\documentclass{ecai} 
\usepackage{latexsym}
\usepackage{amssymb}
\usepackage{amsmath}
\usepackage{amsthm}
\usepackage{booktabs}
\usepackage{enumitem}
\usepackage{graphicx}
\usepackage{color}
\usepackage{multirow}

\usepackage{algorithm}
\usepackage[noend]{algpseudocode}
\usepackage{algorithmicx}
\usepackage{xspace}
\usepackage[table]{xcolor}
\usepackage{microtype}
\usepackage{bbm}

\usepackage{booktabs}
\usepackage{multirow}
\usepackage{enumitem}
\usepackage{graphicx}
\usepackage{color}
\usepackage{amsmath}
\usepackage{amsfonts}
\usepackage{amsthm}
\usepackage{latexsym}
\usepackage{amssymb}

\usepackage{cleveref}

\newtheorem{definition}{Definition}
\newcommand{\xmt}{{\texttt{XMT}}\xspace}
\newcommand{\txmt}{{\texttt{TXMT}}\xspace}
\newcommand{\coopdataset}{{\texttt{Italian Supermarket}}\xspace}
\newcommand{\BibTeX}{B\kern-.05em{\sc i\kern-.025em b}\kern-.08em\TeX}

\begin{document}

%%%%%%%%%%%%%%%%%%%%%%%%%%%%%%%%%%%%%%%%%%%%%%%%%%%%%%%%%%%%%%%%%%%%%%%%

\begin{frontmatter}

%%% Use this command to specify your submission number.
%%% In doubleblind mode, it will be printed on the first page.

\paperid{3466} 

%%% Use this command to specify the title of your paper.

\title{An Interpretable Data-Driven Unsupervised Approach for the Prevention of Forgotten Items}

%%% Use this combinations of commands to specify all authors of your 
%%% paper. Use \fnms{} and \snm{} to indicate everyone's first names 
%%% and surname. This will help the publisher with indexing the 
%%% proceedings. Please use a reasonable approximation in case your 
%%% name does not neatly split into "first names" and "surname".
%%% Specifying your ORCID digital identifier is optional. 
%%% Use the \thanks{} command to indicate one or more corresponding 
%%% authors and their email address(es). If so desired, you can specify
%%% author contributions using the \footnote{} command.

\author[A]{\fnms{Luca}~\snm{Corbucci}\thanks{Corresponding Author Email: luca.corbucci@phd.unipi.it}
}
\author[A]{\fnms{Javier Alejandro}~\snm{Borges Legrottaglie}
}
\author[A]{\fnms{Francesco}~\snm{Spinnato}
} 
\author[A]{\fnms{Anna}~\snm{Monreale}
}
\author[A]{\fnms{Riccardo}~\snm{Guidotti}
} 

\address[A]{University of Pisa}
%\address[B]{ISTI-CNR, Pisa}

%%% Use this environment to include an abstract of your paper.

\begin{abstract}
Accurately identifying items forgotten during a supermarket visit and providing clear, interpretable explanations for recommending them remains an underexplored problem within the Next Basket Prediction (NBP) domain. 
Existing NBP approaches typically only focus on forecasting future purchases, without explicitly addressing the detection of unintentionally omitted items. 
This gap is partly due to the scarcity of real-world datasets that allow for the reliable estimation of forgotten items. 
Furthermore, most current NBP methods rely on black-box models, which lack transparency and limit the ability to justify recommendations to end users.
In this paper, we formally introduce the forgotten item prediction task and propose two novel interpretable-by-design algorithms. 
These methods are tailored to identify forgotten items while offering intuitive, human-understandable explanations.
Experiments on a real-world retail dataset show our algorithms outperform state-of-the-art NBP baselines by 10–15\% across multiple evaluation metrics.
\end{abstract}

\end{frontmatter}

%%%%%%%%%%%%%%%%%%%%%%%%%%%%%%%%%%%%%%%%%%%%%%%%%%%%%%%%%%%%%%%%%%%%%%%%

\section{Introduction}
\label{sec:Introduction}
Recommendation systems are evolving from simple suggestion tools into sophisticated, ubiquitous technologies that deliver personalized item selections across several domains, ranging from books and movies, %and TV series 
to supermarket products and retail transactions.
Especially within supermarkets and online marketplaces~\cite{ko2022survey}, recommendation systems play a crucial role in shaping consumer behavior, improving customer satisfaction, and driving business growth~\cite{sarwar2000analysis}.
A significant portion of current research in retail analytics has focused on Next-Basket Prediction (NBP), the task of forecasting a customer's next purchase based on their historical transaction data~\cite{rendle2010factorizing}. 
Techniques developed for this problem range from models that leverage sequential shopping patterns to complex deep learning architectures. 
While these approaches have shown success in predicting general shopping behaviors, they fall short when it comes to addressing a more nuanced and practically important challenge: predicting forgotten items. 

In the transactional retail setting, forgotten items are products that customers intended to purchase during a shopping trip but inadvertently left behind, often leading to a quick return trip to the store. For instance, we can imagine a customer shopping for their weekly groceries.  Their cart includes jam and peanut butter, but it is missing bread, which they typically purchase along with the other breakfast items. 
A system that detects such omissions could prompt reminders, enhancing customer satisfaction and increasing timely sales. 
This simple scenario exemplifies the need for recommendation models that go beyond predicting what a user might like, but rather infer what they likely intended to buy but overlooked.
Forgotten items create inefficiencies for customers, who must make additional store visits, and retailers, who miss opportunities for immediate sales. Mitigating this problem requires a different methodological approach than conventional NBP, as it involves identifying deviations from an individual’s established shopping patterns, rather than predicting entirely new or evolving preferences.
We empirically demonstrate that, although existing NBP methods can be applied to address the forgotten item problem, their performance on this specific task remains suboptimal.

Another key limitation of today's recommendation systems is that their increasing sophistication often comes at the expense of transparency. Indeed, many state-of-the-art models now operate as black boxes, achieving high predictive accuracy but offering little insight into their decision-making processes~\cite{le2019correlation,yu2016dynamic}. This opacity presents a significant challenge in retail environments, where both businesses and customers benefit from understanding the rationale behind a recommendation~\cite{sinha2002role}. 
The lack of interpretability can undermine user trust and diminish the practical utility of these systems~\cite{rudin2019stop}.
Therefore, there is growing demand for models that not only perform effectively but also offer clear, easy-to-understand explanations, especially in contexts where recommendations influence consumer behavior.

We address these gaps through the following key contributions.

\textit{\textbf{(i)}} We formalize the \textit{Forgotten-Item Prediction} (FIP) problem, establishing a framework for systematically evaluating and comparing methods for this specific task. To support this, we define evaluation metrics that reflect the distinct nature of FIP, setting them apart from standard NBP criteria. 
%    \item 
%Second, 
\textit{\textbf{(ii)}} We propose two 
%interpretable-by-design 
FIP algorithms: \xmt, which integrates historical purchase patterns, temporal dynamics, and contextual signals to identify likely forgotten items; and \txmt, an extension of \xmt that incorporates recurrent purchasing behaviors using TARS~\cite{guidotti2017market,guidotti2019personalized}. Both methods employ a multi-factor scoring system to enhance predictive accuracy. % while preserving interpretability. 
   % \item 
Furthermore, 
\textit{\textbf{(iii)}} \xmt and \txmt are interpretable-by-design, providing human-understandable justifications for each prediction. For instance, revisiting the earlier example, the model may suggest that \textit{bread is a forgotten item because the current basket contains both peanut butter and jam, bread is typically purchased weekly, and the customer last bought it one week ago}.
    %\item 
\textit{\textbf{(iv)}} We evaluate \xmt and \txmt on \textit{real-world retail data}, demonstrating the effectiveness of our models, significantly outperforming both baseline NBP and state-of-the-art FIP methods by margins of 10–15\% across different evaluation metrics.\footnote{
Our code is available here: \url{https://bit.ly/3H0r1C7}}

%The rest of this work is structured as follows. 
%\Cref{sec:related} reports the related works, \Cref{sec:background} recall concepts necessary to understand our proposal.
%In \Cref{sec:problem} we formalize the FIP problem while \Cref{sec:methodology} details the proposed approaches.
%\Cref{sec:experiments} introduces the experimental setting and presents the results of the experiments, \Cref{sec:conclusion} discusses limitations and future works.

\section{Related Works}
\label{sec:related}
Early recommender systems in retail focused on identifying item co-occurrence using collaborative filtering and association rule mining~\cite{sarwar2000analysis}. 
While effective at capturing static affinities, these approaches lacked temporal modeling to anticipate shopping intent. 
To incorporate sequential patterns, various time-aware and personalized strategies emerged.
Simple yet strong baselines include predicting items from the user's most recent basket or selecting the most frequent items from a customer’s purchase history~\cite{cumby}.
More involved approaches use Markov Chains to introduce temporal dynamics as item transitions or binary classifiers on handcrafted temporal features to forecast item reoccurrence~\cite{cumby}.
Matrix factorization methods brought further personalization. For example, Non-negative Matrix Factorization (NMF)~\cite{lee2000algorithms} models latent item-user interactions. 
The Factorized Personalized Markov Chain (FPMC)~\cite{rendle2010factorizing} method combines matrix factorization with personalized Markov chains to account for user preferences and sequential transitions. 
Building on this, Hierarchical Representation Model (HRM)~\cite{wang2015learning} learns unified representations of user and basket histories via hierarchical neural architectures.
The advent of deep learning advanced NBP, enabling modelling of long-range dependencies and complex user behavior through recurrent networks~\cite{yu2016dynamic}, attention mechanisms~\cite{wang2018attention}, and hierarchical encoding~\cite{huang2024knowledge}. 

However, most of these approaches function as black-box models, optimizing predictive performance at the cost of interpretability~\cite{guidotti2019survey}. 
This opacity is particularly problematic in consumer-facing retail environments, where trust, transparency, and actionable insights are paramount.
Indeed, as recommendation models become increasingly complex, explainability has become a crucial concern. 
In contemporary recommender systems, explaining why an item is suggested can be as important as the recommendation’s accuracy for maintaining user trust and satisfaction~\cite{zhang2020explainable}. This is especially true in retail, where customers may be wary of automated suggestions affecting their purchase decisions. 
Historically, straightforward techniques like association rules naturally offered simple explanations, e.g., \textit{``customers who bought $X$ also bought $Y$''}, based on co-purchase statistics~\cite{le2019correlation}. 
However, modern NBP algorithms, from FPMC to deep neural networks, are largely opaque. 
Most next-basket recommendation methods have been optimized for predictive performance, with little attention to transparency~\cite{huang2024knowledge}. 
The lack of interpretability can reduce user acceptance and complicate retailers' validation of recommendations.

Recently, research has started to bridge this gap by infusing explainability into NBP. 
One line of work integrates external knowledge to rationalize recommendations. 
In~\cite{huang2024knowledge}, the authors construct a basket-level knowledge graph using Reinforcement Learning to generate intuitive paths that justify a recommended item, providing explanations alongside the next-basket prediction. Attention-based models also offer a degree of interpretability by revealing which past items or baskets had the most influence on the prediction, albeit implicitly~\cite{wang2018attention}. 
Nevertheless, providing direct and user-friendly explanations in a sequential recommendation setting remains challenging. 
Unlike single-item recommendations, next-basket suggestions may result from complex interactions among multiple prior purchases, which complicates the explanation task.
~\cite{guidotti2017market, guidotti2019personalized} introduces an interpretable Temporal Basket Prediction (TBP) model for NBP based on Temporal Annotated Recurring Sequences (TARS). 
Thanks to TARS, TBP can effectively capture multiple factors influencing customer decision-making, including co-occurrence, sequentiality, periodicity, and recurrence of purchased items.

Within this context, identifying products a user likely intended to purchase but accidentally forgot remains a relatively unexplored task. 
We define this task as Forgotten-Item Prediction (FIP).
Prior work on basket completion touches on related themes, such as suggesting complementary or co-purchased items for an ongoing basket~\cite{guidotti2017txmeans,sarwar2000analysis}, or identifying items that lost purchase relevance over time~\cite{mourao2011oblivion}. However, the task of predicting forgotten items in basket analysis is mostly absent from the literature.
%Although prior work on basket completion touches on related themes, suggesting complementary or co-purchased items given an ongoing basket~\cite{guidotti2017txmeans,sarwar2000analysis}, or identifying items that lost purchase relevance over the years~\cite{mourao2011oblivion}, explicitly addressing forgotten items in basket analysis is mostly absent from the literature.
FIP requires modeling user intention and recall errors, rather than merely predicting what is likely to be bought next, making it a unique and under-investigated extension of the NBP problem.
Indeed, to the best of our knowledge, at the time of writing, the only published work addressing the FIP task, albeit from a slightly different perspective, is~\cite{singh2020prediction}. In~\cite{singh2020prediction}, the authors introduce the customer-centric Interval-Based Predictor (IBP), which addresses the FIP problem by offering customers a list of potentially forgotten items. This is achieved by considering numerical factors such as the average time between visits, the average interval between purchases of a given item, and the time elapsed since the item was last purchased, to estimate when the customer is likely to run out of stock.
However, unlike most existing approaches to NBP, the score for each item is computed independently of the others, without considering item co-occurrence.  
Also, while the numerical estimation allows for some degree of interpretability, it remains notably limited.

In this paper, we address a research gap by formalizing the unexplored FIP problem and proposing an unsupervised, data-driven approach to prevent forgotten items.
Our proposals simultaneously account for co-occurrence, sequentiality, periodicity, recurrence of purchased items, average time between visits, average intervals between purchases of specific items, and the time elapsed since an item was last bought. 
Furthermore, these factors are combined to provide customers with clear and intuitive explanations for why a particular item is being recommended.

\section{Setting the Stage}
\label{sec:background}
This section outlines the key concepts behind our proposal.

\paragraph{Next-Basket Prediction (NBP):}
This task aims to forecast items a user is likely to purchase in their next shopping transaction based on historical shopping behaviors. 
Formally,

\begin{definition}[Next-Basket Prediction]
    Let $U = \{u_1, u_2, \dots, u_n\}$ denote the users and $I = \{i_1, i_2, \dots, i_m\}$ the items available. 
    Each user $u \in U$ has a shopping history $H_u = \{B_1, B_2, \dots, B_{t-1}\}$ with baskets $B_j \subseteq I$. 
    Given $H_u$, NBP predicts the probability distribution $P(i \in B_t \mid H_u)$ for each item $i \in I$.
\end{definition}

\noindent Thus, a NBP function $g$ takes as input the shopping history of a customer $H_u$ and, possibly, also a set of shopping history of a set of customers $\mathcal{H} = \{H_1, \dots, H_n\}$, and returns the predicted next basket $\hat{B}_t = g(H_u, \mathcal{H})$.
Key challenges in the algorithmic structure of $g$ include modeling sequential item dependencies, co-occurrence patterns, and temporal dynamics of user behaviors~\cite{guidotti2019personalized}. 
Temporal Annotated Recurring Sequences (TARS), introduced in~\cite{guidotti2017market,guidotti2019personalized}, capture temporal and recurrent purchasing behavior. In simple terms, a sequence $X \to Y$, is composed of a set of items $X$ called a head, followed later, in another shopping event, by a set of items $Y$ named tail.

We say that a sequence is \textit{minimal} when there is not a more smaller sequence using subsets of these items that occurs within the same time interval. 
TARS are mined from a customer's purchase history $H_u$ by detecting frequent sequences and adapting temporal thresholds to individual behavior. 
Final item scores, $\Omega_i$, assess the relevance of item $i$ appearing in the tails $Y$ of relevant TARS. 
They quantify how relevant each item is for future purchases, based on a customer's recent behavior and historical recurring patterns.

%\smallskip
\paragraph{Explainable AI.} As NBP approaches grow increasingly complex, the need for Explainable AI (XAI) methods to interpret them becomes more pressing. XAI techniques can be classified as either \textit{local} or \textit{global}, depending on whether they explain individual predictions or the model’s overall behavior~\cite{bodria2023benchmarking}. 
They may also be \textit{post-hoc}, applied after training to explain a black-box model, or \textit{ante-hoc}, also known as \textit{interpretable-by-design}, where the model is inherently transparent. 
We focus here on local, interpretable-by-design explainers, which provide direct, faithful insights into decision-making without relying on potentially misleading post-hoc approximations~\cite{rudin2019stop}.

\begin{figure}[t]
    \centering
    \includegraphics[width=0.98\linewidth]{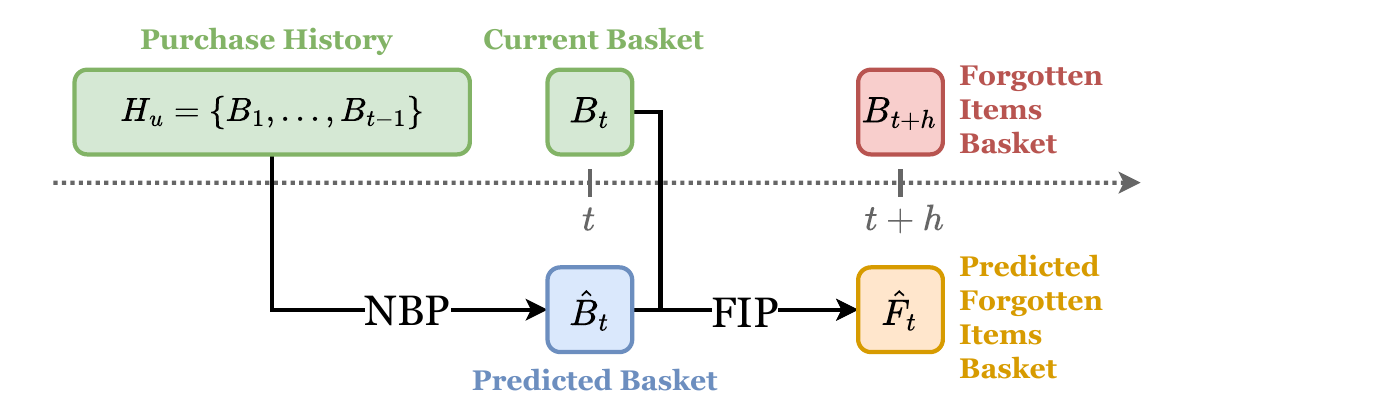}
    \caption{Depiction of the Next Basket Prediction (NBP), and Forgotten Items Prediction (FIP) problems. In NBP, the user's purchase history $H_u$ is used to predict the current basket at time $t$, while FIP is used to predict which items the user is forgetting at the current time $t$.}
    \label{fig:schema}
\end{figure}

\section{Problem Formulation}
\label{sec:problem}
We define the Forgotten-Items Prediction (FIP) problem as follows. 

\begin{definition}[Forgotten-Items Basket]
    Let $B_t$ represent a basket transaction purchased at time $t$. 
    We name the basket $B_{t+h}$, occurring after a temporal horizon $h$ relative to $t$, as a \emph{forgotten-items basket} if the customer intended to purchase the items in $B_{t+h}$ at the time $t$ together with the items in $B_t$, i.e., the actual intended purchase should have been $B^*_t = B_t \cup B_{t+h}$.
\end{definition}
     
\noindent Thus, a FIP function $g$ takes as input the current basket $B_t$, the shopping history of the customer $H_u$, and, depending on its nature, also a set of shopping history of a set of customers $\mathcal{H}$, and returns the predicted forgotten-items basket $\hat{F}_t = g(B_t, H_u, \mathcal{H})$.
The temporal horizon $h$ can be set based on observed shopping patterns and helps differentiate genuine forgotten-item purchases from regular shopping.
For instance, it can be set in an interval between 0 days (i.e., re-purchasing the same day) and 2 days (i.e., repurchasing at most after 2 days). 
The intention of a customer to purchase two sequential baskets together can be inferred through various means, such as customer interviews or heuristics applied to their shopping history.
Details specific to our case study are provided in \Cref{sec:exp_setting}.

\smallskip
We can now introduce the Forgotten-Item Prediction (FIP) problem, which has the objective to predicts items that are likely to be forgotten during the current transaction. 

\begin{definition}[Forgotten-Item Prediction]
    Let $U = \{u_1, u_2, \dots, u_n\}$ denote the users, $I = \{i_1, i_2, \dots, i_m\}$ the items available and $H_u = \{B_1, B_2, \dots, B_{t-1}\}$ the shopping history of each user with baskets $B_j \subseteq I$. 
    Given $H_u$, FIP consists in predicting the probability distribution $P(i \in B_{t+h} \mid H_u, B_t)$ for each item $i \in I$ where $B_{t+h}$ is a forgotten-items basket of $B_t$.
\end{definition}

\noindent Hence, the objective of FIP is to predict the forgotten-items basket $\hat{F}_t\approx B_{t+h}$ given the customer history $H_u$ and the current basket $B_t$. 
A depiction of the NBP and FIP tasks can be found in \Cref{fig:schema}. We report in~\Cref{tab:notation_summary_extended} the list of symbols used in this paper to increase clarity and readability.

\begin{table}[t]
\footnotesize
    \caption{Summary of notation.}
    \label{tab:notation_summary_extended}
    \centering
    \renewcommand{\arraystretch}{1.1}
    
    \begin{tabular}{ll}
    \toprule
    %\textbf{Notation} & \textbf{Meaning} \\
    \multicolumn{2}{l}{\textbf{Entities and Variables}} \\
    \midrule
    $\mathcal{H}, H$ & set of user histories, single history \\
    $U, u$ & set of users, user \\
    $I, i$ & set of items, item \\
    $B_t, B_{t+h}$ & basket at time $t$, forgotten-items basket \\
    $\hat{B}_t, \hat{F}_t$ & predicted basket, predicted forgotten items \\
    $g$ & NBP or FIP function \\
    \midrule
    \multicolumn{2}{l}{\textbf{Item Properties}} \\
    \midrule
    $\eta_i$ & median inter-purchase interval \\
    $\phi_i$ & days since last purchase \\
    $\rho_i$ & seasonal concentration ratio \\
    $\chi(i,j)$ & co-purchase count with item $j$ \\
    $\zeta_i(h)$ & repurchases within $h$ \\
    $\lambda_i$ & repurchase opportunities after large baskets \\
    \midrule
    \multicolumn{2}{l}{\textbf{Item Scores}} \\
    \midrule
    $f_i, \tau_i, \sigma_i$ & frequency, temporal, seasonal scores \\
    $\kappa_i, \psi_i$ & context affinity, repurchase tendency \\
    $\Omega_i$ & TARS-based score \\
    $\mathit{MAP}_i, \mathit{TMAP}_i$ & combined prediction scores \\
    \midrule
    \multicolumn{2}{l}{\textbf{Hyperparameters}} \\
    \midrule
    $\epsilon, \gamma, G$ & min. freq., max. interval mult., discontinue threshold \\
    $\alpha, \upsilon, \beta$ & temporal boost, seasonality thresh., seasonal boost \\
    $\Upsilon, \Theta$ & co-purchase thresh. and boost \\
    $\nu, \Lambda, \Phi$ & repurchase window, boost, rate threshold \\
    \bottomrule
    \end{tabular}
\end{table}

\section{Methodology}
\label{sec:methodology}
We present here two interpretable unsupervised algorithms to tackle the FIP problem: the \textit{eXplainable Missing items considering Time} (\xmt), and \textit{TARS-based \xmt} (\txmt). 
For both algorithms, we face the FIP problem exploiting NBP as follows.

\begin{definition}[Forgotten-Item Prediction via Next-Basket Prediction]
    Let $B_t$ be the basket purchased at current time $t$, $B_{t+h}$ the forgotten-items basket (unknown at time $t$), and $\hat{B}_t \subset I$ the prediction for the basket purchased at time $t$ given by an NBP algorithm $g$.
    We solve FIP via NBP by approximating the prediction of the forgotten items $B_{t+h}$ as the set of items predicted at time $t$ but not purchased $\hat{F}_t = \hat{B}_{t} - B_t$.
\end{definition}

\noindent The objective of FIP via NBP is to define a function $g$ that outperforms the user in remembering which items should be purchased, i.e., achieving higher recall than the user. Consequently, let $\hat{B}_t \cap B_t$ denote the set of items both purchased and correctly predicted to be purchased, $\hat{F}_t$ represent the set of items \textit{potentially forgotten} by the user. 
Therefore, a core challenge in FIP via NBP is distinguishing genuinely forgotten items from intentionally omitted ones while maintaining strong predictive precision.
In this setting, any NBP method can, in principle, be applied to address the FIP problem.
However, in the experimental section, we demonstrate that standard NBP approaches, when evaluated by comparing $B_{t+h}$ against $\hat{F}_t$ in cases where $B_{t+h} \neq \emptyset$, i.e., some items were indeed forgotten, are less effective than the two tailored methods we introduce, which are specifically designed to address FIP, while remaining within the NBP formulation.
Finally, we highlight a latent yet critical issue that must be addressed alongside FIP: identifying real situations in which FIP is needed.
Due to the highly imbalanced nature of real-world data, where users typically do not forget items, we leave this aspect for future research.
In this study, we focus exclusively on scenarios where the next basket is known to be a forgotten-items basket.
However, a simple yet practical strategy is to consistently present to customers at purchase time a small set $\hat{F}_t$ of $k$ potentially forgotten items.
Indeed, in the worst-case scenario, non-forgotten items will simply be ignored, incurring no harm.

\subsection{\xmt}
\label{sec:methodology:xmt}
The \textit{eXplainable Missing items considering Time} (\xmt) method, detailed in \Cref{alg:xmt}, introduces a multi-factor scoring approach that combines historical purchase patterns, temporal relationships, and contextual evidence to identify potentially forgotten items. 
The forgotten item prediction process unfolds in two phases. 
In the first phase (lines \ref{alg:xmt:base_freq}-\ref{alg:xmt:seas_cont} in \Cref{alg:xmt}), \xmt constructs the predicted basket $\hat{B}_t$ by selecting the top items ranked by the temporal and seasonal purchase likelihood, $\sigma_i$. This predicted basket $\hat{B}_t$ (line \ref{alg:xmt:select_basket} in \Cref{alg:xmt}) represents items that the customer is expected to purchase at time $t$, based purely on historical frequency, recency, and seasonality. 

In the second phase (lines \ref{alg:xmt:copurchase}-\ref{alg:xmt:map_score}), the algorithm computes the Multi-factor Adjusted Prediction (MAP) score $\mathit{MAP}_i$ for each item $i \in \hat{B}_t$: 

\begin{equation}
\label{eq:map}
    \mathit{MAP}_i = \sigma_i + \kappa_i + \psi_i
\end{equation}

\noindent which integrates three components: the previously computed seasonally adjusted purchase likelihood $\sigma_i$, the basket context affinity $\kappa_i$ (line \ref{alg:xmt:copurchase}), and the post-basket repurchase tendency $\psi_i$ (line \ref{alg:xmt:repurchase}). 
The final forgotten item predictions are obtained by selecting the top $k$ items from $\hat{F}_t$ ranked by their MAP scores (line \ref{alg:xmt:select_forgotten}). This two-phase process ensures that the predictions are first guided by long-term behavioral signals and then refined by session-specific contextual information. 
In the following, we detail how to compute each component.

\begin{algorithm}[t]
\footnotesize
    \caption{\xmt}
    \begin{algorithmic}[1]
    \Require Customer purchase history $H_u$, current time $t$, minimum appearances $\epsilon$, maximum interval multiplier $\gamma$, discontinuation threshold $G$, temporal score boost $\alpha$, seasonality threshold $\upsilon$, seasonal boost factor $\beta$, co-purchase threshold $\Upsilon$, co-purchase boost $\Theta$, repurchase window $\nu$, repurchase boost $\Lambda$, repurchase threshold $\Phi$, prediction length $k$.
    \Ensure $\hat{F_t}$
    
    \State $f \gets \text{calculateBaseFrequency}(H_u, \epsilon)$ \label{alg:xmt:base_freq} \Comment{Eq. \ref{eq:frequency}}
    \State $\tau \gets \text{calculateTemporalProximity}(H_u, f, t, \gamma, G, \alpha)$  \label{alg:xmt:temp_prox} \Comment{Eq. \ref{eq:tau}}
    \State $\sigma \gets \text{calculateSeasonalContext}(H_u, \tau, \upsilon, \beta)$ \label{alg:xmt:seas_cont} \Comment{Eq. \ref{eq:sigma}}
    \State $\hat{B}_t \gets \{\text{top } |B_t|+k  \text{ items according to }\sigma\}$ \label{alg:xmt:select_basket}
    \State $\kappa \gets \text{calculateCoPurchase}(H_u, B_t, \Upsilon, \Theta)$  \label{alg:xmt:copurchase} \Comment{Eq. \ref{eq:kappa}}
    \State $\psi \gets \text{calculateRepurchasePattern}(H_u, \theta, \nu, \Lambda, \Phi, O)$ \label{alg:xmt:repurchase} \Comment{Eq. \ref{eq:psi}}
    \State $\textit{MAP} \gets \text{calculateMAPScore}(\sigma, \kappa, \psi)$ \label{alg:xmt:map_score} \Comment{Eq. \ref{eq:map}}
    \State $\hat{F}_t \gets \{\text{top $k$ items of } (\hat{B}_t \setminus B_t) \text{ according to }\textit{MAP}\}$ \label{alg:xmt:select_forgotten}

    \State \Return $\hat{F}_t$
    \end{algorithmic}
    \label{alg:xmt}
\end{algorithm}

\paragraph{Temporal and Seasonal Purchase Likelihood.}
The first term $\sigma_i$, captures long-term purchasing patterns, short-term recency effects, and seasonal regularity. The relative purchase frequency is:

\begin{equation}
\label{eq:frequency}
    f_i = \frac{|H_u^{(i)}|}{|H_u|} \cdot \mathbbm{1}\left[|H_u^{(i)}| \geq \epsilon\right],
\end{equation}

\noindent where $H_u$ is the customer’s purchase history, $H_u^{(i)} = \{B_j \in H_u \mid i \in B_j\}$ is the set of baskets in the history containing item $i$, $\mathbbm{1}$ is the indicator function that uses the threshold $\epsilon$ to filter out low-frequency items (Line~\ref{alg:xmt:base_freq}). 
%It is important to choose a reasonable value for $\epsilon$ because it helps in filtering out items with too few purchases. 
We treated $\epsilon$ as a hyperparameter, and we provide more details in Appendix B of Supplementary Material (SM)~\cite{forgotten2025arxiv}. 
As shown by the effectiveness of the TOP method in~\cite{guidotti2017market}, the relative purchase frequency provides an important estimate for an item's inclusion in any basket. 
Items with higher purchase frequencies are more likely to be regular purchases and thus more likely to be forgotten if missing from a large basket.
Then, inspired by IBP~\cite{singh2020prediction}, \xmt incorporates temporal patterns into a scoring component $\tau_i$ that boosts items that are ``due'' for purchase based on their typical purchase cycle (line~\ref{alg:xmt:temp_prox}):

\begin{equation}
\label{eq:tau}
    \tau_i = f_i \cdot \left(1 + (\alpha - 1) \cdot \mathbbm{1}\left[\eta_i \leq \phi_i \leq \eta_i \cdot \gamma \land \phi \leq G\right] \right)
\end{equation}

\noindent where $\phi_i$ is the number of days since the last purchase of item $i$, $\eta_i$ is the median inter-purchase interval, and $\alpha$, $\gamma$, and $G$ are hyperparameters controlling the strength and eligibility of the temporal adjustment; more details on how we chose them are provided in Appendix B of the SM~\cite{forgotten2025arxiv}.
The temporal proximity computation scales the frequency score $f_i$ by a factor $\alpha$, capturing an increased likelihood of imminent repurchase.
This boost prioritizes items that align with the customer's established purchase rhythm. 
On the contrary, the algorithm excludes items that have not been purchased for an extended period to avoid suggesting discontinued purchases. 
After that, \xmt computes a seasonal adjustment $\sigma_i$ that identifies and boosts items with strong seasonal tendencies (Line~\ref{alg:xmt:seas_cont}) computed as follows:

\begin{equation}
\label{eq:sigma}
    \sigma_i = \tau_i \cdot \left(1 + (\beta - 1) \cdot \mathbbm{1}\left[\rho_{i} > \upsilon\right] \right)
\end{equation}

\noindent where $\rho_{i}$ is the fraction of annual purchases of item $i$ that occur in the current season, and $\upsilon$ and $\beta$ are hyperparameters defining the seasonality threshold and seasonal boost, respectively. More details about how we selected them are reported in Appendix B of the SM~\cite{forgotten2025arxiv}.

\paragraph{Basket Context Affinity.}
The second term $\kappa_i$ measures the contextual relevance of item $i$ given the content of the current basket $B_t$. 
This is computed by summing over all items $j \in B_t$ that have a significant history of co-purchase with $i$:

\begin{equation}
\label{eq:kappa}
    \kappa_i = \sum_{j \in B_t} \Theta \cdot \mathbbm{1}\left[\chi(i,j) > \Upsilon\right]
\end{equation}

\noindent where $\chi(i,j)$ is the number of past baskets in which items $i$ and $j$ co-occurred, and $\Upsilon$ and $\Theta$ control the minimum required co-occurrence frequency and the score boost per co-purchase, respectively (details about how we selected them are reported in Appendix B of the SM~\cite{forgotten2025arxiv}). 
This component reflects associative strength and ensures that items frequently bought together are considered in tandem.

\paragraph{Post-Basket Repurchase Tendency.}
The final component $\psi_i$ accounts for items that are commonly repurchased soon after large baskets, indicating a short-term replenishment behavior. 

\begin{equation}
\label{eq:psi}
    \psi_i = \Lambda \cdot \mathbbm{1}\left[ \frac{\sum_{h \leq \nu} \zeta_i(h)}{\lambda_i} > \Phi \right]
\end{equation}

\noindent where $\zeta_i(h)$ is the number of times item $i$ was repurchased within $h$ days after the most recent basket, $\lambda_i$ is the number of such opportunities, $\nu$ is the repurchase window, and $\Phi$ and $\Lambda$ are the repurchase rate threshold and corresponding boost.
More details on how we selected them are reported in Appendix B of the SM~\cite{forgotten2025arxiv}. 
This term ensures that high-velocity repurchase patterns influence the final score even if the item’s regular frequency is moderate.

\subsection{\txmt}
Building upon \xmt, the \textit{TARS-based eXplainable Missing items considering Time} (\txmt) extends its capabilities by integrating temporal patterns extracted through TARS~\cite{guidotti2017market}. %(see~\Cref{sec:background} for a short overview on TARS).
The incorporation of TARS is crucial for capturing complex temporal purchasing behaviors, thereby enhancing the accuracy of the algorithm's predictions.
However, this improvement comes at the cost of increased computational complexity due to the overhead introduced by the TARS extraction process.

To effectively integrate the TARS scores into \xmt, \txmt modifies the MAP score by introducing an additional scoring component. 
First, the TARS scores $\Omega_i$ are computed using the customer's purchase history $H_u$ as described in~\cite{guidotti2017market}.
Subsequently, the TARS scores $\Omega_i$ are normalized by the maximum TARS score. 
These normalized scores are then added to the MAP scores creating the TARS-based MAP score $\mathit{TMAP}_i$ for item $i$ as follows:
\begin{equation}
    \mathit{TMAP}_{i} = \mathit{MAP}_i + \Omega_i
\end{equation}

\noindent Similarly to \xmt, given the $\mathit{TMAP}_{i}$ values, \txmt returns the top $k$ items ranked by their TMAP scores.

\subsection{Interpretability of \xmt and \txmt}
\label{sec:interpretability}
As discussed in~\Cref{sec:background}, the increasing availability of black-box models has improved performance but often at the cost of interpretability, thereby diminishing user trust and limiting practical adoption. In contrast, both \xmt and TXTM are interpretable-by-design, owing to the proposed $\mathit{MAP}$ and $\mathit{TMAP}$ scoring algorithms, which rely solely on the combination of distinct scoring components. Unlike black-box approaches, each score type in our methods offers clear and actionable insights that can be directly communicated to users.

Indeed, through \xmt, the different scores capture immediate contextual evidence and temporal relationships. A high base frequency $f_i$ indicates the item is a regular part of the customer's purchase history. 
A temporal score boost from the Temporal Proximity Component $\tau_i > f_i$ suggests the item is ``due'' for repurchase based on their typical buying cycle. 
A boost from the Seasonal Context Component ($\sigma_i > \tau_i$) points to a higher likelihood of purchase due to the current season. 
A positive Co-Purchase score ($\kappa_i > 0$ indicates the item is frequently bought together with items currently in the basket. 
Finally, a positive Repurchase Pattern score ($\psi_i > 0$) suggests the item is often re-bought shortly after large shopping trips. 
The additive nature of the Multi-factor Adjusted Prediction (MAP) score ($MAP_i = \sigma_i + \kappa_i + \psi_i$) allows for a decomposition of the factors influencing the prediction for each item, enhancing the transparency of the model's reasoning. 
By presenting the contribution of each component, \xmt offers interpretable predictions grounded in distinct aspects of the customer's purchasing behavior.

For \txmt, this combination used to compute $\mathit{TMAP}_{i}$ maintains the interpretability of both \xmt and TARS while allowing each type of evidence to contribute independently to the final prediction. 
The TARS component contributes interpretability by highlighting items that are part of learned recurring purchase sequences, indicating established temporal dependencies in the customer's behavior.

\section{Experiments}
\label{sec:experiments}
We outline here the experimental setup and present a selection of the obtained results. 
We begin by describing the dataset and the preprocessing steps used to identify forgotten-items baskets. 
Next, we define the evaluation metrics employed to assess model performance. 
We then introduce the baseline models used for comparison with \xmt and \txmt. 
Finally, we report the results and provide a qualitative example illustrating the explainability of our proposed methods.

\begin{table}[t]
    \scriptsize
  \centering
  \caption{Dataset Statistics with different parameters to define forgotten‐items baskets.}
  \label{tab:merged_coop_stats_forgotten}

  \resizebox{\linewidth}{!}{%
  \begin{tabular}{cccccc}
    \toprule
    \textbf{Customers} & \textbf{Unique Items} & \textbf{Baskets} & \textbf{Med. Hist. Len} & \textbf{Med. Days} & \textbf{Med. Basket Size} \\
    \midrule
    17,132 & 524 & 10.87M & 452 & 3.50 & 9.52 \\
    \bottomrule
  \end{tabular}
  }%
  
  \begin{tabular}{cccccc}
  & & & & & \\
  \end{tabular}
  \setlength{\tabcolsep}{3.1mm} 
  \begin{tabular}{ccccc}
    \toprule
    $\theta_t$ & $\theta_{t+h}$ & $h$ & \# \textbf{Forgotten Baskets} & \textbf{\% Forgotten Basket} \\
    \midrule
    \multirow{3}{*}{10} & \multirow{3}{*}{10} & 0 &   8,495     &  0.08\%  \\
                       &                     & 1 & 492,044   &  4.53\%  \\
                       &                     & 2 & 1,050,499 &  9.67\%  \\
    \cmidrule(r){1-5}
    \multirow{3}{*}{10} & \multirow{3}{*}{1}  & 0 &  49,314    &  0.45\%  \\
                       &                     & 1 & 1,207,712 & 11.11\%  \\
                       &                     & 2 & 2,043,673 & 18.80\%  \\
    \bottomrule
  \end{tabular}
  %}%
\end{table}

%\subsection{Datasets and Pre-processing}
%\label{sec:exp_setup:dataset}
\subsection{Experimental Setting}
\label{sec:exp_setting}
To evaluate \xmt and \txmt, we relied on a proprietary dataset collected by an Italian supermarket chain. 
~\Cref{tab:merged_coop_stats_forgotten} provides a detailed overview of this dataset, which we call the \coopdataset dataset. It includes 10,867,976 individual transactions by 17,132 customers from 2007 to 2016, with a median basket size of 9.52 (Pre-processing details are in Appendix A of SM~\cite{forgotten2025arxiv}).
The products that are presented in this dataset are uniquely identified by an ID, and we mapped these IDS to a general category or product (524 unique items).
Widely used supermarket datasets, such as the \texttt{Ta Feng}\footnote{\url{https://www.kaggle.com/datasets/chiranjivdas09/ta-feng-grocery-dataset}} dataset, are typically used to evaluate algorithms on the NBP or product assortment decisions task. 
However, these datasets are not well-suited for the FIP task, which requires a large amount of expenses and users to effectively identify and label forgotten expenses in a pre-processing phase. 
With over 10 million transactions, the \coopdataset dataset is a perfect candidate for evaluating \xmt and \txmt on the FIP task. Moreover, it has been used in prior work~\cite{immigrants2021guidotti}, and a synthetic version is publicly available.\footnote{Synthetic dataset available after free registration at \url{http://bit.ly/4lLTVoE}.}

Given the absence of explicit forgotten purchase annotations in \coopdataset, i.e., we have no real ground truth for forgotten-items baskets $B_{t+h}$, it is necessary to define a set of criteria to identify baskets that likely represent forgotten items. % based on the purchasing patterns of the user. 
In particular, we mark a basket as forgotten if:
\begin{itemize}[noitemsep,topsep=0pt]
    \item The forgotten-items basket $B_{t+h}$ must be purchased after a \textit{large basket} $B_t$ containing at least $\theta_t$ items, i.e., $|B_t| > \theta_t$. 
    In our experiments, we selected  $\theta_t=10$ to avoid small daily purchases, such as those containing only bread or milk, from being mistakenly flagged as forgotten expenses.
    %\item \textbf{Large Basket} ($\theta$): Given a threshold $\theta$, number of items that a basket $B_t \subseteq I$ must contain to be considered a \textit{large basket}, where $I = {i_1, i_2, ..., i_m}$ represents the set of all possible items available for purchase. In our experiments, we consider $\theta = 10$.
    \item The forgotten-items basket $B_{t+h}$ must be purchased after a maximum of $h$ \textsc{days} after a large basket.
    We report the results of the experiments obtained with $0 \leq h \leq 2$, i.e., purchases made between the same day of $B_t$, i.e., $h=0$, and two days after $B_t$, i.e., $h=2$. 
    The choice of these values is motivated by the observation that the most likely forgotten-items baskets tend to occur shortly after the \textit{large basket} purchase while extending the horizon further would risk capturing unrelated expenses.
    \item The forgotten-items basket  $B_{t+h}$ cannot more than $\theta_{t+h}$ items. 
    We consider experiments $\theta_{t+h} = 1$ and $\theta_{t+h} = 10$.
    However, to better capture scenarios where customers, the main results reported in the paper are those with $\theta_{t+h} = 10$.
    Indeed, upon returning to the store, a customer can take advantage of the trip to purchase additional items beyond just the forgotten ones.
\end{itemize}

\noindent Alongside the dataset details,~\Cref{tab:merged_coop_stats_forgotten} presents a summary of the forgotten items identified in the dataset following the filtering process. 
Specifically, we compare two different value for the parameter $\theta_{t+h}$, i.e., 1 and 10, and three different values for $h$, i.e., 0, 1, and 2, to show how they impact the amount of baskets labeled as forgotten. 
As shown in~\Cref{tab:merged_coop_stats_forgotten}, increasing $h$, leads to a higher number of forgotten-items baskets. 
In contrast, increasing the $\theta_{t+h}$ parameter results in fewer baskets labeled as forgotten-items baskets. 

It is important to highlight that thanks to our user-centric approach, each user can choose hyperparameters that best suit their needs for more tailored predictions. For example, the size $\theta_t$ of the \textit{large basket} preceding a forgotten-item basket $B_{t+h}$ can be smaller for users with typically small baskets and larger for those with bigger ones.

\smallskip
To evaluate the performance of the proposed methods, we implemented a \textit{split-based} model validation strategy. 
For each customer $u \in U$, we partition their shopping history $H_u = \{B_1, B_2, \dots, B_m\}$ into training and testing sets using different percentage splits. 
For instance, with the \textit{30\% Train} criteria, the model learns from the first 30\% of a customer's transaction history.
Then the models are evaluated to predict the first forgotten-items basket present in the test set (if at least one is present).
The evaluation metrics reported in the paper are the average and the standard deviation of all the predictions computed on the test set. 
This lets us evaluate how the amount of historical data affects model performance.
To assess the quality of a prediction of a model to solve the FIP problem, we employ the following evaluation metrics where $\hat{F}_t$ denotes the set of predicted forgotten items at time $t$, while $B_{t+h}$ represents the set of items actually forgotten at time $t$ and subsequently purchased at time $t+h$.
\begin{itemize}[noitemsep,topsep=0pt]
   \item \textit{Precision}: the ratio of correctly predicted items to the total number of predicted items: 
   $\mathit{Precision} = \frac{|\hat{F}_t \cap B_{t+h}|}{|\hat{F}_t|}$
   where 
   \item \textit{Recall}: the ratio of correctly predicted items to the total number of actual items: $\mathit{Recall} = \frac{|\hat{F}_t \cap B_{t+h}|}{|B_{t+h}|}$.
   \item \textit{F1-Score}: the harmonic mean of Precision and Recall: \\$\mathit{F1} = 2 \cdot \frac{\mathit{Precision} \cdot \mathit{Recall}}{\mathit{Precision} + \mathit{Recall}} = \frac{2|\hat{F}_t \cap B_{t+h}|}{|\hat{F}_t| + |B_{t+h}|}$.
\end{itemize}

\begin{figure}[t]
    \centering
    \includegraphics[width=\linewidth]{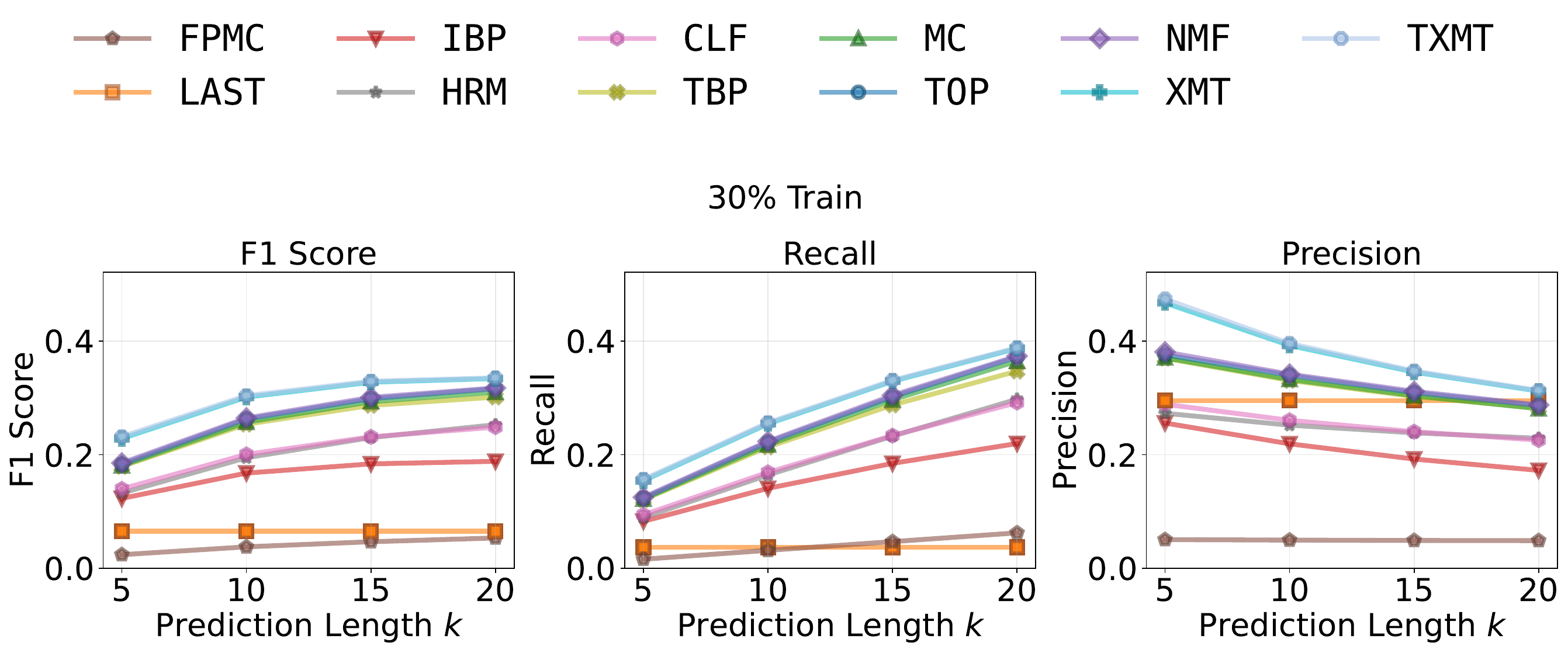}\\
    \vspace{0.1cm}
    \includegraphics[width=\linewidth]{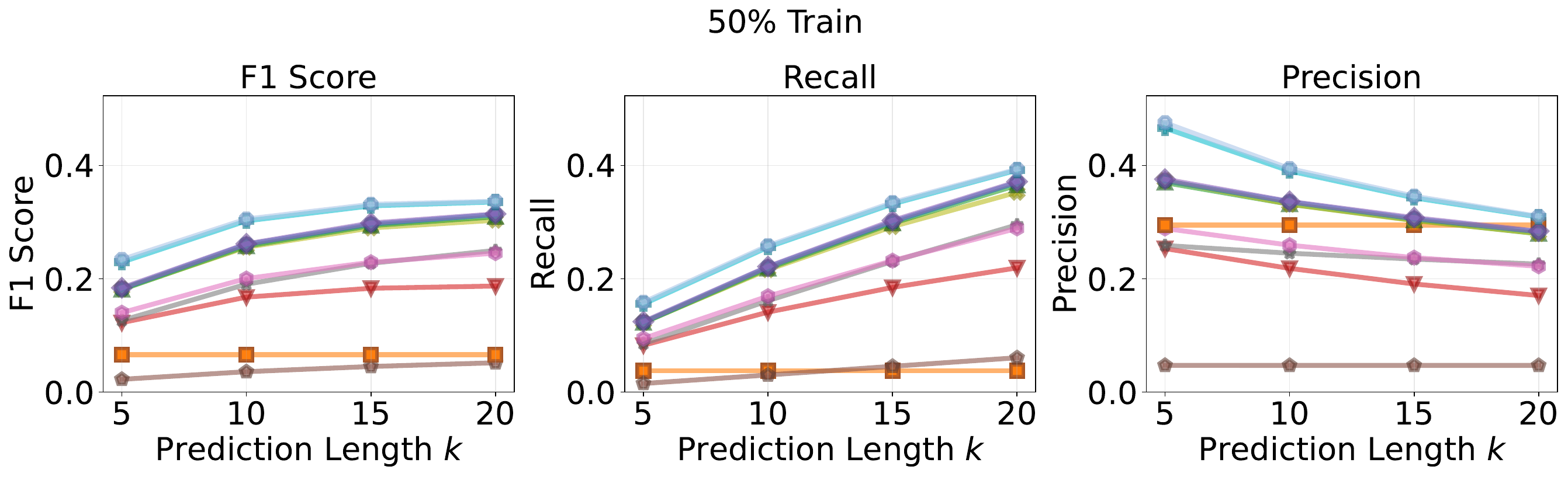}\\\vspace{0.1cm}
    \includegraphics[width=\linewidth, clip, trim= 0 0 0 0]{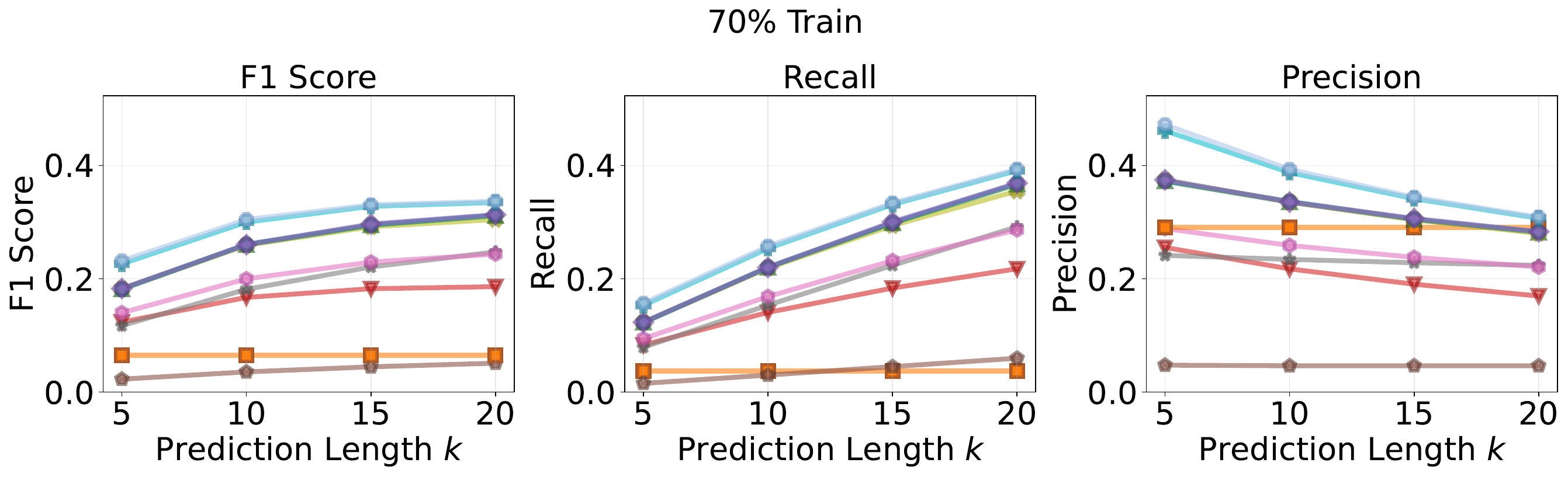}
    \caption{Comparison of \xmt and \txmt against baselines and competitor using \textit{30\%}, \textit{50\%}, \textit{70\%} Train with $h=2$. Prediction length $k$ from 5 to 20.}
    \label{fig:coop_50_day_2}
\end{figure}

\begin{table*}[!t]
\caption{F1-score for different prediction lengths $k$ and data splits. For each row, the best performer is in bold, and the best performer runner-up is underlined.}
\label{tab:f1_score}
\centering
\resizebox{\linewidth}{!}{
\begin{tabular}{cc>{\columncolor{gray!20}}cc>{\columncolor{gray!20}}cc>{\columncolor{gray!20}}cc>{\columncolor{gray!20}}cc>{\columncolor{gray!20}}cc>{\columncolor{gray!20}}c}
\toprule
$k$ & {split} & \texttt{CLF} & \texttt{FPMC} & \texttt{HRM} & \texttt{IBP} & \texttt{LAST} & \texttt{MC} & \texttt{NMF} & \texttt{TBP} & \texttt{TOP} & \texttt{TXMT} & \texttt{XMT} \\
\midrule
5 & 30 & 0.14 $\pm$ 0.10 & 0.02 $\pm$ 0.05 & 0.13 $\pm$ 0.10 & 0.12 $\pm$ 0.10 & 0.07 $\pm$ 0.00 & 0.18 $\pm$ 0.11 & \underline{0.19 $\pm$ 0.11} & 0.18 $\pm$ 0.11 & 0.18 $\pm$ 0.11 & \textbf{0.23 $\pm$ 0.12} & \textbf{0.23 $\pm$ 0.12} \\
5 & 50 & 0.14 $\pm$ 0.10 & 0.02 $\pm$ 0.05 & 0.13 $\pm$ 0.10 & 0.12 $\pm$ 0.10 & 0.07 $\pm$ 0.00 & \underline{0.18 $\pm$ 0.11} & \underline{0.18 $\pm$ 0.11} & \underline{0.18 $\pm$ 0.11} & \underline{0.18 $\pm$ 0.11} & \textbf{0.23 $\pm$ 0.12} & \textbf{0.23 $\pm$ 0.12} \\
5 & 70 & 0.14 $\pm$ 0.10 & 0.02 $\pm$ 0.05 & 0.12 $\pm$ 0.09 & 0.12 $\pm$ 0.10 & 0.07 $\pm$ 0.00 & \underline{0.18 $\pm$ 0.11} & \underline{0.18 $\pm$ 0.11} & \underline{0.18 $\pm$ 0.11} & \underline{0.18 $\pm$ 0.11} & \textbf{0.23 $\pm$ 0.12} & \textbf{0.23 $\pm$ 0.12} \\

\midrule
10 & 30 & 0.20 $\pm$ 0.11 & 0.04 $\pm$ 0.05 & 0.19 $\pm$ 0.10 & 0.17 $\pm$ 0.11 & 0.07 $\pm$ 0.00 & \underline{0.26 $\pm$ 0.12} & \underline{0.26 $\pm$ 0.12} & 0.25 $\pm$ 0.12 & \underline{0.26 $\pm$ 0.12} & \textbf{0.30 $\pm$ 0.13} & \textbf{0.30 $\pm$ 0.13} \\
10 & 50 & 0.20 $\pm$ 0.11 & 0.04 $\pm$ 0.05 & 0.19 $\pm$ 0.10 & 0.17 $\pm$ 0.11 & 0.07 $\pm$ 0.00 & 0.26 $\pm$ 0.12 & 0.26 $\pm$ 0.11 & 0.26 $\pm$ 0.12 & 0.26 $\pm$ 0.12 & \textbf{0.31 $\pm$ 0.13} & \underline{0.30 $\pm$ 0.13} \\
10 & 70 & 0.20 $\pm$ 0.11 & 0.04 $\pm$ 0.05 & 0.18 $\pm$ 0.10 & 0.17 $\pm$ 0.11 & 0.07 $\pm$ 0.00 & \underline{0.26 $\pm$ 0.12} & \underline{0.26 $\pm$ 0.12} & \underline{0.26 $\pm$ 0.12} & \underline{0.26 $\pm$ 0.12} & \textbf{0.30 $\pm$ 0.13} & \textbf{0.30 $\pm$ 0.13} \\

\midrule
15 & 30 & 0.23 $\pm$ 0.11 & 0.05 $\pm$ 0.05 & 0.23 $\pm$ 0.10 & 0.18 $\pm$ 0.11 & 0.07 $\pm$ 0.00 & 0.29 $\pm$ 0.11 & \underline{0.30 $\pm$ 0.11} & 0.29 $\pm$ 0.12 & \underline{0.30 $\pm$ 0.11} & \textbf{0.33 $\pm$ 0.13} & \textbf{0.33 $\pm$ 0.13} \\
15 & 50 & 0.23 $\pm$ 0.11 & 0.05 $\pm$ 0.05 & 0.23 $\pm$ 0.10 & 0.18 $\pm$ 0.10 & 0.07 $\pm$ 0.00 & 0.29 $\pm$ 0.11 & \underline{0.30 $\pm$ 0.11} & 0.29 $\pm$ 0.12 & \underline{0.30 $\pm$ 0.11} & \textbf{0.33 $\pm$ 0.12} & \textbf{0.33 $\pm$ 0.12} \\
15 & 70 & 0.23 $\pm$ 0.10 & 0.04 $\pm$ 0.05 & 0.22 $\pm$ 0.10 & 0.18 $\pm$ 0.11 & 0.07 $\pm$ 0.00 & 0.29 $\pm$ 0.11 & \underline{0.30 $\pm$ 0.11} & 0.29 $\pm$ 0.12 & \underline{0.30 $\pm$ 0.11} & \textbf{0.33 $\pm$ 0.12} & \textbf{0.33 $\pm$ 0.12} \\

\midrule
20 & 30 & 0.25 $\pm$ 0.10 & 0.05 $\pm$ 0.05 & 0.25 $\pm$ 0.10 & 0.19 $\pm$ 0.10 & 0.07 $\pm$ 0.00 & 0.31 $\pm$ 0.11 & 0.32 $\pm$ 0.11 & 0.30 $\pm$ 0.11 & 0.32 $\pm$ 0.11 & \textbf{0.34 $\pm$ 0.12} & \underline{0.33 $\pm$ 0.12} \\
20 & 50 & 0.25 $\pm$ 0.10 & 0.05 $\pm$ 0.05 & 0.25 $\pm$ 0.10 & 0.19 $\pm$ 0.10 & 0.07 $\pm$ 0.00 & \underline{0.31 $\pm$ 0.11} & \underline{0.31 $\pm$ 0.11} & 0.30 $\pm$ 0.11 & \underline{0.31 $\pm$ 0.11} & \textbf{0.34 $\pm$ 0.12} & \textbf{0.34 $\pm$ 0.12} \\
20 & 70 & 0.24 $\pm$ 0.10 & 0.05 $\pm$ 0.05 & 0.25 $\pm$ 0.10 & 0.19 $\pm$ 0.10 & 0.07 $\pm$ 0.00 & 0.31 $\pm$ 0.11 & 0.31 $\pm$ 0.11 & 0.30 $\pm$ 0.11 & 0.31 $\pm$ 0.11 & \textbf{0.34 $\pm$ 0.12} & \underline{0.33 $\pm$ 0.12} \\

\bottomrule
\end{tabular}
}

\end{table*}

\begin{table*}[t]
    \caption{Average Training Time (in seconds) for $h=2$. For each row, the best performer is in bold, and the best performer runner-up is underlined.}
    \label{tab:average_time}
    \centering
    \scriptsize
    \begin{tabular}{cc>{\columncolor{gray!20}}cc>{\columncolor{gray!20}}cc>{\columncolor{gray!20}}cc>{\columncolor{gray!20}}cc>{\columncolor{gray!20}}c}
    \toprule
    {split} & \texttt{CLF} & \texttt{FPMC} & \texttt{HRM} & \texttt{IBP} & \texttt{MC} & \texttt{NMF} & \texttt{TBP} & \texttt{TXMT} & \texttt{XMT} \\
    \midrule
    30 & 85.0 $\pm$ 12.5 & 2.4 $\pm$ 0.3 & 1595.3 $\pm$ 241.5 & \textbf{1.3 $\pm$ 0.1} & \underline{1.5 $\pm$ 0.2} & 29.1 $\pm$ 26.9 & 4.3 $\pm$ 4.1 & 1524.2 $\pm$ 932.7 & 4.8 $\pm$ 0.3 \\
    50 & 147.9 $\pm$ 14.6 & 2.9 $\pm$ 0.6 & 2208.5 $\pm$ 203.3 & \textbf{1.5 $\pm$ 0.1} & \underline{2.4 $\pm$ 0.8} & 42.0 $\pm$ 22.4 & 5.6 $\pm$ 5.8 & 2511.8 $\pm$ 1447.5 & 5.9 $\pm$ 0.4 \\
    70 & 188.8 $\pm$ 22.4 & 3.4 $\pm$ 0.5 & 2434.3 $\pm$ 224.0 & \textbf{1.8 $\pm$ 0.3} & \underline{3.1 $\pm$ 0.3} & 47.5 $\pm$ 15.9 & 7.6 $\pm$ 8.3 & 3486.6 $\pm$ 1851.3 & 6.7 $\pm$ 0.5 \\
    \bottomrule
    \end{tabular}
\end{table*}

\begin{table*}[!t]
    \caption{Example of explanations for the forgotten item \textit{vegetables}.}
    \label{tab:xmt_txmt_explanation}
    \centering
    \scriptsize
    \begin{tabular}{llll}
    \toprule
    \multicolumn{4}{l}{\textbf{Current Basket} $B_t = \{\mathit{rice},
    \mathit{snacks},
    \mathit{prepared}\; \mathit{food},
    \mathit{bread},  
    \mathit{tomatoes},
    \mathit{hazelnut~} \mathit{fruit},
    \mathit{wine},
    \mathit{olive}\; \mathit{oil},
    \mathit{cold}\; \mathit{cuts}\}$} \\
    \multicolumn{4}{l}{\textbf{Forgotten Items} $\hat{F}_t = \{ \mathit{drinks}, \mathit{dried}\; \mathit{fruit}, \mathit{rabbit}, \mathit{yogurt}, \mathit{vegetables}\}$} \\
    \multicolumn{4}{l}{\textbf{Forgotten Item Analyzed:} \textit{vegetables}} \\
    \midrule
    \multirow{3}{*}{\xmt} 
      & Last purchased & 3 days ago & Typically bought every 2.0 days \\
      & Often bought with & bread, cold cuts, wine & Based on co-occurrence \\
      & Often repurchased soon & 71.3\% & Of past opportunities \\
    \midrule
    \multirow{1}{*}{\txmt} 
      %& Last purchased & 3 days ago & Typically bought every 2.0 days \\
      %& Often bought with & bread, cold cuts, wine & Based on co-occurrence \\
      %& Often repurchased soon & 71.3\% & Of past opportunities \\
      & TARS pattern likelihood & 50.5\% & Confidence in pattern match \\
    \bottomrule
    \end{tabular}
\end{table*}

\smallskip
In the experiments, we compare \xmt and \txmt, with eight solutions: four baseline methods introduced in~\cite{cumby} and four state-of-the-art approaches 
\texttt{LAST}~\cite{cumby}: predicts the same products contained in the customer's previous basket, 
\texttt{TOP}~\cite{cumby}: predicts the top most frequent items with respect to their appearance in the customer's history, 
\texttt{MC}~\cite{cumby}:  makes the prediction based on the last customer purchase and on a Markov chain calculated on the customer's history,
\texttt{CLF}~\cite{cumby}: uses a binary classifier based on temporal purchase history features to predict if an item will be in the next basket,
Non-negative Matrix Factorization (\texttt{NMF})~\cite{lee2000algorithms}: uses non-negative matrix factorization on the customer-item matrix, 
Factorizing personalized Markov Chain (\texttt{FPMC})~\cite{rendle2010factorizing}: combines personalized Markov chains with matrix factorization, 
Hierarchical Representation Model (\texttt{HRM})~\cite{wang2015learning}: combines personalized Markov chains with matrix factorization, and 
Interval Based Predictor (\texttt{IBP})~\cite{singh2020prediction}. %: estimates when a customer is likely to repurchase an item by considering the average time between customer's visits and the average time between purchases of that item.
We remind the reader that, among the various baselines, all, except for \texttt{IBP}, are designed for NBP rather than FIP.
Therefore, \texttt{IBP} serves as the only true competitor.

\subsection{Experimental Results}
\label{sec:exp_results}
This section reports results obtained using \xmt and \txmt, and all baselines and competitors, on the \coopdataset dataset, with $h=2$. 
This setting considers an item as forgotten if not purchased within two days of the previous expense. 
We evaluate performance using three train-test splits: \textit{30\% Train}, \textit{50\% Train}, and \textit{70\% Train} to analyze the training size impact on prediction performance. 
Due to space constraints, we report in Appendix C of the SM~\cite{forgotten2025arxiv} further results with similar findings, including those with other split percentage values and with $h=0$ and $h=1$.

\Cref{fig:coop_50_day_2} reports the \textit{Recall}, \textit{Precision} and \textit{F1-score} obtained by the various methods while varying the prediction lengths $k$ from 5 to 20 in the \textit{30\% Train}, \textit{50\% Train} and \textit{70\% Train} settings, respectively. 
All the metrics are averaged over the different customers on the test set with at least a forgotten-items basket. 
We can notice that \xmt and \txmt always outperform all the competitors. 
In particular, \texttt{TOP}, \texttt{TBP}, \texttt{NMF}, and \texttt{MC} demonstrate relatively strong performance, while  \texttt{FPMC}, \texttt{LAST}, and \texttt{HRM} perform poorly across all metrics, showing limited prediction power in solving the FIP problem. 
Notably, both \xmt and \txmt outperform \texttt{IBP}~\cite{singh2020prediction}, which is the state-of-the-art method for the FIP problem.
From these results, we observe that \textit{Recall} consistently increases, while \textit{Precision} decreases.
This behavior is expected: as prediction length $k$ increases, the likelihood of capturing more correct items rises, but so does the risk of including incorrect ones.

\Cref{tab:f1_score} presents a detailed summary of the F1-scores obtained by each method for a subset of prediction length $k$ equals to 5, 10, 15 and 20. 
As the table shows, \xmt and \txmt demonstrate a substantial performance increase, outperforming all other competitor methods by approximately 5\% to 85\% in terms of F1-score across the different settings of prediction lengths and data splits.
Notably, increasing the amount of training data does not consistently correspond to an improvement in predictive performance across all models. 
This lack of gain is particularly evident when examining~\Cref{tab:f1_score}, where all the models, including \txmt and \xmt, show stable F1-score as the training split increases from 30\% to 70\% for a given prediction length $k$. 
This finding is particularly relevant for the design of personalized recommender systems, as having a method that achieves strong performance with a relatively small amount of training data is crucial for allowing the training of user-specific models that preserve privacy without sacrificing utility. 
Moreover,~\Cref{tab:f1_score} reveals a trend of increasing performance measured with F1-score as the prediction length $k$ increases. 
In particular, for the proposed methods \xmt and \txmt, we can notice an increase in the F1 Score of 47\% when transitioning from $k=5$ to $k=20$, showing how leveraging a longer prediction length could lead to more accurate predictions. 

To compare \xmt with \txmt and consider their real-world application, we present the training times (in seconds) for the evaluated methods in~\Cref{tab:average_time}. 
We excluded the baseline methods \texttt{TOP} and \texttt{LAST}, which output the most frequent and most recent purchases, respectively.  
Regarding training time efficiency, \texttt{IBP} consistently outperforms all other methods. 
However, our proposed approach offers higher performance compared with \texttt{IBP}. 
This advantage is particularly noticeable when comparing \texttt{IBP} with \xmt, which achieves higher performance with comparable training duration.
Compared with \xmt, \txmt requires significantly more computational resources, with training times up to 912x longer. Incorporating TARS patterns in \txmt enables more nuanced and explainable predictions, potentially justifying the longer training times through enhanced customer experience and potential revenue benefits. 
Conversely, \xmt offers a competitive option for scenarios prioritizing rapid deployment, substantially reducing computation time for a minor trade-off in performance.

\subsection{Interpretability of \xmt and \txmt}
Both $\xmt$ and $\txmt$ provide interpretable justifications for predicted forgotten items, yet they differ in the type and richness of the information they offer. 
To illustrate this, we examine a FIP scenario with a prediction length of $k = 5$, in which the items \textit{drinks}, \textit{dried fruit}, \textit{rabbit}, \textit{yogurt}, and \textit{vegetables} were predicted as forgotten, i.e., $\hat{F}_r = \{ \mathit{drinks}, \mathit{dried} \mathit{fruit}, \mathit{rabbit}, \mathit{yogurt}, \mathit{vegetables}\}$ given the following current basket $B_t = \{\mathit{rice},
\mathit{snacks},
\mathit{prepared~} \mathit{food},
\mathit{bread},$ $
\mathit{tomatoes},
\mathit{hazelnut~} \mathit{fruit},
\mathit{wine},
\mathit{olive} \mathit{oil},
\mathit{cold~} \mathit{cuts}\}$.

Table~\ref{tab:xmt_txmt_explanation} compares the explanations provided by $\xmt$ and $\txmt$ for the item \textit{vegetables}.
The explanation generated by $\xmt$ consists of three elements: the recency of the last purchase of \textit{vegetables} (3 days ago), its typical purchase frequency (every 2.0 days), and its co-occurrence with items such as \textit{bread, cold cuts,} and \textit{wine}. 
Additionally, $\xmt$ reports that \textit{vegetables} are frequently repurchased shortly after being bought, occurring in 71.3\% of relevant historical contexts. 
This explanation reflects aggregated behavioral patterns grounded in time and co-occurrence statistics.
As another example of co-occurrence useful in FIP, we can consider the \textit{dried fruit} item that is predicted as forgotten and a reason for suggesting it is the purchase of \textit{hazelnut fruit} at time $t$.
$\txmt$ complements the aforementioned justifications with an additional layer, i.e., the likelihood of \textit{vegetables} being needed again based on recurring temporal purchase patterns. 
Specifically, it reports a 50.5\% confidence that \textit{vegetables} will reoccur according to past behavior captured by mined sequences. 
This TARS-derived signal goes beyond surface-level frequency and highlights that the current context strongly aligns with a pattern where \textit{vegetables} have consistently followed previous purchases. 
Thus, TARS likelihood provides a temporally structured rationale that improves interpretability by linking forgotten item to recurring behavioral sequences instead of isolated co-occurrence or recency metrics alone. 
Additional item explanations are in Appendix C of the SM~\cite{forgotten2025arxiv}, while Appendix D shows a dashboard visualization for user explanations.

\section{Conclusion}
\label{sec:conclusion}
In this paper, we introduced \texttt{XMT} and \texttt{TXMT} for tackling the Forgotten Item Prediction (FIP) problem.
Evaluated on a complex real-world dataset, our models outperform nine existing methods, including those designed for both NBP and FIP tasks. Our user-centric approach lets individuals use their shopping history to train models that suggest forgotten items, ensuring highly personalized recommendations while preserving privacy, as no data is shared externally.
A key advantage of \texttt{XMT} and \texttt{TXMT} is their interpretability, as they provide clear explanations for why specific items are predicted as forgotten, enhancing user trust and offering actionable insights.
A limitation of our work is the use of a single dataset, due to scarce public data suited for FIP.
Future work includes incorporating collaborative signals from similar users', helpful in cold-start scenarios. To reduce the privacy risks, we plan to explore privacy-preserving methods such as Federated Learning~\cite{fl}, enabling decentralized model training without centralizing user data.

\begin{ack}
This work has been partially supported by the Italian Project Fondo Italiano per la Scienza FIS00001966 ``MIMOSA'', by the PRIN 2022 framework project ``PIANO'' (Personalized Interventions Against Online Toxicity) under CUP B53D23013290006, by the European Community Horizon~2020 programme under the funding schemes ERC-2018-ADG G.A. 834756 ``XAI'', by the European Commission under the NextGeneration EU programme – National Recovery and Resilience Plan (Piano Nazionale di Ripresa e Resilienza, PNRR) Project: ``SoBigData.it – Strengthening the Italian RI for Social Mining and Big Data Analytics'' – Prot. IR0000013 –  Av. n. 3264 del 28/12/2021, M4C2 - Investimento 1.3, Partenariato Esteso PE00000013 - ``FAIR'' - Future Artificial Intelligence Research'' - Spoke 1 ``Human-centered AI'', ``FINDHR'' that has received funding from the European Union's Horizon Europe research and innovation program under G.A. 101070212 and ``TANGO'' under G.A. 101120763. Views and opinions expressed are, however, those of the author(s) only and do not necessarily reflect those of the European Union or the European Health and Digital Executive Agency (HaDEA). Neither the European Union nor the granting authority can be held responsible for them.
\end{ack}

%%% Use this command to include your bibliography file.
%\clearpage
%\clearpage
\bibliography{mybibfile}

\clearpage

\appendix

\section{Pre-Processing}

We applied a series of filtering rules to identify valid customers for our experiments in our dataset. These rules must be satisfied before applying the splitting algorithm to each customer’s purchase history.

\begin{itemize}
\item \textbf{Minimum number of baskets per customer}: 10;
\item \textbf{Minimum basket size}: $|B| \geq 1$;
\item \textbf{Maximum basket size}: $|B| \leq \infty$;
\item \textbf{Minimum item occurrences}: $f_i \geq 1$, where $f_i$ represents the frequency of item $i$;
\end{itemize}
\vspace{-1.0em}

The splitting algorithm works in two phases. First, it analyzes the transaction history of each customer; those that do not satisfy the criteria are discarded. Second, it applies the split type, e.g., 30\% Train, to divide them into training and test sets.

After applying the splitting criteria, we retained 17,131 (99.99\%) customers from the \coopdataset datasets.

\section{Hyperparameters}
\label{sec:appendix:hyperparameters}

The current implementation of \xmt and \txmt relies on a series of hyperparameters whose values were chosen based on heuristic reasoning rather than systematic optimization. For the base frequency component, we set the minimum appearances threshold $\epsilon$ to 5, reasoning that fewer than five occurrences would provide insufficient data to establish reliable purchase patterns while being strict enough to filter out truly sporadic purchases.

In the temporal proximity component, we chose a temporal score boost $\alpha$ of 1.5 to provide a noticeable but not overwhelming increase in prediction likelihood. The maximum interval multiplier $\gamma$ was set to 3.0, operating under the assumption that gaps in purchasing more than three times the usual interval likely indicate a change in buying pattern rather than forgetfulness. The 90-day discontinuation threshold $G$ represents a conservative estimate of when an item should no longer be considered part of a customer's regular purchases.

For seasonal adjustments, we set the seasonality threshold $\upsilon$ to 0.4, considering that in a uniform distribution, each season would account for 25\% of purchases, thus, 40\% in a single season suggests meaningful seasonal preference. The seasonal boost factor $\beta$ of 1.5 matches the temporal boost to maintain consistency in how we weigh different types of temporal evidence.

The co-purchase component uses a co-occurrence threshold $\Upsilon$ of 5 and a boost parameter $\Theta$ of 0.2, chosen to ensure that co-purchase evidence contributes meaningfully to predictions without dominating other factors. Similarly, the repurchase pattern component parameters ($\nu = 2$ days, $\Lambda = 0.5$, $\Phi = 0.3$, $O = 5$) were selected to identify patterns that seem intuitively likely to indicate forgotten items.

While these parameter choices have yielded promising results in our experiments, we acknowledge that they represent educated guesses rather than optimized values. Moreover, using fixed parameters across all customers may not be optimal given the highly personal nature of shopping behaviors. Future work should explore both systematic hyperparameter tuning through techniques such as grid search or Bayesian optimization~\cite{nguyen2019bayesian}, and the potential for dynamic, user-specific parameters that adapt to individual shopping patterns and purchase frequencies. Such personalization could increase performance, particularly for customers with unusual shopping cadences or those whose behaviour varies significantly from the norm.

\section{Other Results}
\label{sec:appendix:other_results}

\begin{figure}[ht]
    \centering
    \includegraphics[width=\linewidth]{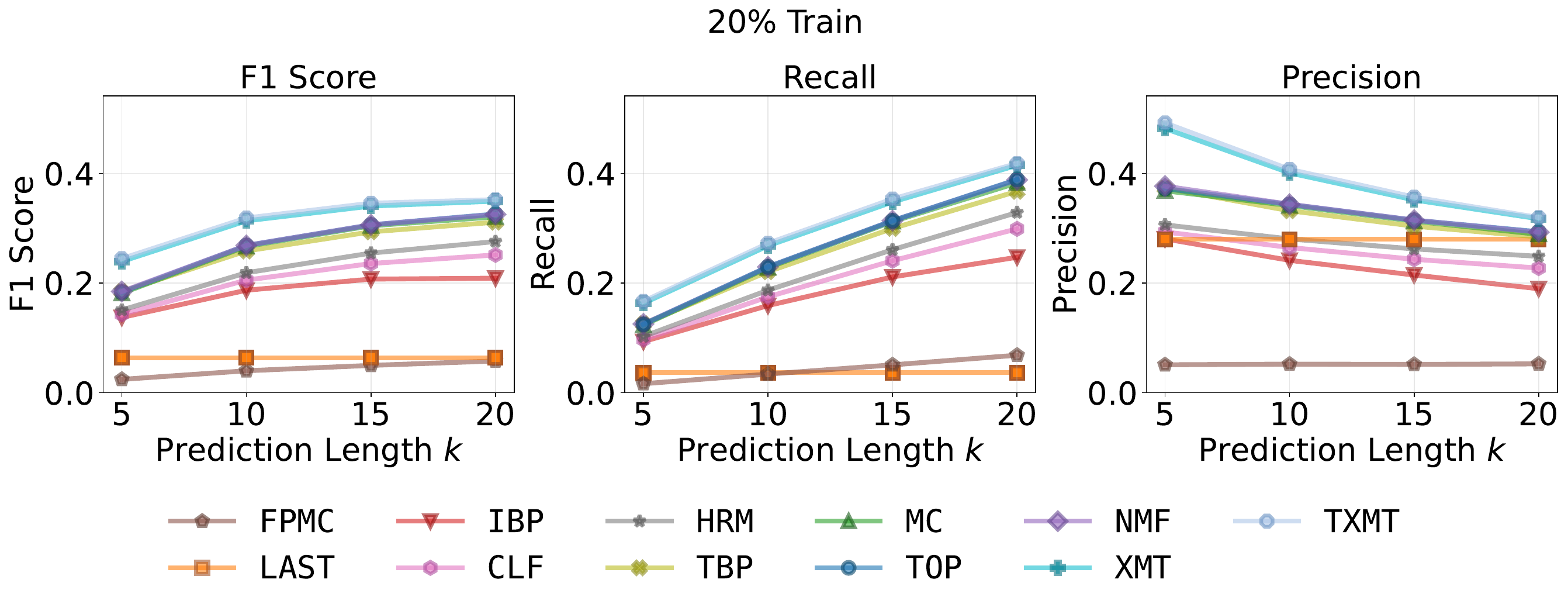}
    \caption{Comparison of \xmt and \txmt with Baseline methods using a \textit{20\% Train} and $h$=0}
    \label{fig:coop_20_day_0}
\end{figure}

\begin{figure}[ht]
    \centering
    \includegraphics[width=\linewidth]{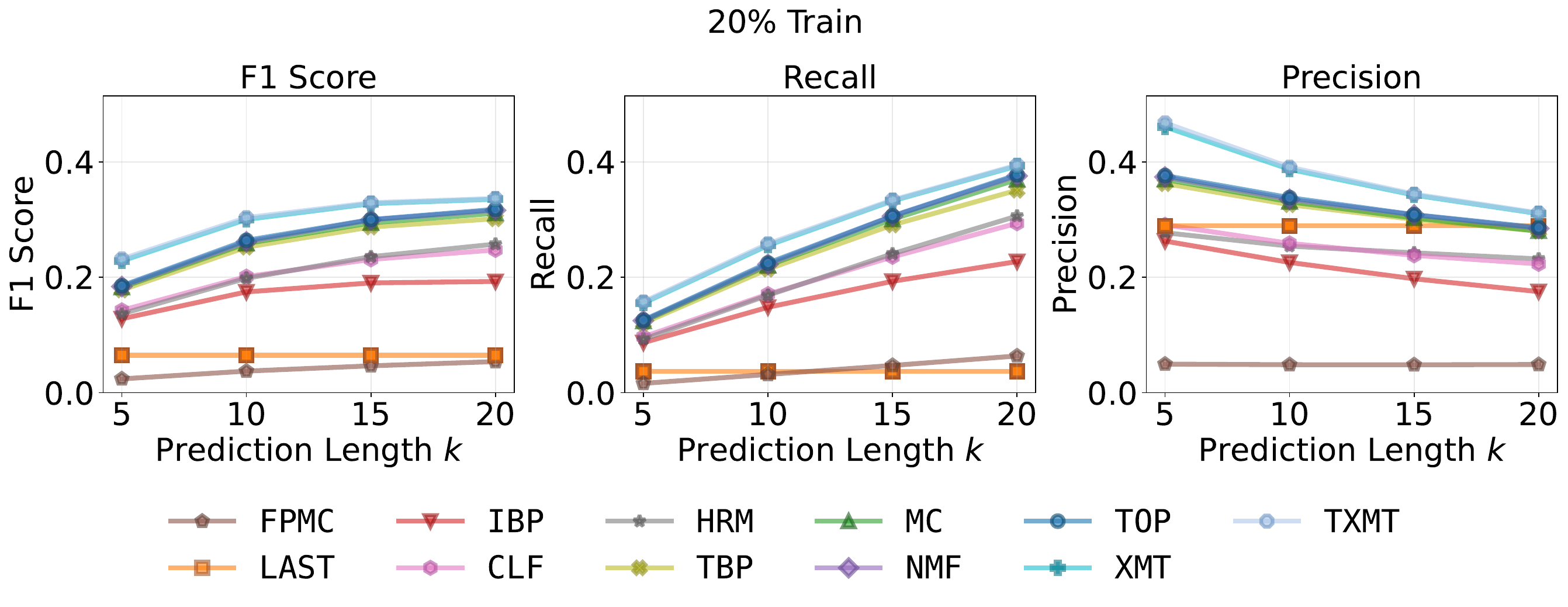}
    \caption{Comparison of \xmt and \txmt with Baseline methods using a \textit{20\% Train} and $h$=1}
    \label{fig:coop_20_day_1}
\end{figure}

\begin{figure}[ht]
    \centering
    \includegraphics[width=\linewidth]{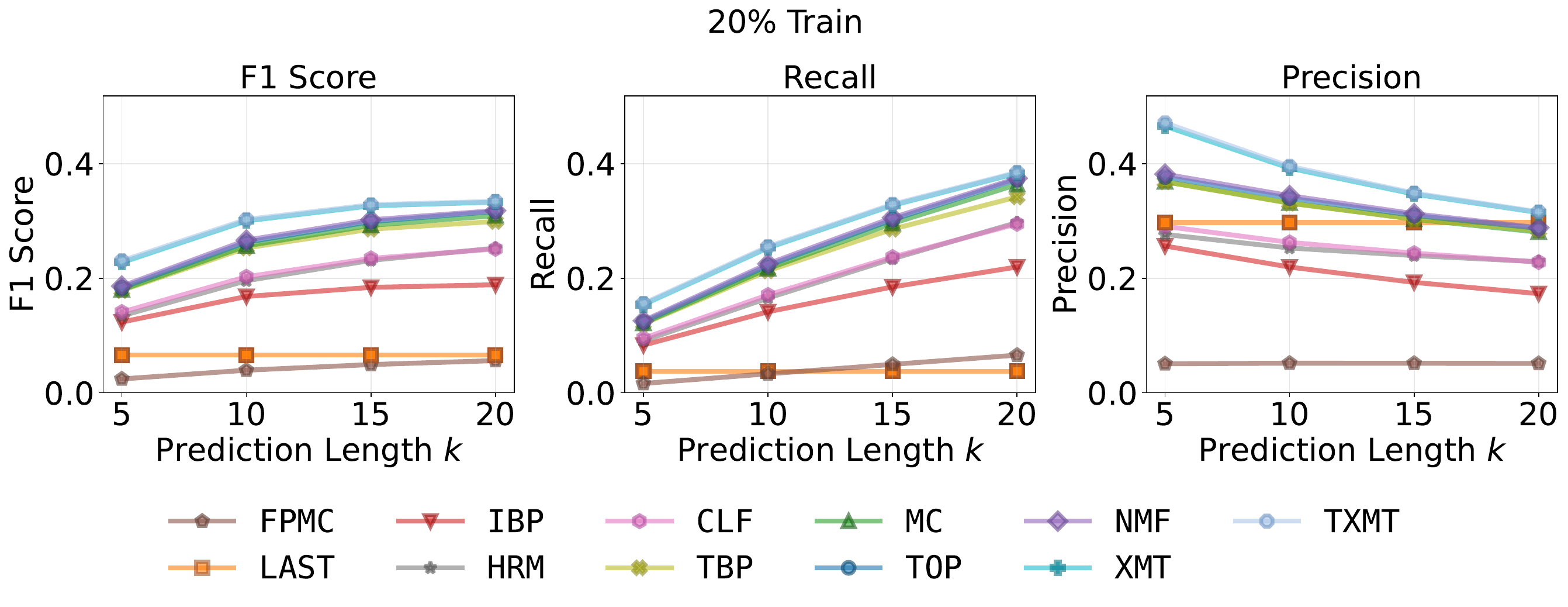}
    \caption{Comparison of \xmt and \txmt with Baseline methods using a \textit{20\% Train} and $h$=2}
    \label{fig:coop_20_day_2}
\end{figure}

In~\Cref{sec:exp_results} we reported the results obtained using the \coopdataset with $h = 2$ and \textit{30\% Train},\textit{50\% Train}, and \textit{70\% Train} i.e., using 30\%, 50\% and 70\% of the dataset for training.
In our experiments, we also tested other combinations of these parameters. In particular, we considered $h = 0$ and $h = 1$, i.e., the forgotten-item basket is purchased after a max of 0 or 1 day after the large basket. Moreover, we also considered different training dataset sizes ranging from 20\% to 80\%.
Overall, the results obtained using these settings are coherent with the ones presented in~\Cref{sec:exp_results}.
%20% SPLIT
Figures~\ref{fig:coop_20_day_0},~\ref{fig:coop_20_day_1} and~\ref{fig:coop_20_day_2} presents the result with \textit{20\% Train} and $h = 0$, $h = 1$ and $h = 2$ respectively. Tables~\ref{tab:comparison_precision_20},~\ref{tab:comparison_recall_20} and~\ref{tab:comparison_recall_20} report respectively the precision, recall and F1 Score with \textit{20\% Train}. Even in this scenario, in which we use a small part of the dataset to train our model, \xmt and \txmt offer performance comparable to the one we obtain when increasing the training dataset size. As highlighted in~\Cref{sec:exp_results}, this result is particularly important because it allows us to train models that offer good performance even with limited data, for example, when a new supermarket in the chain opens or when a new customer begins shopping and has no prior purchase history.

\begin{table*}[ht]
\caption{Precision comparison for different prediction lengths k and days $h$ with \textit{20\% Train}}
\label{tab:comparison_precision_20}
\vspace{0.5cm}
\centering
\small
\setlength{\tabcolsep}{4pt}
\resizebox{\linewidth}{!}{
\begin{tabular}{cc>{\columncolor{gray!20}}cc>{\columncolor{gray!20}}cc>{\columncolor{gray!20}}cc>{\columncolor{gray!20}}cc>{\columncolor{gray!20}}cc>{\columncolor{gray!20}}c}
\hline
\multicolumn{13}{c}{20\% Train} \\
\hline
\textbf{k} & \textbf{$h$} & \textbf{clf} & \textbf{fpmc} & \textbf{hrm} & \textbf{ibp} & \textbf{last} & \textbf{mc} & \textbf{nmf} & \textbf{tbp} & \textbf{top} & \textbf{txmt} & \textbf{xmt} \\
\hline
\multirow{3}{*}{5} &  0 & 0.29 $\pm$ 0.22 & 0.05 $\pm$ 0.10 & 0.31 $\pm$ 0.22 & 0.28 $\pm$ 0.22 & 0.28 $\pm$ 0.00 & 0.37 $\pm$ 0.23 & 0.38 $\pm$ 0.23 & 0.37 $\pm$ 0.23 & 0.37 $\pm$ 0.23 & \textbf{0.49 $\pm$ 0.26} & \underline{0.48 $\pm$ 0.25} \\
 & 1 & 0.29 $\pm$ 0.21 & 0.05 $\pm$ 0.10 & 0.28 $\pm$ 0.21 & 0.26 $\pm$ 0.22 & 0.29 $\pm$ 0.00 & 0.37 $\pm$ 0.23 & 0.37 $\pm$ 0.23 & 0.36 $\pm$ 0.23 & 0.38 $\pm$ 0.23 & \textbf{0.47 $\pm$ 0.25} & \underline{0.46 $\pm$ 0.25} \\
 & 2 & 0.29 $\pm$ 0.21 & 0.05 $\pm$ 0.10 & 0.28 $\pm$ 0.\multirow{3}{*}{20} & 0.26 $\pm$ 0.21 & 0.30 $\pm$ 0.00 & 0.37 $\pm$ 0.23 & \underline{0.38 $\pm$ 0.23} & 0.37 $\pm$ 0.23 & \underline{0.38 $\pm$ 0.23} & \textbf{0.47 $\pm$ 0.25} & \textbf{0.47 $\pm$ 0.25} \\
\midrule
\multirow{3}{*}{10} & 0 & 0.26 $\pm$ 0.16 & 0.05 $\pm$ 0.07 & 0.28 $\pm$ 0.15 & 0.24 $\pm$ 0.16 & 0.28 $\pm$ 0.00 & 0.34 $\pm$ 0.17 & 0.34 $\pm$ 0.17 & 0.33 $\pm$ 0.17 & 0.34 $\pm$ 0.17 & \textbf{0.41 $\pm$ 0.19} & \underline{0.40 $\pm$ 0.19} \\
 & 1 & 0.26 $\pm$ 0.15 & 0.05 $\pm$ 0.07 & 0.25 $\pm$ 0.14 & 0.23 $\pm$ 0.16 & 0.29 $\pm$ 0.00 & 0.33 $\pm$ 0.17 & 0.33 $\pm$ 0.17 & 0.33 $\pm$ 0.17 & \underline{0.34 $\pm$ 0.17} & \textbf{0.39 $\pm$ 0.19} & \textbf{0.39 $\pm$ 0.18} \\
 & 2 & 0.26 $\pm$ 0.15 & 0.05 $\pm$ 0.07 & 0.25 $\pm$ 0.14 & 0.22 $\pm$ 0.15 & 0.30 $\pm$ 0.00 & 0.33 $\pm$ 0.17 & 0.34 $\pm$ 0.17 & 0.33 $\pm$ 0.17 & 0.34 $\pm$ 0.17 & \textbf{0.40 $\pm$ 0.19} & \underline{0.39 $\pm$ 0.19} \\
\midrule
\multirow{3}{*}{15} & 0 & 0.24 $\pm$ 0.13 & 0.05 $\pm$ 0.06 & 0.26 $\pm$ 0.13 & 0.21 $\pm$ 0.13 & 0.28 $\pm$ 0.00 & 0.31 $\pm$ 0.14 & 0.31 $\pm$ 0.14 & 0.30 $\pm$ 0.14 & 0.31 $\pm$ 0.14 & \textbf{0.36 $\pm$ 0.16} & \underline{0.35 $\pm$ 0.16} \\
 & 1 & 0.24 $\pm$ 0.12 & 0.05 $\pm$ 0.06 & 0.24 $\pm$ 0.12 & 0.20 $\pm$ 0.13 & 0.29 $\pm$ 0.00 & 0.30 $\pm$ 0.14 & \underline{0.31 $\pm$ 0.14} & 0.30 $\pm$ 0.14 & \underline{0.31 $\pm$ 0.14} & \textbf{0.34 $\pm$ 0.16} & \textbf{0.34 $\pm$ 0.15} \\
 & 2 & 0.24 $\pm$ 0.13 & 0.05 $\pm$ 0.06 & 0.24 $\pm$ 0.12 & 0.19 $\pm$ 0.12 & 0.30 $\pm$ 0.00 & 0.30 $\pm$ 0.14 & \underline{0.31 $\pm$ 0.14} & 0.30 $\pm$ 0.14 & \underline{0.31 $\pm$ 0.14} & \textbf{0.35 $\pm$ 0.16} & \textbf{0.35 $\pm$ 0.16} \\
\midrule
\multirow{3}{*}{20} & 0 & 0.23 $\pm$ 0.11 & 0.05 $\pm$ 0.05 & 0.25 $\pm$ 0.11 & 0.19 $\pm$ 0.11 & 0.28 $\pm$ 0.00 & \underline{0.29 $\pm$ 0.12} & \underline{0.29 $\pm$ 0.12} & 0.28 $\pm$ 0.13 & \underline{0.29 $\pm$ 0.12} & \textbf{0.32 $\pm$ 0.14} & \textbf{0.32 $\pm$ 0.14} \\
 & 1 & 0.22 $\pm$ 0.11 & 0.05 $\pm$ 0.05 & 0.23 $\pm$ 0.11 & 0.18 $\pm$ 0.11 & \underline{0.29 $\pm$ 0.00} & 0.28 $\pm$ 0.12 & 0.28 $\pm$ 0.12 & 0.28 $\pm$ 0.13 & \underline{0.29 $\pm$ 0.12} & \textbf{0.31 $\pm$ 0.14} & \textbf{0.31 $\pm$ 0.14} \\
 & 2 & 0.23 $\pm$ 0.11 & 0.05 $\pm$ 0.05 & 0.23 $\pm$ 0.10 & 0.17 $\pm$ 0.11 & 0.30 $\pm$ 0.00 & 0.28 $\pm$ 0.12 & 0.29 $\pm$ 0.12 & 0.28 $\pm$ 0.13 & 0.29 $\pm$ 0.12 & \textbf{0.32 $\pm$ 0.14} & \underline{0.31 $\pm$ 0.14} \\
\midrule
\end{tabular}}

\end{table*}

\begin{table*}[ht]
\caption{Recall comparison for different prediction lengths (k) and days (d) with \textit{20\% Train}}
\label{tab:comparison_recall_20}
\vspace{0.5cm}
\centering
\small
\setlength{\tabcolsep}{4pt}
\resizebox{\linewidth}{!}{
\begin{tabular}{cc>{\columncolor{gray!20}}cc>{\columncolor{gray!20}}cc>{\columncolor{gray!20}}cc>{\columncolor{gray!20}}cc>{\columncolor{gray!20}}cc>{\columncolor{gray!20}}c}
\hline
\multicolumn{13}{c}{20\% Train} \\
\hline
\textbf{k} & \textbf{$h$} & \textbf{clf} & \textbf{fpmc} & \textbf{hrm} & \textbf{ibp} & \textbf{last} & \textbf{mc} & \textbf{nmf} & \textbf{tbp} & \textbf{top} & \textbf{txmt} & \textbf{xmt} \\
\hline
\multirow{3}{*}{5} & 0 & 0.10 $\pm$ 0.07 & 0.02 $\pm$ 0.03 & 0.10 $\pm$ 0.08 & 0.09 $\pm$ 0.07 & 0.04 $\pm$ 0.00 & 0.12 $\pm$ 0.08 & 0.13 $\pm$ 0.08 & 0.13 $\pm$ 0.08 & 0.12 $\pm$ 0.08 & \textbf{0.17 $\pm$ 0.09} & \underline{0.16 $\pm$ 0.09} \\
 & 1 & 0.10 $\pm$ 0.07 & 0.02 $\pm$ 0.03 & 0.09 $\pm$ 0.07 & 0.09 $\pm$ 0.07 & 0.04 $\pm$ 0.00 & 0.12 $\pm$ 0.08 & 0.12 $\pm$ 0.08 & 0.12 $\pm$ 0.08 & \underline{0.13 $\pm$ 0.08} & \textbf{0.16 $\pm$ 0.09} & \textbf{0.16 $\pm$ 0.09} \\
 & 2 & 0.09 $\pm$ 0.07 & 0.02 $\pm$ 0.03 & 0.09 $\pm$ 0.07 & 0.08 $\pm$ 0.07 & 0.04 $\pm$ 0.00 & 0.12 $\pm$ 0.08 & 0.13 $\pm$ 0.08 & 0.12 $\pm$ 0.08 & 0.12 $\pm$ 0.08 & \textbf{0.16 $\pm$ 0.09} & \underline{0.15 $\pm$ 0.09} \\
\midrule
\multirow{3}{*}{10} & 0 & 0.17 $\pm$ 0.10 & 0.03 $\pm$ 0.05 & 0.19 $\pm$ 0.10 & 0.16 $\pm$ 0.10 & 0.04 $\pm$ 0.00 & \underline{0.23 $\pm$ 0.11} & \underline{0.23 $\pm$ 0.11} & 0.22 $\pm$ 0.11 & \underline{0.23 $\pm$ 0.11} & \textbf{0.27 $\pm$ 0.12} & \textbf{0.27 $\pm$ 0.12} \\
 & 1 & 0.17 $\pm$ 0.10 & 0.03 $\pm$ 0.05 & 0.17 $\pm$ 0.09 & 0.15 $\pm$ 0.10 & 0.04 $\pm$ 0.00 & \underline{0.22 $\pm$ 0.11} & \underline{0.22 $\pm$ 0.10} & 0.21 $\pm$ 0.11 & \underline{0.22 $\pm$ 0.11} & \textbf{0.26 $\pm$ 0.12} & \textbf{0.26 $\pm$ 0.12} \\
 & 2 & 0.17 $\pm$ 0.09 & 0.03 $\pm$ 0.05 & 0.17 $\pm$ 0.09 & 0.14 $\pm$ 0.09 & 0.04 $\pm$ 0.00 & 0.22 $\pm$ 0.10 & 0.23 $\pm$ 0.11 & 0.21 $\pm$ 0.11 & 0.22 $\pm$ 0.11 & \textbf{0.26 $\pm$ 0.12} & \underline{0.25 $\pm$ 0.12} \\
\midrule
\multirow{3}{*}{15} & 0 & 0.24 $\pm$ 0.12 & 0.05 $\pm$ 0.06 & 0.26 $\pm$ 0.12 & 0.21 $\pm$ 0.12 & 0.04 $\pm$ 0.00 & \underline{0.31 $\pm$ 0.13} & \underline{0.31 $\pm$ 0.13} & 0.30 $\pm$ 0.13 & \underline{0.31 $\pm$ 0.13} & \textbf{0.35 $\pm$ 0.14} & \textbf{0.35 $\pm$ 0.14} \\
 & 1 & 0.24 $\pm$ 0.11 & 0.05 $\pm$ 0.06 & 0.24 $\pm$ 0.11 & 0.19 $\pm$ 0.11 & 0.04 $\pm$ 0.00 & 0.30 $\pm$ 0.12 & \underline{0.31 $\pm$ 0.12} & 0.29 $\pm$ 0.13 & \underline{0.31 $\pm$ 0.12} & \textbf{0.33 $\pm$ 0.14} & \textbf{0.33 $\pm$ 0.14} \\
 & 2 & 0.24 $\pm$ 0.11 & 0.05 $\pm$ 0.06 & 0.23 $\pm$ 0.10 & 0.18 $\pm$ 0.11 & 0.04 $\pm$ 0.00 & 0.30 $\pm$ 0.12 & \underline{0.31 $\pm$ 0.12} & 0.29 $\pm$ 0.12 & 0.30 $\pm$ 0.12 & \textbf{0.33 $\pm$ 0.14} & \textbf{0.33 $\pm$ 0.14} \\
\midrule
\multirow{3}{*}{20} & 0 & 0.30 $\pm$ 0.13 & 0.07 $\pm$ 0.06 & 0.33 $\pm$ 0.13 & 0.25 $\pm$ 0.13 & 0.04 $\pm$ 0.00 & 0.38 $\pm$ 0.14 & 0.39 $\pm$ 0.14 & 0.37 $\pm$ 0.14 & 0.39 $\pm$ 0.14 & \textbf{0.42 $\pm$ 0.15} & \underline{0.41 $\pm$ 0.15} \\
 & 1 & 0.29 $\pm$ 0.12 & 0.06 $\pm$ 0.06 & 0.31 $\pm$ 0.12 & 0.23 $\pm$ 0.12 & 0.04 $\pm$ 0.00 & 0.37 $\pm$ 0.13 & \underline{0.38 $\pm$ 0.13} & 0.35 $\pm$ 0.14 & \underline{0.38 $\pm$ 0.13} & \textbf{0.39 $\pm$ 0.15} & \textbf{0.39 $\pm$ 0.15} \\
 & 2 & 0.29 $\pm$ 0.12 & 0.07 $\pm$ 0.06 & 0.30 $\pm$ 0.12 & 0.22 $\pm$ 0.12 & 0.04 $\pm$ 0.00 & 0.36 $\pm$ 0.13 & \underline{0.37 $\pm$ 0.13} & 0.34 $\pm$ 0.14 & \underline{0.37 $\pm$ 0.13} & \textbf{0.38 $\pm$ 0.15} & \textbf{0.38 $\pm$ 0.15} \\
\midrule
\end{tabular}
}

\end{table*}

\begin{table*}[ht]
\caption{F1 Score comparison for different prediction lengths (k) and days (d)  with \textit{20\% Train}}
\label{tab:comparison_f1_score_20}
\vspace{0.5cm}
\centering
\small
\setlength{\tabcolsep}{4pt}
\resizebox{\linewidth}{!}{
\begin{tabular}{cc>{\columncolor{gray!20}}cc>{\columncolor{gray!20}}cc>{\columncolor{gray!20}}cc>{\columncolor{gray!20}}cc>{\columncolor{gray!20}}cc>{\columncolor{gray!20}}c}
\hline
\multicolumn{13}{c}{20\% Train} \\
\hline
\textbf{k} & \textbf{$h$} & \textbf{clf} & \textbf{fpmc} & \textbf{hrm} & \textbf{ibp} & \textbf{last} & \textbf{mc} & \textbf{nmf} & \textbf{tbp} & \textbf{top} & \textbf{txmt} & \textbf{xmt} \\
\hline
\multirow{3}{*}{5} & 0 & 0.14 $\pm$ 0.10 & 0.02 $\pm$ 0.05 & 0.15 $\pm$ 0.11 & 0.14 $\pm$ 0.11 & 0.06 $\pm$ 0.00 & \underline{0.18 $\pm$ 0.11} & \underline{0.18 $\pm$ 0.11} & \underline{0.18 $\pm$ 0.11} & \underline{0.18 $\pm$ 0.11} & \textbf{0.24 $\pm$ 0.13} & \textbf{0.24 $\pm$ 0.13} \\
 & 1 & 0.14 $\pm$ 0.10 & 0.02 $\pm$ 0.05 & 0.14 $\pm$ 0.10 & 0.13 $\pm$ 0.10 & 0.06 $\pm$ 0.00 & 0.18 $\pm$ 0.11 & 0.18 $\pm$ 0.11 & 0.18 $\pm$ 0.11 & \underline{0.19 $\pm$ 0.11} & \textbf{0.23 $\pm$ 0.12} & \textbf{0.23 $\pm$ 0.12} \\
 & 2 & 0.14 $\pm$ 0.10 & 0.02 $\pm$ 0.05 & 0.13 $\pm$ 0.10 & 0.12 $\pm$ 0.10 & 0.07 $\pm$ 0.00 & 0.18 $\pm$ 0.11 & \underline{0.19 $\pm$ 0.11} & 0.18 $\pm$ 0.11 & 0.18 $\pm$ 0.11 & \textbf{0.23 $\pm$ 0.12} & \textbf{0.23 $\pm$ 0.12} \\
\midrule
\multirow{3}{*}{10} & 0 & 0.21 $\pm$ 0.11 & 0.04 $\pm$ 0.05 & 0.22 $\pm$ 0.11 & 0.19 $\pm$ 0.11 & 0.06 $\pm$ 0.00 & 0.27 $\pm$ 0.12 & 0.27 $\pm$ 0.12 & 0.26 $\pm$ 0.12 & 0.27 $\pm$ 0.12 & \textbf{0.32 $\pm$ 0.14} & \underline{0.31 $\pm$ 0.13} \\
 & 1 & 0.20 $\pm$ 0.11 & 0.04 $\pm$ 0.05 & 0.20 $\pm$ 0.11 & 0.17 $\pm$ 0.11 & 0.06 $\pm$ 0.00 & \underline{0.26 $\pm$ 0.12} & \underline{0.26 $\pm$ 0.12} & 0.25 $\pm$ 0.12 & \underline{0.26 $\pm$ 0.12} & \textbf{0.30 $\pm$ 0.13} & \textbf{0.30 $\pm$ 0.13} \\
 & 2 & 0.20 $\pm$ 0.11 & 0.04 $\pm$ 0.05 & 0.20 $\pm$ 0.10 & 0.17 $\pm$ 0.11 & 0.07 $\pm$ 0.00 & 0.26 $\pm$ 0.12 & \underline{0.27 $\pm$ 0.12} & 0.25 $\pm$ 0.12 & 0.26 $\pm$ 0.12 & \textbf{0.30 $\pm$ 0.13} & \textbf{0.30 $\pm$ 0.13} \\
\midrule
\multirow{3}{*}{15} & 0 & 0.24 $\pm$ 0.11 & 0.05 $\pm$ 0.05 & 0.25 $\pm$ 0.11 & 0.21 $\pm$ 0.11 & 0.06 $\pm$ 0.00 & 0.30 $\pm$ 0.12 & 0.31 $\pm$ 0.12 & 0.29 $\pm$ 0.12 & 0.31 $\pm$ 0.12 & \textbf{0.35 $\pm$ 0.13} & \underline{0.34 $\pm$ 0.13} \\
 & 1 & 0.23 $\pm$ 0.11 & 0.05 $\pm$ 0.05 & 0.24 $\pm$ 0.10 & 0.19 $\pm$ 0.11 & 0.06 $\pm$ 0.00 & 0.29 $\pm$ 0.11 & \underline{0.30 $\pm$ 0.11} & 0.29 $\pm$ 0.12 & \underline{0.30 $\pm$ 0.11} & \textbf{0.33 $\pm$ 0.13} & \textbf{0.33 $\pm$ 0.13} \\
 & 2 & 0.23 $\pm$ 0.11 & 0.05 $\pm$ 0.05 & 0.23 $\pm$ 0.10 & 0.18 $\pm$ 0.11 & 0.07 $\pm$ 0.00 & 0.29 $\pm$ 0.12 & \underline{0.30 $\pm$ 0.12} & 0.29 $\pm$ 0.12 & \underline{0.30 $\pm$ 0.11} & \textbf{0.33 $\pm$ 0.13} & \textbf{0.33 $\pm$ 0.13} \\
\midrule
\multirow{3}{*}{20} & 0 & 0.25 $\pm$ 0.11 & 0.06 $\pm$ 0.05 & 0.28 $\pm$ 0.10 & 0.21 $\pm$ 0.11 & 0.06 $\pm$ 0.00 & 0.32 $\pm$ 0.11 & \underline{0.33 $\pm$ 0.11} & 0.31 $\pm$ 0.12 & \underline{0.33 $\pm$ 0.11} & \textbf{0.35 $\pm$ 0.12} & \textbf{0.35 $\pm$ 0.12} \\
 & 1 & 0.25 $\pm$ 0.10 & 0.05 $\pm$ 0.05 & 0.26 $\pm$ 0.10 & 0.19 $\pm$ 0.11 & 0.06 $\pm$ 0.00 & 0.31 $\pm$ 0.11 & \underline{0.32 $\pm$ 0.11} & 0.30 $\pm$ 0.11 & \underline{0.32 $\pm$ 0.11} & \textbf{0.34 $\pm$ 0.12} & \textbf{0.34 $\pm$ 0.12} \\
 & 2 & 0.25 $\pm$ 0.10 & 0.06 $\pm$ 0.05 & 0.25 $\pm$ 0.10 & 0.19 $\pm$ 0.10 & 0.07 $\pm$ 0.00 & 0.31 $\pm$ 0.11 & \underline{0.32 $\pm$ 0.11} & 0.30 $\pm$ 0.11 & \underline{0.32 $\pm$ 0.11} & \textbf{0.33 $\pm$ 0.12} & \textbf{0.33 $\pm$ 0.12} \\
\midrule
\end{tabular}
}

\end{table*}

%30% SPLIT

\begin{figure}[ht]
    \centering
    \includegraphics[width=\linewidth]{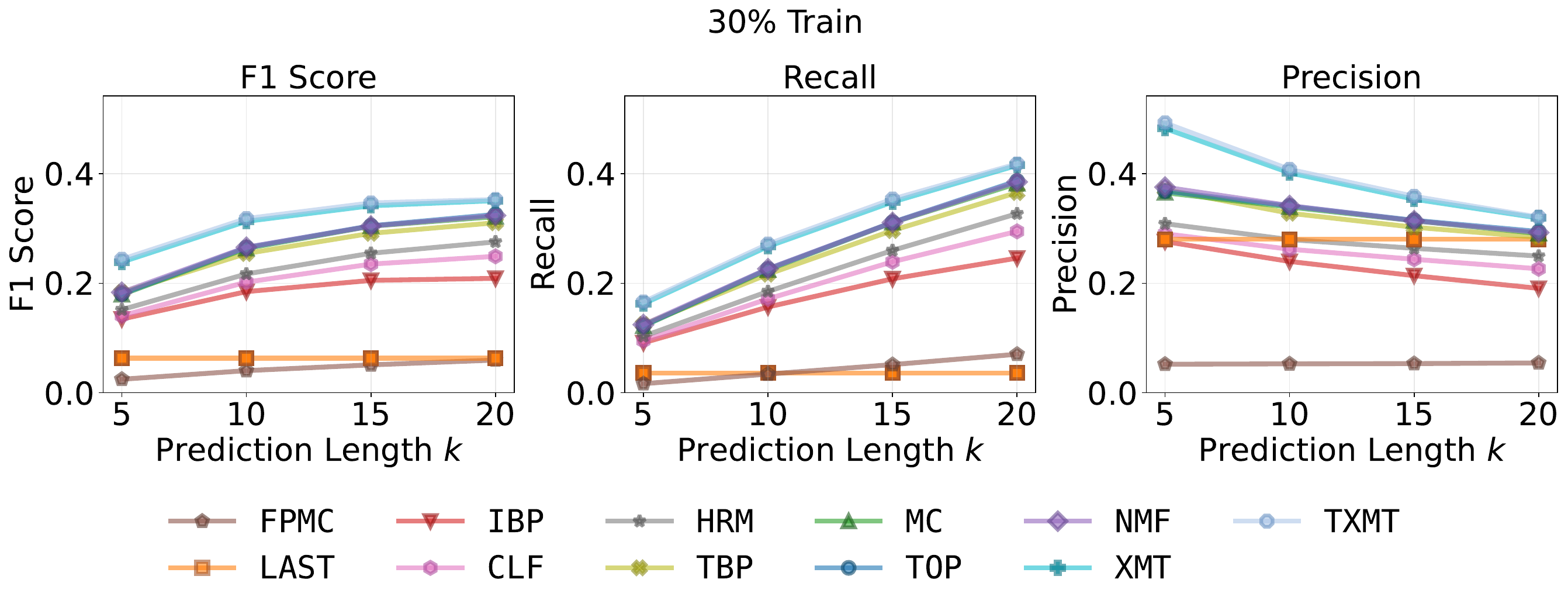}
    \caption{Comparison of \xmt and \txmt with Baseline methods using a \textit{30\% Train} and $h$=0}
    \label{fig:coop_30_day_0}
\end{figure}

\begin{figure}[ht]
    \centering
    \includegraphics[width=\linewidth]{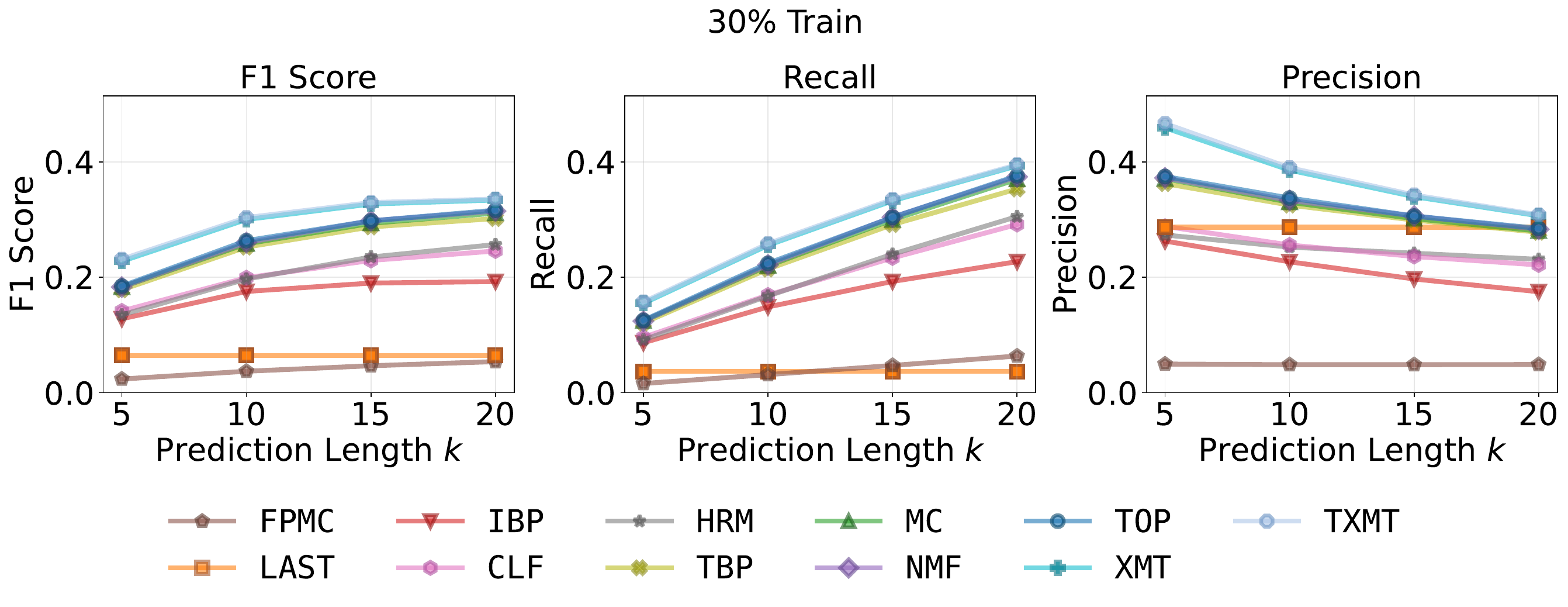}
    \caption{Comparison of \xmt and \txmt with Baseline methods using a \textit{30\% Train} and $h$=1}
    \label{fig:coop_30_day_1}
\end{figure}

Figures~\ref{fig:coop_30_day_0}, and~\ref{fig:coop_30_day_1} presents the result with \textit{30\% Train} and $h = 0$ and $h = 1$ respectively. Tables~\ref{tab:comparison_precision_30},~\ref{tab:comparison_recall_30} and~\ref{tab:comparison_recall_30} report respectively the precision, recall and F1 Score with \textit{30\% Train}. This is one of the settings we considered in~\Cref{sec:exp_results}, here we offer a more in-depth view of the results, exploring how they change when we decrease the $h$.
In particular, when $h$ = 0, we can see a small increase in the performance of both \xmt and \txmt. This is probably because using a low value for the $h$ we better capture the forgotten baskets, and our methods can provide better results.


\begin{table*}[ht]
    \caption{Precision comparison for different prediction lengths k and days $h$ with \textit{30\% Train}}
    \label{tab:comparison_precision_30}
    \vspace{0.5cm}
    \centering
    \small
    \setlength{\tabcolsep}{4pt}
    \resizebox{\linewidth}{!}{
\begin{tabular}{cc>{\columncolor{gray!20}}cc>{\columncolor{gray!20}}cc>{\columncolor{gray!20}}cc>{\columncolor{gray!20}}cc>{\columncolor{gray!20}}cc>{\columncolor{gray!20}}c}
    \hline
    \multicolumn{13}{c}{30\% Train} \\
    \hline
    \textbf{k} & \textbf{$h$} & \textbf{clf} & \textbf{fpmc} & \textbf{hrm} & \textbf{ibp} & \textbf{last} & \textbf{mc} & \textbf{nmf} & \textbf{tbp} & \textbf{top} & \textbf{txmt} & \textbf{xmt} \\
    \hline
    \multirow{3}{*}{5} & 0 & 0.29 $\pm$ 0.22 & 0.05 $\pm$ 0.10 & 0.31 $\pm$ 0.22 & 0.28 $\pm$ 0.22 & 0.28 $\pm$ 0.00 & 0.37 $\pm$ 0.22 & 0.38 $\pm$ 0.22 & 0.37 $\pm$ 0.23 & 0.37 $\pm$ 0.23 & \textbf{0.49 $\pm$ 0.26} & \underline{0.48 $\pm$ 0.25} \\
     & 1 & 0.29 $\pm$ 0.21 & 0.05 $\pm$ 0.10 & 0.27 $\pm$ 0.\multirow{3}{*}{20} & 0.26 $\pm$ 0.22 & 0.29 $\pm$ 0.00 & 0.37 $\pm$ 0.23 & 0.37 $\pm$ 0.23 & 0.36 $\pm$ 0.23 & 0.37 $\pm$ 0.23 & \textbf{0.47 $\pm$ 0.25} & \underline{0.46 $\pm$ 0.24} \\
     & 2 & 0.29 $\pm$ 0.21 & 0.05 $\pm$ 0.10 & 0.27 $\pm$ 0.\multirow{3}{*}{20} & 0.26 $\pm$ 0.21 & 0.30 $\pm$ 0.00 & 0.37 $\pm$ 0.23 & \underline{0.38 $\pm$ 0.23} & 0.37 $\pm$ 0.23 & \underline{0.38 $\pm$ 0.23} & \textbf{0.47 $\pm$ 0.25} & \textbf{0.47 $\pm$ 0.25} \\
    \midrule
    \multirow{3}{*}{10} & 0 & 0.26 $\pm$ 0.16 & 0.05 $\pm$ 0.07 & 0.28 $\pm$ 0.15 & 0.24 $\pm$ 0.16 & 0.28 $\pm$ 0.00 & 0.34 $\pm$ 0.17 & 0.34 $\pm$ 0.17 & 0.33 $\pm$ 0.17 & 0.34 $\pm$ 0.17 & \textbf{0.41 $\pm$ 0.19} & \underline{0.40 $\pm$ 0.19} \\
     & 1 & 0.26 $\pm$ 0.15 & 0.05 $\pm$ 0.07 & 0.25 $\pm$ 0.14 & 0.23 $\pm$ 0.16 & 0.29 $\pm$ 0.00 & 0.33 $\pm$ 0.17 & 0.33 $\pm$ 0.17 & 0.33 $\pm$ 0.17 & \underline{0.34 $\pm$ 0.17} & \textbf{0.39 $\pm$ 0.18} & \textbf{0.39 $\pm$ 0.18} \\
     & 2 & 0.26 $\pm$ 0.15 & 0.05 $\pm$ 0.07 & 0.25 $\pm$ 0.14 & 0.22 $\pm$ 0.15 & 0.30 $\pm$ 0.00 & 0.33 $\pm$ 0.17 & 0.34 $\pm$ 0.17 & 0.33 $\pm$ 0.17 & 0.34 $\pm$ 0.17 & \textbf{0.40 $\pm$ 0.18} & \underline{0.39 $\pm$ 0.18} \\
    \midrule
    \multirow{3}{*}{15} & 0 & 0.24 $\pm$ 0.13 & 0.05 $\pm$ 0.06 & 0.26 $\pm$ 0.13 & 0.21 $\pm$ 0.13 & 0.28 $\pm$ 0.00 & 0.31 $\pm$ 0.14 & 0.31 $\pm$ 0.14 & 0.30 $\pm$ 0.14 & 0.31 $\pm$ 0.14 & \textbf{0.36 $\pm$ 0.16} & \underline{0.35 $\pm$ 0.16} \\
     & 1 & 0.24 $\pm$ 0.12 & 0.05 $\pm$ 0.06 & 0.24 $\pm$ 0.12 & 0.20 $\pm$ 0.13 & 0.29 $\pm$ 0.00 & 0.30 $\pm$ 0.14 & \underline{0.31 $\pm$ 0.14} & 0.30 $\pm$ 0.14 & \underline{0.31 $\pm$ 0.14} & \textbf{0.34 $\pm$ 0.15} & \textbf{0.34 $\pm$ 0.15} \\
     & 2 & 0.24 $\pm$ 0.12 & 0.05 $\pm$ 0.06 & 0.24 $\pm$ 0.12 & 0.19 $\pm$ 0.12 & 0.30 $\pm$ 0.00 & 0.30 $\pm$ 0.14 & \underline{0.31 $\pm$ 0.14} & 0.30 $\pm$ 0.14 & \underline{0.31 $\pm$ 0.14} & \textbf{0.35 $\pm$ 0.15} & \textbf{0.35 $\pm$ 0.15} \\
    \midrule
    \multirow{3}{*}{20} & 0 & 0.23 $\pm$ 0.11 & 0.05 $\pm$ 0.05 & 0.25 $\pm$ 0.11 & 0.19 $\pm$ 0.11 & 0.28 $\pm$ 0.00 & \underline{0.29 $\pm$ 0.12} & \underline{0.29 $\pm$ 0.12} & 0.28 $\pm$ 0.13 & \underline{0.29 $\pm$ 0.12} & \textbf{0.32 $\pm$ 0.13} & \textbf{0.32 $\pm$ 0.14} \\
     & 1 & 0.22 $\pm$ 0.11 & 0.05 $\pm$ 0.05 & 0.23 $\pm$ 0.10 & 0.17 $\pm$ 0.11 & \underline{0.29 $\pm$ 0.00} & 0.28 $\pm$ 0.12 & 0.28 $\pm$ 0.12 & 0.28 $\pm$ 0.13 & 0.28 $\pm$ 0.12 & \textbf{0.31 $\pm$ 0.13} & \textbf{0.31 $\pm$ 0.13} \\
     & 2 & 0.23 $\pm$ 0.11 & 0.05 $\pm$ 0.05 & 0.23 $\pm$ 0.10 & 0.17 $\pm$ 0.11 & \underline{0.30 $\pm$ 0.00} & 0.28 $\pm$ 0.12 & 0.29 $\pm$ 0.12 & 0.28 $\pm$ 0.13 & 0.29 $\pm$ 0.12 & \textbf{0.31 $\pm$ 0.14} & \textbf{0.31 $\pm$ 0.14} \\
    \midrule
    \end{tabular}}
    
    \end{table*}
\begin{table*}[ht]
\caption{Recall comparison for different prediction lengths (k) and days (d) with \textit{30\% Train}}
    \label{tab:comparison_recall_30}
\vspace{0.5cm}
    \centering
    \small
    \setlength{\tabcolsep}{4pt}
    \resizebox{\linewidth}{!}{
    \begin{tabular}{cc>{\columncolor{gray!20}}cc>{\columncolor{gray!20}}cc>{\columncolor{gray!20}}cc>{\columncolor{gray!20}}cc>{\columncolor{gray!20}}cc>{\columncolor{gray!20}}c}
    \hline
    \multicolumn{13}{c}{30\% Train} \\
    \hline
    \textbf{k} & \textbf{$h$} & \textbf{clf} & \textbf{fpmc} & \textbf{hrm} & \textbf{ibp} & \textbf{last} & \textbf{mc} & \textbf{nmf} & \textbf{tbp} & \textbf{top} & \textbf{txmt} & \textbf{xmt} \\
    \hline
    \multirow{3}{*}{5} & 0 & 0.10 $\pm$ 0.07 & 0.02 $\pm$ 0.03 & 0.10 $\pm$ 0.08 & 0.09 $\pm$ 0.07 & 0.04 $\pm$ 0.00 & 0.12 $\pm$ 0.08 & 0.12 $\pm$ 0.07 & 0.12 $\pm$ 0.08 & 0.12 $\pm$ 0.08 & \textbf{0.17 $\pm$ 0.09} & \underline{0.16 $\pm$ 0.09} \\
     & 1 & 0.10 $\pm$ 0.07 & 0.02 $\pm$ 0.03 & 0.09 $\pm$ 0.07 & 0.09 $\pm$ 0.07 & 0.04 $\pm$ 0.00 & 0.12 $\pm$ 0.08 & 0.12 $\pm$ 0.08 & 0.12 $\pm$ 0.08 & 0.13 $\pm$ 0.08 & \textbf{0.16 $\pm$ 0.09} & \underline{0.15 $\pm$ 0.09} \\
     & 2 & 0.09 $\pm$ 0.07 & 0.02 $\pm$ 0.03 & 0.09 $\pm$ 0.07 & 0.08 $\pm$ 0.07 & 0.04 $\pm$ 0.00 & 0.12 $\pm$ 0.08 & 0.13 $\pm$ 0.08 & 0.12 $\pm$ 0.08 & 0.12 $\pm$ 0.08 & \textbf{0.16 $\pm$ 0.09} & \underline{0.15 $\pm$ 0.09} \\
    \midrule
    \multirow{3}{*}{10} & 0 & 0.17 $\pm$ 0.10 & 0.03 $\pm$ 0.05 & 0.18 $\pm$ 0.10 & 0.16 $\pm$ 0.10 & 0.04 $\pm$ 0.00 & 0.22 $\pm$ 0.11 & \underline{0.23 $\pm$ 0.10} & 0.22 $\pm$ 0.11 & \underline{0.23 $\pm$ 0.11} & \textbf{0.27 $\pm$ 0.12} & \textbf{0.27 $\pm$ 0.12} \\
     & 1 & 0.17 $\pm$ 0.09 & 0.03 $\pm$ 0.05 & 0.17 $\pm$ 0.09 & 0.15 $\pm$ 0.10 & 0.04 $\pm$ 0.00 & \underline{0.22 $\pm$ 0.11} & \underline{0.22 $\pm$ 0.10} & 0.21 $\pm$ 0.10 & \underline{0.22 $\pm$ 0.11} & \textbf{0.26 $\pm$ 0.12} & \textbf{0.26 $\pm$ 0.12} \\
     & 2 & 0.17 $\pm$ 0.09 & 0.03 $\pm$ 0.05 & 0.16 $\pm$ 0.09 & 0.14 $\pm$ 0.09 & 0.04 $\pm$ 0.00 & 0.22 $\pm$ 0.10 & 0.22 $\pm$ 0.11 & 0.21 $\pm$ 0.10 & 0.22 $\pm$ 0.10 & \textbf{0.26 $\pm$ 0.12} & \underline{0.25 $\pm$ 0.12} \\
    \midrule
    \multirow{3}{*}{15} & 0 & 0.24 $\pm$ 0.11 & 0.05 $\pm$ 0.06 & 0.26 $\pm$ 0.11 & 0.21 $\pm$ 0.12 & 0.04 $\pm$ 0.00 & \underline{0.31 $\pm$ 0.13} & \underline{0.31 $\pm$ 0.12} & 0.30 $\pm$ 0.12 & \underline{0.31 $\pm$ 0.12} & \textbf{0.35 $\pm$ 0.14} & \textbf{0.35 $\pm$ 0.14} \\
     & 1 & 0.23 $\pm$ 0.11 & 0.05 $\pm$ 0.06 & 0.24 $\pm$ 0.11 & 0.19 $\pm$ 0.11 & 0.04 $\pm$ 0.00 & 0.30 $\pm$ 0.12 & 0.30 $\pm$ 0.12 & 0.29 $\pm$ 0.12 & 0.30 $\pm$ 0.12 & \textbf{0.34 $\pm$ 0.14} & \underline{0.33 $\pm$ 0.13} \\
     & 2 & 0.23 $\pm$ 0.11 & 0.05 $\pm$ 0.05 & 0.23 $\pm$ 0.10 & 0.18 $\pm$ 0.11 & 0.04 $\pm$ 0.00 & \underline{0.30 $\pm$ 0.12} & \underline{0.30 $\pm$ 0.12} & 0.29 $\pm$ 0.12 & \underline{0.30 $\pm$ 0.12} & \textbf{0.33 $\pm$ 0.13} & \textbf{0.33 $\pm$ 0.13} \\
    \midrule
    \multirow{3}{*}{20} & 0 & 0.29 $\pm$ 0.12 & 0.07 $\pm$ 0.07 & 0.33 $\pm$ 0.12 & 0.25 $\pm$ 0.13 & 0.04 $\pm$ 0.00 & 0.38 $\pm$ 0.13 & \underline{0.39 $\pm$ 0.13} & 0.37 $\pm$ 0.14 & \underline{0.39 $\pm$ 0.13} & \textbf{0.42 $\pm$ 0.15} & \textbf{0.42 $\pm$ 0.15} \\
     & 1 & 0.29 $\pm$ 0.12 & 0.06 $\pm$ 0.06 & 0.31 $\pm$ 0.12 & 0.23 $\pm$ 0.12 & 0.04 $\pm$ 0.00 & 0.37 $\pm$ 0.13 & 0.37 $\pm$ 0.13 & 0.35 $\pm$ 0.14 & \underline{0.38 $\pm$ 0.13} & \textbf{0.39 $\pm$ 0.15} & \textbf{0.39 $\pm$ 0.15} \\
     & 2 & 0.29 $\pm$ 0.12 & 0.06 $\pm$ 0.06 & 0.30 $\pm$ 0.12 & 0.22 $\pm$ 0.12 & 0.04 $\pm$ 0.00 & 0.36 $\pm$ 0.13 & \underline{0.37 $\pm$ 0.13} & 0.35 $\pm$ 0.14 & \underline{0.37 $\pm$ 0.13} & \textbf{0.39 $\pm$ 0.15} & \textbf{0.39 $\pm$ 0.15} \\
    \midrule
    \end{tabular}
    }
    
    \end{table*}
\begin{table*}[ht]
\caption{F1 Score comparison for different prediction lengths (k) and days (d)  with \textit{30\% Train}}
\label{tab:comparison_f1_score_30}
\vspace{0.5cm}
\centering
\small
\setlength{\tabcolsep}{4pt}
\resizebox{\linewidth}{!}{
\begin{tabular}{cc>{\columncolor{gray!20}}cc>{\columncolor{gray!20}}cc>{\columncolor{gray!20}}cc>{\columncolor{gray!20}}cc>{\columncolor{gray!20}}cc>{\columncolor{gray!20}}c}
\hline
\multicolumn{13}{c}{30\% Train} \\
\hline
\textbf{k} & \textbf{$h$} & \textbf{clf} & \textbf{fpmc} & \textbf{hrm} & \textbf{ibp} & \textbf{last} & \textbf{mc} & \textbf{nmf} & \textbf{tbp} & \textbf{top} & \textbf{txmt} & \textbf{xmt} \\
\hline
\multirow{3}{*}{5} & 0 & 0.14 $\pm$ 0.10 & 0.02 $\pm$ 0.05 & 0.15 $\pm$ 0.11 & 0.13 $\pm$ 0.10 & 0.06 $\pm$ 0.00 & \underline{0.18 $\pm$ 0.11} & \underline{0.18 $\pm$ 0.11} & \underline{0.18 $\pm$ 0.11} & \underline{0.18 $\pm$ 0.11} & \textbf{0.24 $\pm$ 0.13} & \textbf{0.24 $\pm$ 0.13} \\
 & 1 & 0.14 $\pm$ 0.10 & 0.02 $\pm$ 0.05 & 0.13 $\pm$ 0.10 & 0.13 $\pm$ 0.10 & 0.06 $\pm$ 0.00 & \underline{0.18 $\pm$ 0.11} & \underline{0.18 $\pm$ 0.11} & \underline{0.18 $\pm$ 0.11} & \underline{0.18 $\pm$ 0.11} & \textbf{0.23 $\pm$ 0.12} & \textbf{0.23 $\pm$ 0.12} \\
 & 2 & 0.14 $\pm$ 0.10 & 0.02 $\pm$ 0.05 & 0.13 $\pm$ 0.10 & 0.12 $\pm$ 0.10 & 0.07 $\pm$ 0.00 & 0.18 $\pm$ 0.11 & \underline{0.19 $\pm$ 0.11} & 0.18 $\pm$ 0.11 & 0.18 $\pm$ 0.11 & \textbf{0.23 $\pm$ 0.12} & \textbf{0.23 $\pm$ 0.12} \\
\midrule
\multirow{3}{*}{10} & 0 & 0.20 $\pm$ 0.11 & 0.04 $\pm$ 0.05 & 0.22 $\pm$ 0.11 & 0.18 $\pm$ 0.11 & 0.06 $\pm$ 0.00 & 0.26 $\pm$ 0.12 & 0.27 $\pm$ 0.12 & 0.25 $\pm$ 0.12 & 0.27 $\pm$ 0.12 & \textbf{0.32 $\pm$ 0.13} & \underline{0.31 $\pm$ 0.13} \\
 & 1 & 0.20 $\pm$ 0.11 & 0.04 $\pm$ 0.05 & 0.20 $\pm$ 0.10 & 0.18 $\pm$ 0.11 & 0.06 $\pm$ 0.00 & \underline{0.26 $\pm$ 0.12} & \underline{0.26 $\pm$ 0.12} & 0.25 $\pm$ 0.12 & \underline{0.26 $\pm$ 0.12} & \textbf{0.30 $\pm$ 0.13} & \textbf{0.30 $\pm$ 0.13} \\
 & 2 & 0.20 $\pm$ 0.11 & 0.04 $\pm$ 0.05 & 0.19 $\pm$ 0.10 & 0.17 $\pm$ 0.11 & 0.07 $\pm$ 0.00 & \underline{0.26 $\pm$ 0.12} & \underline{0.26 $\pm$ 0.12} & 0.25 $\pm$ 0.12 & \underline{0.26 $\pm$ 0.12} & \textbf{0.30 $\pm$ 0.13} & \textbf{0.30 $\pm$ 0.13} \\
\midrule
\multirow{3}{*}{15} & 0 & 0.24 $\pm$ 0.11 & 0.05 $\pm$ 0.05 & 0.25 $\pm$ 0.11 & 0.21 $\pm$ 0.11 & 0.06 $\pm$ 0.00 & 0.30 $\pm$ 0.12 & 0.30 $\pm$ 0.11 & 0.29 $\pm$ 0.12 & 0.31 $\pm$ 0.12 & \textbf{0.35 $\pm$ 0.13} & \underline{0.34 $\pm$ 0.13} \\
 & 1 & 0.23 $\pm$ 0.11 & 0.05 $\pm$ 0.05 & 0.24 $\pm$ 0.10 & 0.19 $\pm$ 0.11 & 0.06 $\pm$ 0.00 & 0.29 $\pm$ 0.11 & \underline{0.30 $\pm$ 0.11} & 0.29 $\pm$ 0.12 & \underline{0.30 $\pm$ 0.11} & \textbf{0.33 $\pm$ 0.13} & \textbf{0.33 $\pm$ 0.13} \\
 & 2 & 0.23 $\pm$ 0.11 & 0.05 $\pm$ 0.05 & 0.23 $\pm$ 0.10 & 0.18 $\pm$ 0.11 & 0.07 $\pm$ 0.00 & 0.29 $\pm$ 0.11 & \underline{0.30 $\pm$ 0.11} & 0.29 $\pm$ 0.12 & \underline{0.30 $\pm$ 0.11} & \textbf{0.33 $\pm$ 0.13} & \textbf{0.33 $\pm$ 0.13} \\
\midrule
\multirow{3}{*}{20} & 0 & 0.25 $\pm$ 0.10 & 0.06 $\pm$ 0.05 & 0.28 $\pm$ 0.10 & 0.21 $\pm$ 0.11 & 0.06 $\pm$ 0.00 & 0.32 $\pm$ 0.11 & 0.32 $\pm$ 0.11 & 0.31 $\pm$ 0.12 & \underline{0.33 $\pm$ 0.11} & \textbf{0.35 $\pm$ 0.12} & \textbf{0.35 $\pm$ 0.12} \\
 & 1 & 0.25 $\pm$ 0.10 & 0.05 $\pm$ 0.05 & 0.26 $\pm$ 0.10 & 0.19 $\pm$ 0.10 & 0.06 $\pm$ 0.00 & 0.31 $\pm$ 0.11 & 0.31 $\pm$ 0.11 & 0.30 $\pm$ 0.11 & 0.32 $\pm$ 0.11 & \textbf{0.34 $\pm$ 0.12} & \underline{0.33 $\pm$ 0.12} \\
 & 2 & 0.25 $\pm$ 0.10 & 0.05 $\pm$ 0.05 & 0.25 $\pm$ 0.10 & 0.19 $\pm$ 0.10 & 0.07 $\pm$ 0.00 & 0.31 $\pm$ 0.11 & 0.32 $\pm$ 0.11 & 0.30 $\pm$ 0.11 & 0.32 $\pm$ 0.11 & \textbf{0.34 $\pm$ 0.12} & \underline{0.33 $\pm$ 0.12} \\
\midrule
\end{tabular}}

\end{table*}

\begin{figure}[ht]
    \centering
    \includegraphics[width=\linewidth]{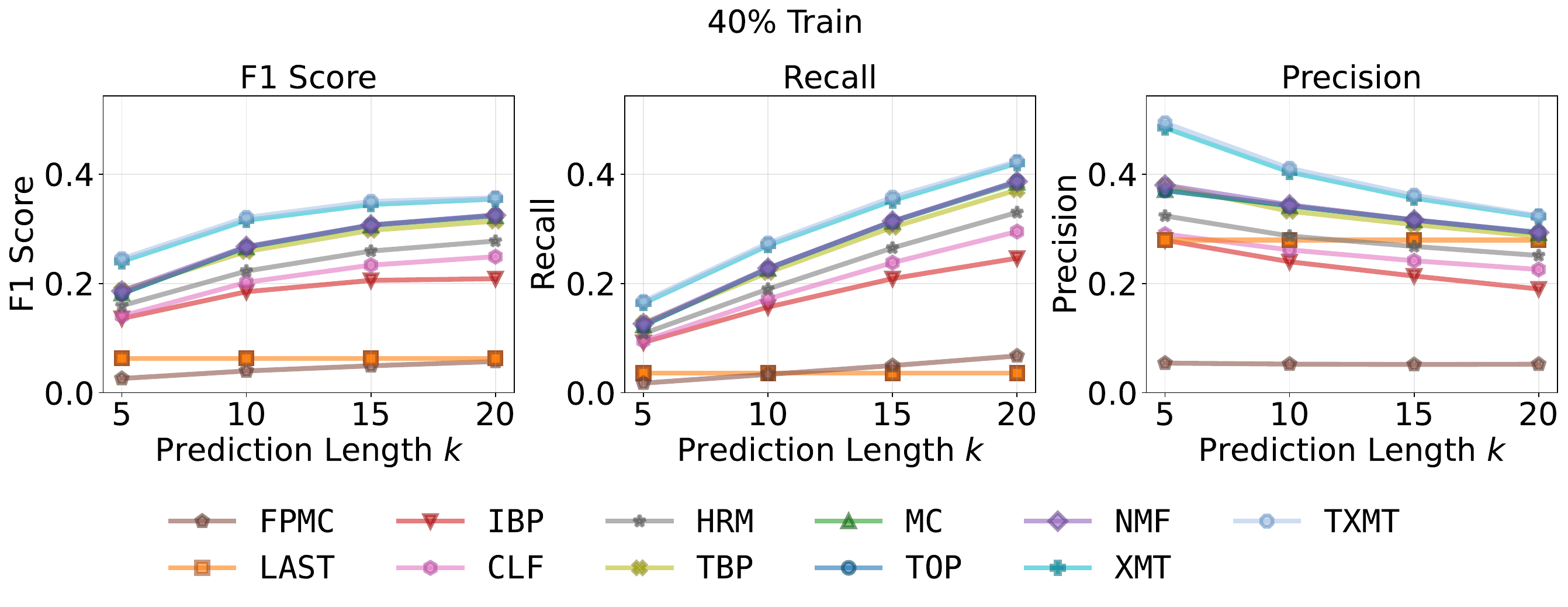}
    \caption{Comparison of \xmt and \txmt with Baseline methods using a \textit{40\% Train} and $h$=0}
    \label{fig:coop_40_day_0}
\end{figure}

\begin{figure}[ht]
    \centering
    \includegraphics[width=\linewidth]{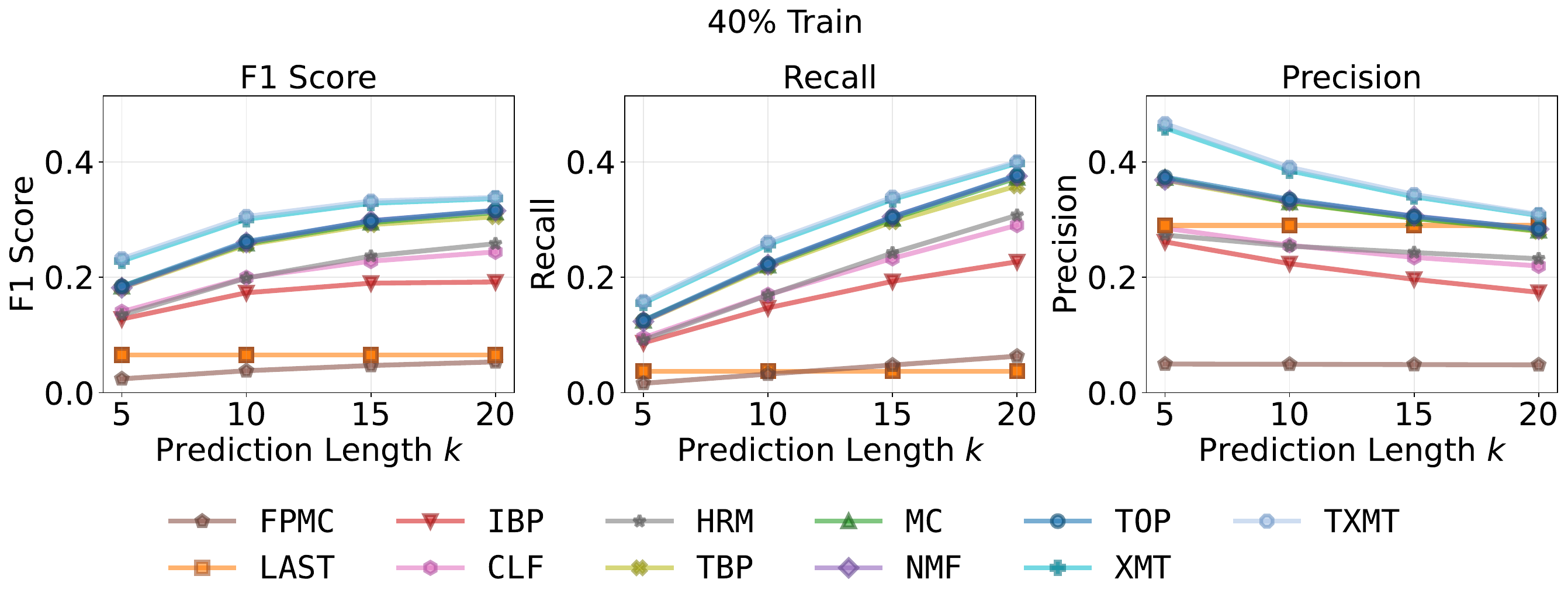}
    \caption{Comparison of \xmt and \txmt with Baseline methods using a \textit{40\% Train} and $h$=1}
    \label{fig:coop_40_day_1}
\end{figure}

\begin{figure}[ht]
    \centering
    \includegraphics[width=\linewidth]{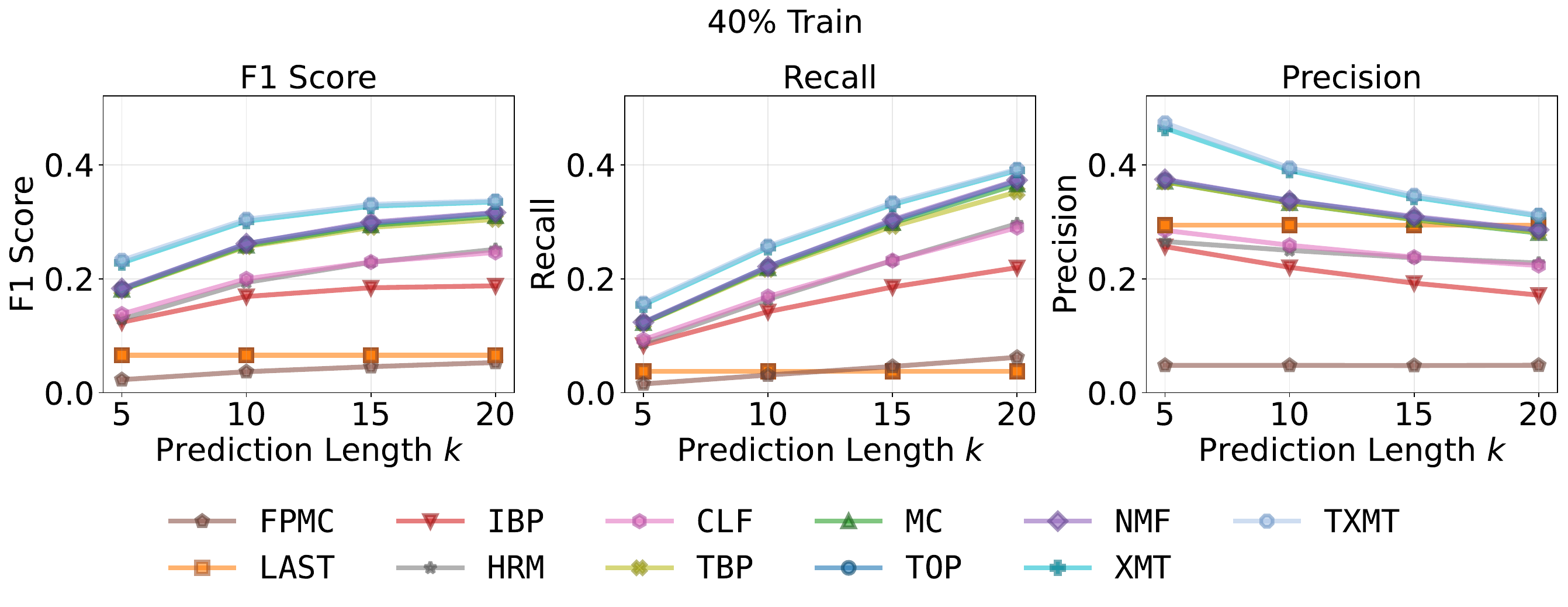}
    \caption{Comparison of \xmt and \txmt with Baseline methods using a \textit{40\% Train} and $h$=2}
    \label{fig:coop_40_day_2}
\end{figure}

%40% SPLIT
Figures~\ref{fig:coop_40_day_0}, ~\ref{fig:coop_40_day_1}and~\ref{fig:coop_40_day_2} presents the result with \textit{40\% Train} and $h = 0$, $h = 1$ and  $h = 2$ respectively. Tables~\ref{tab:comparison_precision_40},~\ref{tab:comparison_recall_40} and~\ref{tab:comparison_recall_40} report respectively the precision, recall and F1 Score with \textit{40\% Train}. Increasing the size of the training dataset does not have an impact on the results, which are similar to the ones shown in previous tables and plots. 


\begin{table*}[ht]
        \caption{Precision comparison for different prediction lengths k and days $h$ with \textit{40\% Train}}
        \label{tab:comparison_precision_40}
        \vspace{0.5cm}
        \centering
        \small
        \setlength{\tabcolsep}{4pt}
        \resizebox{\linewidth}{!}{
\begin{tabular}{cc>{\columncolor{gray!20}}cc>{\columncolor{gray!20}}cc>{\columncolor{gray!20}}cc>{\columncolor{gray!20}}cc>{\columncolor{gray!20}}cc>{\columncolor{gray!20}}c}
        \hline
        \multicolumn{13}{c}{40\% Train} \\
        \hline
        \textbf{k} & \textbf{$h$} & \textbf{clf} & \textbf{fpmc} & \textbf{hrm} & \textbf{ibp} & \textbf{last} & \textbf{mc} & \textbf{nmf} & \textbf{tbp} & \textbf{top} & \textbf{txmt} & \textbf{xmt} \\
        \hline
        \multirow{3}{*}{5} & 0 & 0.29 $\pm$ 0.22 & 0.05 $\pm$ 0.10 & 0.32 $\pm$ 0.22 & 0.28 $\pm$ 0.21 & 0.28 $\pm$ 0.00 & 0.37 $\pm$ 0.23 & 0.38 $\pm$ 0.23 & 0.38 $\pm$ 0.24 & 0.37 $\pm$ 0.23 & \textbf{0.49 $\pm$ 0.25} & \underline{0.48 $\pm$ 0.25} \\
         & 1 & 0.29 $\pm$ 0.21 & 0.05 $\pm$ 0.10 & 0.27 $\pm$ 0.\multirow{3}{*}{20} & 0.26 $\pm$ 0.21 & 0.29 $\pm$ 0.00 & 0.37 $\pm$ 0.23 & 0.37 $\pm$ 0.23 & 0.37 $\pm$ 0.23 & 0.37 $\pm$ 0.23 & \textbf{0.47 $\pm$ 0.24} & \underline{0.46 $\pm$ 0.24} \\
         & 2 & 0.29 $\pm$ 0.21 & 0.05 $\pm$ 0.10 & 0.27 $\pm$ 0.\multirow{3}{*}{20} & 0.26 $\pm$ 0.21 & 0.29 $\pm$ 0.00 & 0.37 $\pm$ 0.23 & 0.37 $\pm$ 0.23 & 0.37 $\pm$ 0.23 & 0.37 $\pm$ 0.23 & \textbf{0.47 $\pm$ 0.25} & \underline{0.46 $\pm$ 0.24} \\
        \midrule
        10 & 0 & 0.26 $\pm$ 0.16 & 0.05 $\pm$ 0.07 & 0.29 $\pm$ 0.16 & 0.24 $\pm$ 0.16 & 0.28 $\pm$ 0.00 & 0.34 $\pm$ 0.17 & 0.34 $\pm$ 0.17 & 0.33 $\pm$ 0.17 & 0.34 $\pm$ 0.17 & \textbf{0.41 $\pm$ 0.19} & \underline{0.40 $\pm$ 0.18} \\
         & 1 & 0.26 $\pm$ 0.15 & 0.05 $\pm$ 0.07 & 0.25 $\pm$ 0.14 & 0.22 $\pm$ 0.15 & 0.29 $\pm$ 0.00 & 0.33 $\pm$ 0.17 & 0.33 $\pm$ 0.17 & 0.33 $\pm$ 0.17 & 0.33 $\pm$ 0.17 & \textbf{0.39 $\pm$ 0.18} & \underline{0.38 $\pm$ 0.18} \\
         & 2 & 0.26 $\pm$ 0.15 & 0.05 $\pm$ 0.07 & 0.25 $\pm$ 0.14 & 0.22 $\pm$ 0.15 & 0.29 $\pm$ 0.00 & 0.33 $\pm$ 0.17 & \underline{0.34 $\pm$ 0.17} & 0.33 $\pm$ 0.17 & \underline{0.34 $\pm$ 0.17} & \textbf{0.39 $\pm$ 0.18} & \textbf{0.39 $\pm$ 0.18} \\
        \midrule
        15 & 0 & 0.24 $\pm$ 0.13 & 0.05 $\pm$ 0.06 & 0.27 $\pm$ 0.13 & 0.21 $\pm$ 0.13 & 0.28 $\pm$ 0.00 & \underline{0.32 $\pm$ 0.14} & \underline{0.32 $\pm$ 0.14} & 0.31 $\pm$ 0.14 & \underline{0.32 $\pm$ 0.14} & \textbf{0.36 $\pm$ 0.15} & \textbf{0.36 $\pm$ 0.15} \\
         & 1 & 0.23 $\pm$ 0.12 & 0.05 $\pm$ 0.06 & 0.24 $\pm$ 0.12 & 0.20 $\pm$ 0.13 & 0.29 $\pm$ 0.00 & 0.30 $\pm$ 0.14 & \underline{0.31 $\pm$ 0.14} & 0.30 $\pm$ 0.14 & \underline{0.31 $\pm$ 0.14} & \textbf{0.34 $\pm$ 0.15} & \textbf{0.34 $\pm$ 0.15} \\
         & 2 & 0.24 $\pm$ 0.12 & 0.05 $\pm$ 0.06 & 0.24 $\pm$ 0.12 & 0.19 $\pm$ 0.12 & 0.29 $\pm$ 0.00 & 0.30 $\pm$ 0.14 & 0.31 $\pm$ 0.14 & 0.30 $\pm$ 0.14 & 0.31 $\pm$ 0.13 & \textbf{0.35 $\pm$ 0.15} & \underline{0.34 $\pm$ 0.15} \\
        \midrule
        \multirow{3}{*}{20} & 0 & 0.23 $\pm$ 0.11 & 0.05 $\pm$ 0.05 & 0.25 $\pm$ 0.11 & 0.19 $\pm$ 0.11 & 0.28 $\pm$ 0.00 & \underline{0.29 $\pm$ 0.12} & \underline{0.29 $\pm$ 0.12} & \underline{0.29 $\pm$ 0.12} & \underline{0.29 $\pm$ 0.12} & \textbf{0.32 $\pm$ 0.13} & \textbf{0.32 $\pm$ 0.13} \\
         & 1 & 0.22 $\pm$ 0.11 & 0.05 $\pm$ 0.05 & 0.23 $\pm$ 0.10 & 0.17 $\pm$ 0.11 & \underline{0.29 $\pm$ 0.00} & 0.28 $\pm$ 0.12 & 0.28 $\pm$ 0.12 & 0.28 $\pm$ 0.13 & 0.28 $\pm$ 0.12 & \textbf{0.31 $\pm$ 0.13} & \textbf{0.31 $\pm$ 0.13} \\
         & 2 & 0.22 $\pm$ 0.11 & 0.05 $\pm$ 0.05 & 0.23 $\pm$ 0.10 & 0.17 $\pm$ 0.11 & \underline{0.29 $\pm$ 0.00} & 0.28 $\pm$ 0.12 & \underline{0.29 $\pm$ 0.12} & 0.28 $\pm$ 0.13 & \underline{0.29 $\pm$ 0.12} & \textbf{0.31 $\pm$ 0.13} & \textbf{0.31 $\pm$ 0.13} \\
        \midrule
        \end{tabular}}
        
        \end{table*}
\begin{table*}[ht]
    \caption{Recall comparison for different prediction lengths (k) and days (d) with \textit{40\% Train}}
        \label{tab:comparison_recall_40}
    \vspace{0.5cm}
        \centering
        \small
        \setlength{\tabcolsep}{4pt}
        \resizebox{\linewidth}{!}{
        \begin{tabular}{cc>{\columncolor{gray!20}}cc>{\columncolor{gray!20}}cc>{\columncolor{gray!20}}cc>{\columncolor{gray!20}}cc>{\columncolor{gray!20}}cc>{\columncolor{gray!20}}c}
        \hline
        \multicolumn{13}{c}{40\% Train} \\
        \hline
        \textbf{k} & \textbf{$h$} & \textbf{clf} & \textbf{fpmc} & \textbf{hrm} & \textbf{ibp} & \textbf{last} & \textbf{mc} & \textbf{nmf} & \textbf{tbp} & \textbf{top} & \textbf{txmt} & \textbf{xmt} \\
        \hline
        \multirow{3}{*}{5} & 0 & 0.09 $\pm$ 0.07 & 0.02 $\pm$ 0.03 & 0.11 $\pm$ 0.08 & 0.09 $\pm$ 0.07 & 0.04 $\pm$ 0.00 & 0.12 $\pm$ 0.08 & 0.13 $\pm$ 0.08 & 0.13 $\pm$ 0.08 & 0.12 $\pm$ 0.08 & \textbf{0.17 $\pm$ 0.09} & \underline{0.16 $\pm$ 0.09} \\
         & 1 & 0.10 $\pm$ 0.07 & 0.02 $\pm$ 0.03 & 0.09 $\pm$ 0.07 & 0.09 $\pm$ 0.07 & 0.04 $\pm$ 0.00 & 0.12 $\pm$ 0.08 & 0.12 $\pm$ 0.08 & 0.12 $\pm$ 0.08 & \underline{0.13 $\pm$ 0.08} & \textbf{0.16 $\pm$ 0.09} & \textbf{0.16 $\pm$ 0.09} \\
         & 2 & 0.09 $\pm$ 0.07 & 0.02 $\pm$ 0.03 & 0.09 $\pm$ 0.07 & 0.08 $\pm$ 0.07 & 0.04 $\pm$ 0.00 & 0.12 $\pm$ 0.08 & 0.12 $\pm$ 0.08 & 0.12 $\pm$ 0.08 & 0.12 $\pm$ 0.08 & \textbf{0.16 $\pm$ 0.09} & \underline{0.15 $\pm$ 0.08} \\
        \midrule
        \multirow{3}{*}{10} & 0 & 0.17 $\pm$ 0.10 & 0.03 $\pm$ 0.05 & 0.19 $\pm$ 0.10 & 0.16 $\pm$ 0.10 & 0.04 $\pm$ 0.00 & \underline{0.23 $\pm$ 0.10} & \underline{0.23 $\pm$ 0.10} & 0.22 $\pm$ 0.11 & \underline{0.23 $\pm$ 0.10} & \textbf{0.27 $\pm$ 0.12} & \textbf{0.27 $\pm$ 0.12} \\
         & 1 & 0.17 $\pm$ 0.10 & 0.03 $\pm$ 0.05 & 0.17 $\pm$ 0.09 & 0.15 $\pm$ 0.10 & 0.04 $\pm$ 0.00 & \underline{0.22 $\pm$ 0.11} & \underline{0.22 $\pm$ 0.10} & \underline{0.22 $\pm$ 0.11} & \underline{0.22 $\pm$ 0.11} & \textbf{0.26 $\pm$ 0.12} & \textbf{0.26 $\pm$ 0.12} \\
         & 2 & 0.17 $\pm$ 0.09 & 0.03 $\pm$ 0.04 & 0.16 $\pm$ 0.09 & 0.14 $\pm$ 0.09 & 0.04 $\pm$ 0.00 & 0.22 $\pm$ 0.10 & 0.22 $\pm$ 0.10 & 0.22 $\pm$ 0.11 & 0.22 $\pm$ 0.10 & \textbf{0.26 $\pm$ 0.12} & \underline{0.25 $\pm$ 0.11} \\
        \midrule
        \multirow{3}{*}{15} & 0 & 0.24 $\pm$ 0.11 & 0.05 $\pm$ 0.06 & 0.27 $\pm$ 0.11 & 0.21 $\pm$ 0.11 & 0.04 $\pm$ 0.00 & 0.31 $\pm$ 0.12 & 0.31 $\pm$ 0.12 & 0.30 $\pm$ 0.12 & 0.31 $\pm$ 0.12 & \textbf{0.36 $\pm$ 0.13} & \underline{0.35 $\pm$ 0.13} \\
         & 1 & 0.23 $\pm$ 0.11 & 0.05 $\pm$ 0.06 & 0.24 $\pm$ 0.11 & 0.19 $\pm$ 0.11 & 0.04 $\pm$ 0.00 & 0.30 $\pm$ 0.12 & 0.30 $\pm$ 0.12 & 0.30 $\pm$ 0.12 & 0.30 $\pm$ 0.12 & \textbf{0.34 $\pm$ 0.13} & \underline{0.33 $\pm$ 0.13} \\
         & 2 & 0.23 $\pm$ 0.11 & 0.05 $\pm$ 0.05 & 0.23 $\pm$ 0.11 & 0.19 $\pm$ 0.11 & 0.04 $\pm$ 0.00 & \underline{0.30 $\pm$ 0.12} & \underline{0.30 $\pm$ 0.12} & 0.29 $\pm$ 0.12 & \underline{0.30 $\pm$ 0.12} & \textbf{0.33 $\pm$ 0.13} & \textbf{0.33 $\pm$ 0.13} \\
        \midrule
        \multirow{3}{*}{20} & 0 & 0.30 $\pm$ 0.12 & 0.07 $\pm$ 0.07 & 0.33 $\pm$ 0.12 & 0.25 $\pm$ 0.12 & 0.04 $\pm$ 0.00 & 0.38 $\pm$ 0.13 & \underline{0.39 $\pm$ 0.13} & 0.37 $\pm$ 0.14 & \underline{0.39 $\pm$ 0.13} & \textbf{0.42 $\pm$ 0.15} & \textbf{0.42 $\pm$ 0.14} \\
         & 1 & 0.29 $\pm$ 0.12 & 0.06 $\pm$ 0.06 & 0.31 $\pm$ 0.12 & 0.23 $\pm$ 0.12 & 0.04 $\pm$ 0.00 & 0.37 $\pm$ 0.13 & \underline{0.38 $\pm$ 0.13} & 0.36 $\pm$ 0.14 & \underline{0.38 $\pm$ 0.13} & \textbf{0.40 $\pm$ 0.15} & \textbf{0.40 $\pm$ 0.15} \\
         & 2 & 0.29 $\pm$ 0.12 & 0.06 $\pm$ 0.06 & 0.30 $\pm$ 0.12 & 0.22 $\pm$ 0.12 & 0.04 $\pm$ 0.00 & \underline{0.37 $\pm$ 0.13} & \underline{0.37 $\pm$ 0.13} & 0.35 $\pm$ 0.14 & \underline{0.37 $\pm$ 0.13} & \textbf{0.39 $\pm$ 0.15} & \textbf{0.39 $\pm$ 0.15} \\
        \midrule
        \hline
        \end{tabular}
        }
        
        \end{table*} 
\begin{table*}[ht]
\caption{F1 Score comparison for different prediction lengths (k) and days (d)  with \textit{40\% Train}}
\label{tab:comparison_f1_score_40}
\vspace{0.5cm}
\centering
\small
\setlength{\tabcolsep}{4pt}
\resizebox{\linewidth}{!}{
\begin{tabular}{cc>{\columncolor{gray!20}}cc>{\columncolor{gray!20}}cc>{\columncolor{gray!20}}cc>{\columncolor{gray!20}}cc>{\columncolor{gray!20}}cc>{\columncolor{gray!20}}c}
\hline
\multicolumn{13}{c}{40\% Train} \\
\hline
\textbf{k} & \textbf{$h$} & \textbf{clf} & \textbf{fpmc} & \textbf{hrm} & \textbf{ibp} & \textbf{last} & \textbf{mc} & \textbf{nmf} & \textbf{tbp} & \textbf{top} & \textbf{txmt} & \textbf{xmt} \\
\hline
\multirow{3}{*}{5} & 0 & 0.14 $\pm$ 0.10 & 0.03 $\pm$ 0.05 & 0.16 $\pm$ 0.11 & 0.14 $\pm$ 0.10 & 0.06 $\pm$ 0.00 & 0.18 $\pm$ 0.11 & 0.19 $\pm$ 0.11 & 0.19 $\pm$ 0.12 & 0.18 $\pm$ 0.11 & \textbf{0.25 $\pm$ 0.13} & \underline{0.24 $\pm$ 0.12} \\
 & 1 & 0.14 $\pm$ 0.10 & 0.02 $\pm$ 0.05 & 0.13 $\pm$ 0.10 & 0.13 $\pm$ 0.10 & 0.07 $\pm$ 0.00 & \underline{0.18 $\pm$ 0.11} & \underline{0.18 $\pm$ 0.11} & \underline{0.18 $\pm$ 0.11} & \underline{0.18 $\pm$ 0.11} & \textbf{0.23 $\pm$ 0.12} & \textbf{0.23 $\pm$ 0.12} \\
 & 2 & 0.14 $\pm$ 0.10 & 0.02 $\pm$ 0.05 & 0.13 $\pm$ 0.10 & 0.12 $\pm$ 0.10 & 0.07 $\pm$ 0.00 & \underline{0.18 $\pm$ 0.11} & \underline{0.18 $\pm$ 0.11} & \underline{0.18 $\pm$ 0.11} & \underline{0.18 $\pm$ 0.11} & \textbf{0.23 $\pm$ 0.12} & \textbf{0.23 $\pm$ 0.12} \\
\midrule
\multirow{3}{*}{10} & 0 & 0.20 $\pm$ 0.11 & 0.04 $\pm$ 0.05 & 0.22 $\pm$ 0.11 & 0.18 $\pm$ 0.11 & 0.06 $\pm$ 0.00 & \underline{0.27 $\pm$ 0.12} & \underline{0.27 $\pm$ 0.12} & 0.26 $\pm$ 0.12 & \underline{0.27 $\pm$ 0.12} & \textbf{0.32 $\pm$ 0.13} & \textbf{0.32 $\pm$ 0.13} \\
 & 1 & 0.20 $\pm$ 0.11 & 0.04 $\pm$ 0.05 & 0.20 $\pm$ 0.10 & 0.17 $\pm$ 0.11 & 0.07 $\pm$ 0.00 & 0.26 $\pm$ 0.12 & 0.26 $\pm$ 0.12 & 0.26 $\pm$ 0.12 & 0.26 $\pm$ 0.12 & \textbf{0.31 $\pm$ 0.13} & \underline{0.30 $\pm$ 0.13} \\
 & 2 & 0.20 $\pm$ 0.11 & 0.04 $\pm$ 0.05 & 0.19 $\pm$ 0.10 & 0.17 $\pm$ 0.11 & 0.07 $\pm$ 0.00 & \underline{0.26 $\pm$ 0.12} & \underline{0.26 $\pm$ 0.12} & \underline{0.26 $\pm$ 0.12} & \underline{0.26 $\pm$ 0.12} & \textbf{0.30 $\pm$ 0.13} & \textbf{0.30 $\pm$ 0.13} \\
\midrule
\multirow{3}{*}{15} & 0 & 0.23 $\pm$ 0.11 & 0.05 $\pm$ 0.05 & 0.26 $\pm$ 0.11 & 0.21 $\pm$ 0.11 & 0.06 $\pm$ 0.00 & 0.31 $\pm$ 0.12 & 0.31 $\pm$ 0.11 & 0.30 $\pm$ 0.12 & 0.31 $\pm$ 0.12 & \textbf{0.35 $\pm$ 0.13} & \underline{0.34 $\pm$ 0.13} \\
 & 1 & 0.23 $\pm$ 0.10 & 0.05 $\pm$ 0.05 & 0.24 $\pm$ 0.10 & 0.19 $\pm$ 0.11 & 0.07 $\pm$ 0.00 & 0.29 $\pm$ 0.11 & \underline{0.30 $\pm$ 0.11} & 0.29 $\pm$ 0.12 & \underline{0.30 $\pm$ 0.11} & \textbf{0.33 $\pm$ 0.12} & \textbf{0.33 $\pm$ 0.12} \\
 & 2 & 0.23 $\pm$ 0.11 & 0.05 $\pm$ 0.05 & 0.23 $\pm$ 0.10 & 0.18 $\pm$ 0.11 & 0.07 $\pm$ 0.00 & 0.29 $\pm$ 0.11 & \underline{0.30 $\pm$ 0.11} & 0.29 $\pm$ 0.12 & \underline{0.30 $\pm$ 0.11} & \textbf{0.33 $\pm$ 0.12} & \textbf{0.33 $\pm$ 0.12} \\
\midrule
\multirow{3}{*}{20} & 0 & 0.25 $\pm$ 0.10 & 0.06 $\pm$ 0.05 & 0.28 $\pm$ 0.10 & 0.21 $\pm$ 0.11 & 0.06 $\pm$ 0.00 & 0.32 $\pm$ 0.11 & 0.33 $\pm$ 0.11 & 0.31 $\pm$ 0.11 & 0.33 $\pm$ 0.11 & \textbf{0.36 $\pm$ 0.12} & \underline{0.35 $\pm$ 0.12} \\
 & 1 & 0.24 $\pm$ 0.10 & 0.05 $\pm$ 0.05 & 0.26 $\pm$ 0.10 & 0.19 $\pm$ 0.10 & 0.07 $\pm$ 0.00 & 0.31 $\pm$ 0.11 & \underline{0.32 $\pm$ 0.11} & 0.30 $\pm$ 0.11 & \underline{0.32 $\pm$ 0.11} & \textbf{0.34 $\pm$ 0.12} & \textbf{0.34 $\pm$ 0.12} \\
 & 2 & 0.25 $\pm$ 0.10 & 0.05 $\pm$ 0.05 & 0.25 $\pm$ 0.10 & 0.19 $\pm$ 0.10 & 0.07 $\pm$ 0.00 & 0.31 $\pm$ 0.11 & \underline{0.32 $\pm$ 0.11} & 0.30 $\pm$ 0.11 & \underline{0.32 $\pm$ 0.11} & \textbf{0.34 $\pm$ 0.12} & \textbf{0.34 $\pm$ 0.12} \\
\midrule
\end{tabular}}

\end{table*}

\begin{figure}[ht]
    \centering
    \includegraphics[width=\linewidth]{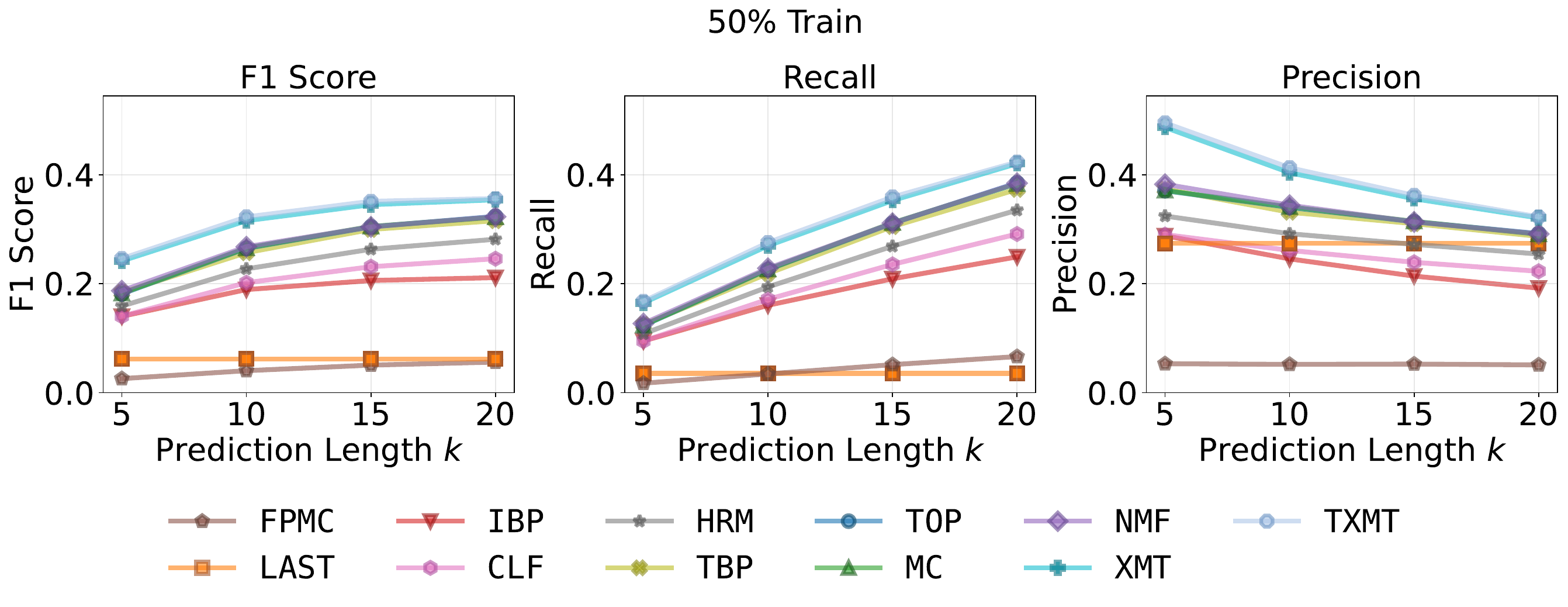}
    \caption{Comparison of \xmt and \txmt with Baseline methods using a \textit{50\% Train} and $h$=0}
    \label{fig:coop_50_day_0}
\end{figure}

\begin{figure}[ht]
    \centering
    \includegraphics[width=\linewidth]{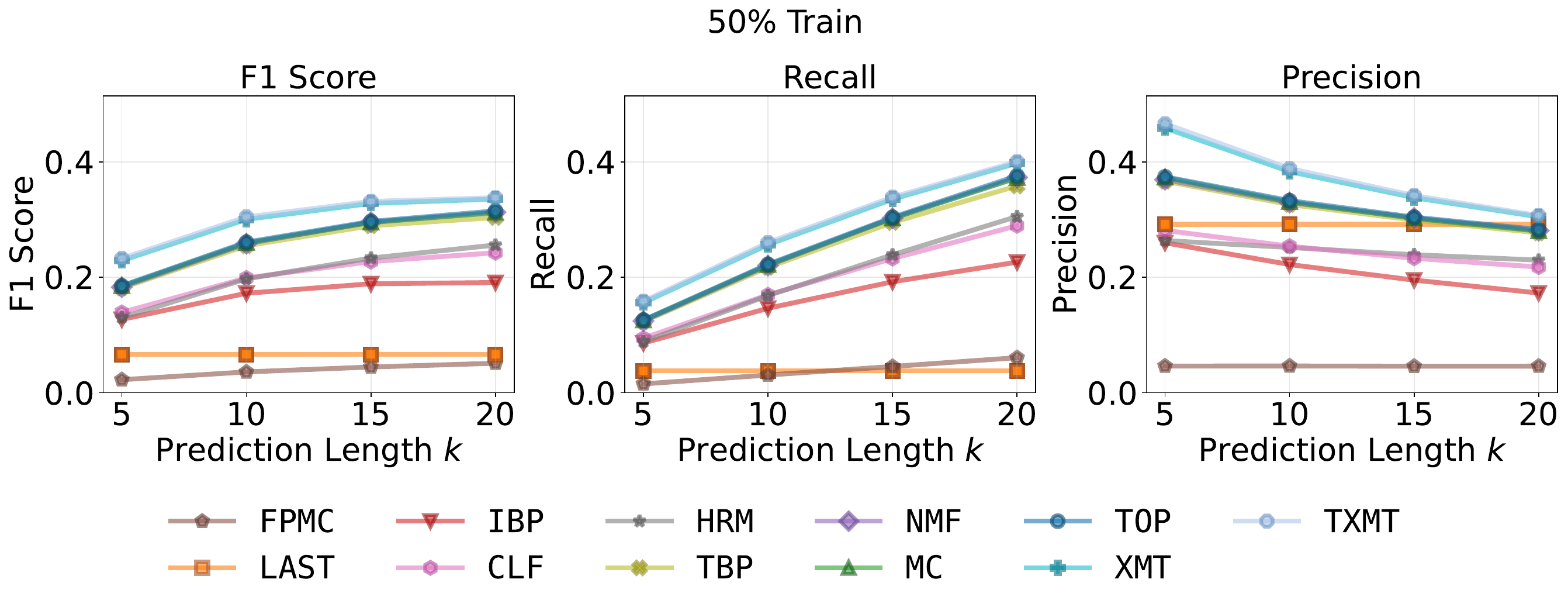}
    \caption{Comparison of \xmt and \txmt with Baseline methods using a \textit{50\% Train} and $h$=1}
    \label{fig:coop_50_day_1}
\end{figure}

%50% SPLIT
Figures~\ref{fig:coop_50_day_0}, and~\ref{fig:coop_50_day_1} presents the result with \textit{50\% Train} and $h = 0$ and $h = 1$ respectively. Tables~\ref{tab:comparison_precision_50},~\ref{tab:comparison_recall_50} and~\ref{tab:comparison_recall_50} report respectively the precision, recall and F1 Score with \textit{50\% Train}. This is one of the settings we considered in~\Cref{sec:exp_results}, here we offer a more in-depth view of the results, exploring how they change when we decrease the $h$.
In particular, when $h$ = 0, we can see a small increase in the recall and F1 Score of both \xmt and \txmt.


\begin{table*}[ht]
        \caption{Precision comparison for different prediction lengths k and days $h$ with \textit{50\% Train}}
            \label{tab:comparison_precision_50}
        \vspace{0.5cm}
            \centering
            \small
            \setlength{\tabcolsep}{4pt}
            \resizebox{\linewidth}{!}{
\begin{tabular}{cc>{\columncolor{gray!20}}cc>{\columncolor{gray!20}}cc>{\columncolor{gray!20}}cc>{\columncolor{gray!20}}cc>{\columncolor{gray!20}}cc>{\columncolor{gray!20}}c}
            \hline
            \multicolumn{13}{c}{50\% Train} \\
            \hline
            \textbf{k} & \textbf{$h$} & \textbf{clf} & \textbf{fpmc} & \textbf{hrm} & \textbf{ibp} & \textbf{last} & \textbf{mc} & \textbf{nmf} & \textbf{tbp} & \textbf{top} & \textbf{txmt} & \textbf{xmt} \\
            \hline
            \multirow{3}{*}{5} & 0 & 0.29 $\pm$ 0.22 & 0.05 $\pm$ 0.10 & 0.32 $\pm$ 0.22 & 0.29 $\pm$ 0.22 & 0.27 $\pm$ 0.00 & 0.37 $\pm$ 0.23 & 0.38 $\pm$ 0.23 & 0.37 $\pm$ 0.24 & 0.37 $\pm$ 0.23 & \textbf{0.50 $\pm$ 0.26} & \underline{0.49 $\pm$ 0.25} \\
             & 1 & 0.28 $\pm$ 0.21 & 0.05 $\pm$ 0.10 & 0.26 $\pm$ 0.\multirow{3}{*}{20} & 0.26 $\pm$ 0.22 & 0.29 $\pm$ 0.00 & 0.37 $\pm$ 0.23 & 0.37 $\pm$ 0.23 & 0.37 $\pm$ 0.23 & 0.37 $\pm$ 0.23 & \textbf{0.47 $\pm$ 0.24} & \underline{0.46 $\pm$ 0.24} \\
             & 2 & 0.29 $\pm$ 0.21 & 0.05 $\pm$ 0.10 & 0.26 $\pm$ 0.\multirow{3}{*}{20} & 0.25 $\pm$ 0.21 & 0.29 $\pm$ 0.00 & 0.37 $\pm$ 0.23 & 0.38 $\pm$ 0.23 & 0.37 $\pm$ 0.23 & 0.37 $\pm$ 0.23 & \textbf{0.48 $\pm$ 0.24} & \underline{0.47 $\pm$ 0.24} \\
            \midrule
            \multirow{3}{*}{10} & 0 & 0.26 $\pm$ 0.16 & 0.05 $\pm$ 0.07 & 0.29 $\pm$ 0.15 & 0.25 $\pm$ 0.15 & 0.27 $\pm$ 0.00 & 0.34 $\pm$ 0.17 & 0.34 $\pm$ 0.17 & 0.33 $\pm$ 0.17 & 0.34 $\pm$ 0.17 & \textbf{0.41 $\pm$ 0.19} & \underline{0.40 $\pm$ 0.18} \\
             & 1 & 0.25 $\pm$ 0.15 & 0.05 $\pm$ 0.07 & 0.25 $\pm$ 0.14 & 0.22 $\pm$ 0.15 & 0.29 $\pm$ 0.00 & 0.33 $\pm$ 0.17 & 0.33 $\pm$ 0.16 & 0.33 $\pm$ 0.17 & 0.33 $\pm$ 0.17 & \textbf{0.39 $\pm$ 0.18} & \underline{0.38 $\pm$ 0.18} \\
             & 2 & 0.26 $\pm$ 0.15 & 0.05 $\pm$ 0.07 & 0.25 $\pm$ 0.14 & 0.22 $\pm$ 0.15 & 0.29 $\pm$ 0.00 & 0.33 $\pm$ 0.17 & \underline{0.34 $\pm$ 0.16} & 0.33 $\pm$ 0.17 & \underline{0.34 $\pm$ 0.16} & \textbf{0.39 $\pm$ 0.18} & \textbf{0.39 $\pm$ 0.18} \\
            \midrule
            \multirow{3}{*}{15} & 0 & 0.24 $\pm$ 0.13 & 0.05 $\pm$ 0.06 & 0.27 $\pm$ 0.13 & 0.21 $\pm$ 0.13 & 0.27 $\pm$ 0.00 & \underline{0.31 $\pm$ 0.14} & \underline{0.31 $\pm$ 0.14} & \underline{0.31 $\pm$ 0.14} & \underline{0.31 $\pm$ 0.14} & \textbf{0.36 $\pm$ 0.15} & \textbf{0.36 $\pm$ 0.15} \\
             & 1 & 0.23 $\pm$ 0.12 & 0.05 $\pm$ 0.05 & 0.24 $\pm$ 0.12 & 0.20 $\pm$ 0.13 & 0.29 $\pm$ 0.00 & \underline{0.30 $\pm$ 0.14} & \underline{0.30 $\pm$ 0.14} & \underline{0.30 $\pm$ 0.14} & \underline{0.30 $\pm$ 0.14} & \textbf{0.34 $\pm$ 0.15} & \textbf{0.34 $\pm$ 0.15} \\
             & 2 & 0.24 $\pm$ 0.12 & 0.05 $\pm$ 0.06 & 0.23 $\pm$ 0.12 & 0.19 $\pm$ 0.12 & 0.29 $\pm$ 0.00 & 0.30 $\pm$ 0.14 & 0.31 $\pm$ 0.13 & 0.30 $\pm$ 0.14 & 0.31 $\pm$ 0.13 & \textbf{0.35 $\pm$ 0.15} & \underline{0.34 $\pm$ 0.15} \\
            \midrule
            \multirow{3}{*}{20} & 0 & 0.22 $\pm$ 0.11 & 0.05 $\pm$ 0.05 & 0.25 $\pm$ 0.11 & 0.19 $\pm$ 0.11 & 0.27 $\pm$ 0.00 & \underline{0.29 $\pm$ 0.12} & \underline{0.29 $\pm$ 0.12} & \underline{0.29 $\pm$ 0.13} & \underline{0.29 $\pm$ 0.12} & \textbf{0.32 $\pm$ 0.13} & \textbf{0.32 $\pm$ 0.13} \\
             & 1 & 0.22 $\pm$ 0.10 & 0.05 $\pm$ 0.05 & 0.23 $\pm$ 0.10 & 0.17 $\pm$ 0.11 & 0.29 $\pm$ 0.00 & 0.28 $\pm$ 0.12 & 0.28 $\pm$ 0.12 & 0.28 $\pm$ 0.12 & 0.28 $\pm$ 0.12 & \textbf{0.31 $\pm$ 0.13} & \underline{0.30 $\pm$ 0.13} \\
             & 2 & 0.22 $\pm$ 0.11 & 0.05 $\pm$ 0.05 & 0.23 $\pm$ 0.10 & 0.17 $\pm$ 0.10 & \underline{0.29 $\pm$ 0.00} & 0.28 $\pm$ 0.12 & 0.28 $\pm$ 0.12 & 0.28 $\pm$ 0.12 & 0.28 $\pm$ 0.12 & \textbf{0.31 $\pm$ 0.13} & \textbf{0.31 $\pm$ 0.13} \\
            \midrule
            \end{tabular}}
            
            \end{table*}

\begin{table*}[ht]
        \caption{Recall comparison for different prediction lengths (k) and days (d) with \textit{50\% Train}}
            \label{tab:comparison_recall_50}
        \vspace{0.5cm}
            \centering
            \small
            \setlength{\tabcolsep}{4pt}
            \resizebox{\linewidth}{!}{
            \begin{tabular}{cc>{\columncolor{gray!20}}cc>{\columncolor{gray!20}}cc>{\columncolor{gray!20}}cc>{\columncolor{gray!20}}cc>{\columncolor{gray!20}}cc>{\columncolor{gray!20}}c}
            \hline
            \multicolumn{13}{c}{50\% Train} \\
            \hline
            \textbf{k} & \textbf{$h$} & \textbf{clf} & \textbf{fpmc} & \textbf{hrm} & \textbf{ibp} & \textbf{last} & \textbf{mc} & \textbf{nmf} & \textbf{tbp} & \textbf{top} & \textbf{txmt} & \textbf{xmt} \\
            \hline
            \multirow{3}{*}{5} & 0 & 0.09 $\pm$ 0.07 & 0.02 $\pm$ 0.03 & 0.11 $\pm$ 0.07 & 0.09 $\pm$ 0.07 & 0.04 $\pm$ 0.00 & 0.12 $\pm$ 0.08 & 0.13 $\pm$ 0.08 & 0.12 $\pm$ 0.08 & 0.12 $\pm$ 0.08 & \textbf{0.17 $\pm$ 0.09} & \underline{0.16 $\pm$ 0.09} \\
             & 1 & 0.09 $\pm$ 0.07 & 0.02 $\pm$ 0.03 & 0.09 $\pm$ 0.07 & 0.09 $\pm$ 0.07 & 0.04 $\pm$ 0.00 & \underline{0.13 $\pm$ 0.08} & 0.12 $\pm$ 0.08 & 0.12 $\pm$ 0.08 & \underline{0.13 $\pm$ 0.08} & \textbf{0.16 $\pm$ 0.09} & \textbf{0.16 $\pm$ 0.09} \\
             & 2 & 0.09 $\pm$ 0.07 & 0.02 $\pm$ 0.03 & 0.09 $\pm$ 0.07 & 0.08 $\pm$ 0.07 & 0.04 $\pm$ 0.00 & \underline{0.12 $\pm$ 0.08} & \underline{0.12 $\pm$ 0.08} & \underline{0.12 $\pm$ 0.08} & \underline{0.12 $\pm$ 0.08} & \textbf{0.16 $\pm$ 0.09} & \textbf{0.16 $\pm$ 0.08} \\
            \midrule
            \multirow{3}{*}{10} & 0 & 0.17 $\pm$ 0.10 & 0.03 $\pm$ 0.05 & 0.19 $\pm$ 0.10 & 0.16 $\pm$ 0.09 & 0.04 $\pm$ 0.00 & 0.23 $\pm$ 0.10 & 0.23 $\pm$ 0.10 & 0.22 $\pm$ 0.11 & 0.23 $\pm$ 0.11 & \textbf{0.28 $\pm$ 0.12} & \underline{0.27 $\pm$ 0.12} \\
             & 1 & 0.17 $\pm$ 0.10 & 0.03 $\pm$ 0.05 & 0.17 $\pm$ 0.09 & 0.15 $\pm$ 0.10 & 0.04 $\pm$ 0.00 & \underline{0.22 $\pm$ 0.11} & \underline{0.22 $\pm$ 0.10} & \underline{0.22 $\pm$ 0.11} & \underline{0.22 $\pm$ 0.10} & \textbf{0.26 $\pm$ 0.12} & \textbf{0.26 $\pm$ 0.12} \\
             & 2 & 0.17 $\pm$ 0.09 & 0.03 $\pm$ 0.04 & 0.16 $\pm$ 0.09 & 0.14 $\pm$ 0.09 & 0.04 $\pm$ 0.00 & \underline{0.22 $\pm$ 0.10} & \underline{0.22 $\pm$ 0.10} & \underline{0.22 $\pm$ 0.10} & \underline{0.22 $\pm$ 0.10} & \textbf{0.26 $\pm$ 0.11} & \textbf{0.26 $\pm$ 0.11} \\
            \midrule
            \multirow{3}{*}{15} & 0 & 0.24 $\pm$ 0.12 & 0.05 $\pm$ 0.06 & 0.27 $\pm$ 0.11 & 0.21 $\pm$ 0.11 & 0.04 $\pm$ 0.00 & 0.31 $\pm$ 0.12 & 0.31 $\pm$ 0.12 & 0.30 $\pm$ 0.12 & 0.31 $\pm$ 0.12 & \textbf{0.36 $\pm$ 0.14} & \underline{0.35 $\pm$ 0.14} \\
             & 1 & 0.23 $\pm$ 0.11 & 0.05 $\pm$ 0.06 & 0.24 $\pm$ 0.11 & 0.19 $\pm$ 0.11 & 0.04 $\pm$ 0.00 & \underline{0.30 $\pm$ 0.12} & \underline{0.30 $\pm$ 0.12} & 0.29 $\pm$ 0.12 & \underline{0.30 $\pm$ 0.12} & \textbf{0.34 $\pm$ 0.13} & \textbf{0.34 $\pm$ 0.13} \\
             & 2 & 0.23 $\pm$ 0.11 & 0.05 $\pm$ 0.05 & 0.23 $\pm$ 0.10 & 0.18 $\pm$ 0.11 & 0.04 $\pm$ 0.00 & \underline{0.30 $\pm$ 0.12} & \underline{0.30 $\pm$ 0.12} & 0.29 $\pm$ 0.12 & \underline{0.30 $\pm$ 0.12} & \textbf{0.33 $\pm$ 0.13} & \textbf{0.33 $\pm$ 0.13} \\
            \midrule
            \multirow{3}{*}{20} & 0 & 0.29 $\pm$ 0.12 & 0.07 $\pm$ 0.06 & 0.33 $\pm$ 0.12 & 0.25 $\pm$ 0.12 & 0.04 $\pm$ 0.00 & \underline{0.38 $\pm$ 0.13} & \underline{0.38 $\pm$ 0.13} & 0.37 $\pm$ 0.14 & \underline{0.38 $\pm$ 0.13} & \textbf{0.42 $\pm$ 0.15} & \textbf{0.42 $\pm$ 0.15} \\
             & 1 & 0.29 $\pm$ 0.12 & 0.06 $\pm$ 0.06 & 0.31 $\pm$ 0.12 & 0.23 $\pm$ 0.12 & 0.04 $\pm$ 0.00 & 0.37 $\pm$ 0.13 & 0.37 $\pm$ 0.13 & 0.36 $\pm$ 0.14 & \underline{0.38 $\pm$ 0.13} & \textbf{0.40 $\pm$ 0.14} & \textbf{0.40 $\pm$ 0.14} \\
             & 2 & 0.29 $\pm$ 0.12 & 0.06 $\pm$ 0.06 & 0.30 $\pm$ 0.12 & 0.22 $\pm$ 0.12 & 0.04 $\pm$ 0.00 & \underline{0.37 $\pm$ 0.13} & \underline{0.37 $\pm$ 0.13} & 0.35 $\pm$ 0.13 & \underline{0.37 $\pm$ 0.13} & \textbf{0.39 $\pm$ 0.14} & \textbf{0.39 $\pm$ 0.14} \\
            \midrule
            \end{tabular}
            }
            
            \end{table*}
\begin{table*}[ht]
\caption{F1 Score comparison for different prediction lengths (k) and days (d)  with \textit{50\% Train}}
\label{tab:comparison_f1_score_50}
\vspace{0.5cm}
\centering
\small
\setlength{\tabcolsep}{4pt}
\resizebox{\linewidth}{!}{
\begin{tabular}{cc>{\columncolor{gray!20}}cc>{\columncolor{gray!20}}cc>{\columncolor{gray!20}}cc>{\columncolor{gray!20}}cc>{\columncolor{gray!20}}cc>{\columncolor{gray!20}}c}
\hline
\multicolumn{13}{c}{50\% Train} \\
\hline
\textbf{k} & \textbf{$h$} & \textbf{clf} & \textbf{fpmc} & \textbf{hrm} & \textbf{ibp} & \textbf{last} & \textbf{mc} & \textbf{nmf} & \textbf{tbp} & \textbf{top} & \textbf{txmt} & \textbf{xmt} \\
\hline
\multirow{3}{*}{5} & 0 & 0.14 $\pm$ 0.10 & 0.03 $\pm$ 0.05 & 0.16 $\pm$ 0.10 & 0.14 $\pm$ 0.10 & 0.06 $\pm$ 0.00 & 0.18 $\pm$ 0.11 & 0.19 $\pm$ 0.11 & 0.18 $\pm$ 0.12 & 0.18 $\pm$ 0.11 & \textbf{0.25 $\pm$ 0.13} & \underline{0.24 $\pm$ 0.12} \\
 & 1 & 0.14 $\pm$ 0.10 & 0.02 $\pm$ 0.05 & 0.13 $\pm$ 0.10 & 0.13 $\pm$ 0.10 & 0.07 $\pm$ 0.00 & 0.18 $\pm$ 0.11 & 0.18 $\pm$ 0.11 & 0.18 $\pm$ 0.11 & \underline{0.19 $\pm$ 0.11} & \textbf{0.23 $\pm$ 0.12} & \textbf{0.23 $\pm$ 0.12} \\
 & 2 & 0.14 $\pm$ 0.10 & 0.02 $\pm$ 0.05 & 0.13 $\pm$ 0.10 & 0.12 $\pm$ 0.10 & 0.07 $\pm$ 0.00 & \underline{0.18 $\pm$ 0.11} & \underline{0.18 $\pm$ 0.11} & \underline{0.18 $\pm$ 0.11} & \underline{0.18 $\pm$ 0.11} & \textbf{0.23 $\pm$ 0.12} & \textbf{0.23 $\pm$ 0.12} \\
\midrule
\multirow{3}{*}{10} & 0 & 0.20 $\pm$ 0.12 & 0.04 $\pm$ 0.05 & 0.23 $\pm$ 0.11 & 0.19 $\pm$ 0.11 & 0.06 $\pm$ 0.00 & \underline{0.27 $\pm$ 0.12} & \underline{0.27 $\pm$ 0.12} & 0.26 $\pm$ 0.12 & 0.26 $\pm$ 0.12 & \textbf{0.32 $\pm$ 0.13} & \textbf{0.32 $\pm$ 0.13} \\
 & 1 & 0.20 $\pm$ 0.11 & 0.04 $\pm$ 0.05 & 0.20 $\pm$ 0.10 & 0.17 $\pm$ 0.11 & 0.07 $\pm$ 0.00 & \underline{0.26 $\pm$ 0.12} & \underline{0.26 $\pm$ 0.12} & 0.25 $\pm$ 0.12 & \underline{0.26 $\pm$ 0.12} & \textbf{0.30 $\pm$ 0.13} & \textbf{0.30 $\pm$ 0.13} \\
 & 2 & 0.20 $\pm$ 0.11 & 0.04 $\pm$ 0.05 & 0.19 $\pm$ 0.10 & 0.17 $\pm$ 0.11 & 0.07 $\pm$ 0.00 & 0.26 $\pm$ 0.12 & 0.26 $\pm$ 0.11 & 0.26 $\pm$ 0.12 & 0.26 $\pm$ 0.12 & \textbf{0.31 $\pm$ 0.13} & \underline{0.30 $\pm$ 0.13} \\
\midrule
\multirow{3}{*}{15} & 0 & 0.23 $\pm$ 0.11 & 0.05 $\pm$ 0.05 & 0.26 $\pm$ 0.11 & 0.21 $\pm$ 0.11 & 0.06 $\pm$ 0.00 & \underline{0.31 $\pm$ 0.12} & 0.30 $\pm$ 0.11 & 0.30 $\pm$ 0.12 & 0.30 $\pm$ 0.12 & \textbf{0.35 $\pm$ 0.13} & \textbf{0.35 $\pm$ 0.13} \\
 & 1 & 0.23 $\pm$ 0.10 & 0.04 $\pm$ 0.05 & 0.23 $\pm$ 0.10 & 0.19 $\pm$ 0.11 & 0.07 $\pm$ 0.00 & 0.29 $\pm$ 0.11 & \underline{0.30 $\pm$ 0.11} & 0.29 $\pm$ 0.12 & \underline{0.30 $\pm$ 0.11} & \textbf{0.33 $\pm$ 0.12} & \textbf{0.33 $\pm$ 0.12} \\
 & 2 & 0.23 $\pm$ 0.11 & 0.05 $\pm$ 0.05 & 0.23 $\pm$ 0.10 & 0.18 $\pm$ 0.10 & 0.07 $\pm$ 0.00 & 0.29 $\pm$ 0.11 & \underline{0.30 $\pm$ 0.11} & 0.29 $\pm$ 0.12 & \underline{0.30 $\pm$ 0.11} & \textbf{0.33 $\pm$ 0.12} & \textbf{0.33 $\pm$ 0.12} \\
\midrule
\multirow{3}{*}{20} & 0 & 0.25 $\pm$ 0.11 & 0.06 $\pm$ 0.05 & 0.28 $\pm$ 0.10 & 0.21 $\pm$ 0.11 & 0.06 $\pm$ 0.00 & 0.32 $\pm$ 0.11 & 0.32 $\pm$ 0.11 & 0.32 $\pm$ 0.12 & 0.32 $\pm$ 0.11 & \textbf{0.36 $\pm$ 0.12} & \underline{0.35 $\pm$ 0.12} \\
 & 1 & 0.24 $\pm$ 0.10 & 0.05 $\pm$ 0.05 & 0.26 $\pm$ 0.10 & 0.19 $\pm$ 0.10 & 0.07 $\pm$ 0.00 & 0.31 $\pm$ 0.11 & 0.31 $\pm$ 0.11 & 0.30 $\pm$ 0.11 & \underline{0.32 $\pm$ 0.11} & \textbf{0.34 $\pm$ 0.12} & \textbf{0.34 $\pm$ 0.12} \\
 & 2 & 0.25 $\pm$ 0.10 & 0.05 $\pm$ 0.05 & 0.25 $\pm$ 0.10 & 0.19 $\pm$ 0.10 & 0.07 $\pm$ 0.00 & \underline{0.31 $\pm$ 0.11} & \underline{0.31 $\pm$ 0.11} & 0.30 $\pm$ 0.11 & \underline{0.31 $\pm$ 0.11} & \textbf{0.34 $\pm$ 0.12} & \textbf{0.34 $\pm$ 0.12} \\
\midrule
\end{tabular}}

\end{table*}

\begin{figure}[ht]
    \centering
    \includegraphics[width=\linewidth]{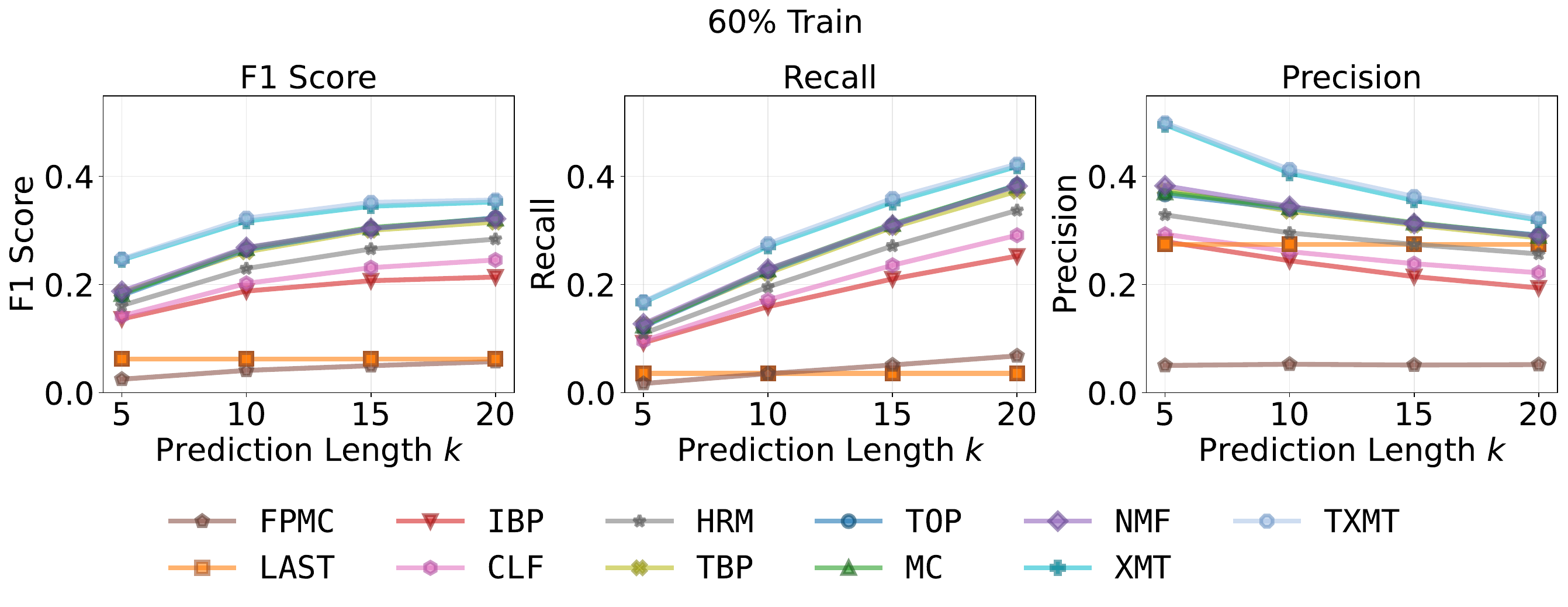}
    \caption{Comparison of \xmt and \txmt with Baseline methods using a \textit{60\% Train} and $h$=0}
    \label{fig:coop_60_day_0}
\end{figure}

\begin{figure}[ht]
    \centering
    \includegraphics[width=\linewidth]{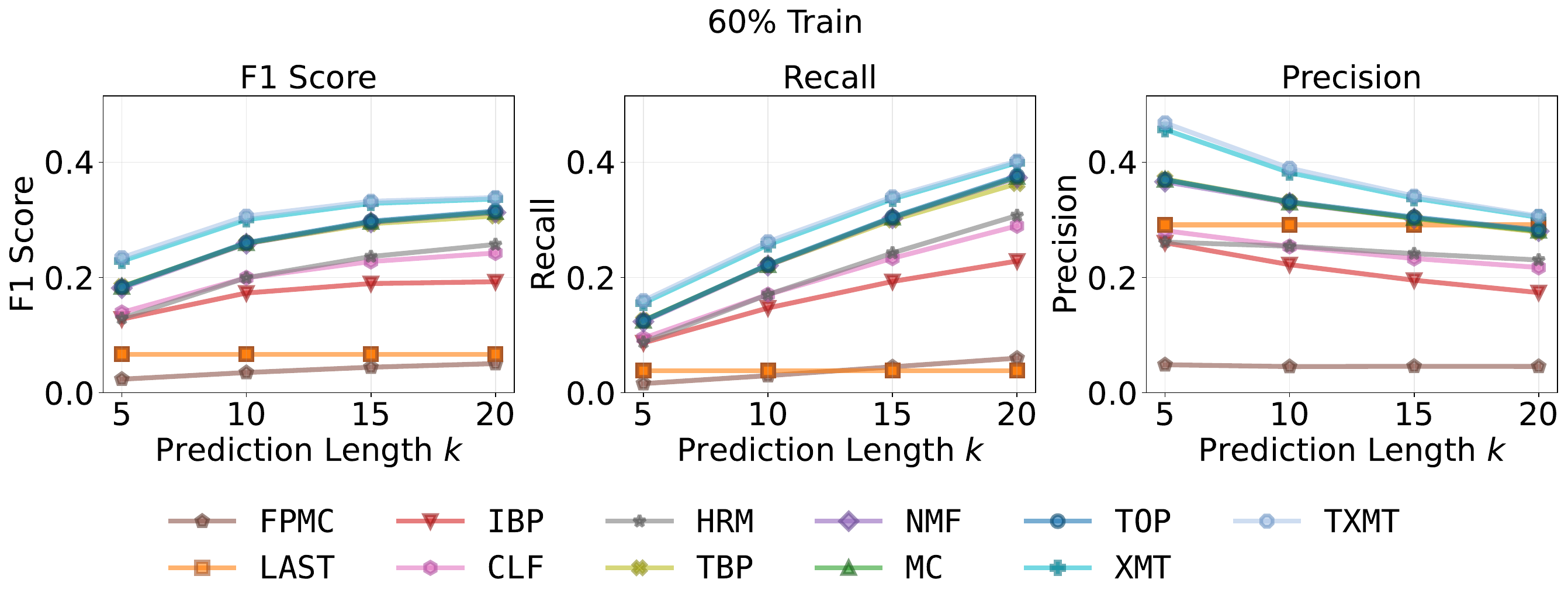}
    \caption{Comparison of \xmt and \txmt with Baseline methods using a \textit{60\% Train} and $h$=1}
    \label{fig:coop_60_day_1}
\end{figure}

\begin{figure}[ht]
    \centering
    \includegraphics[width=\linewidth]{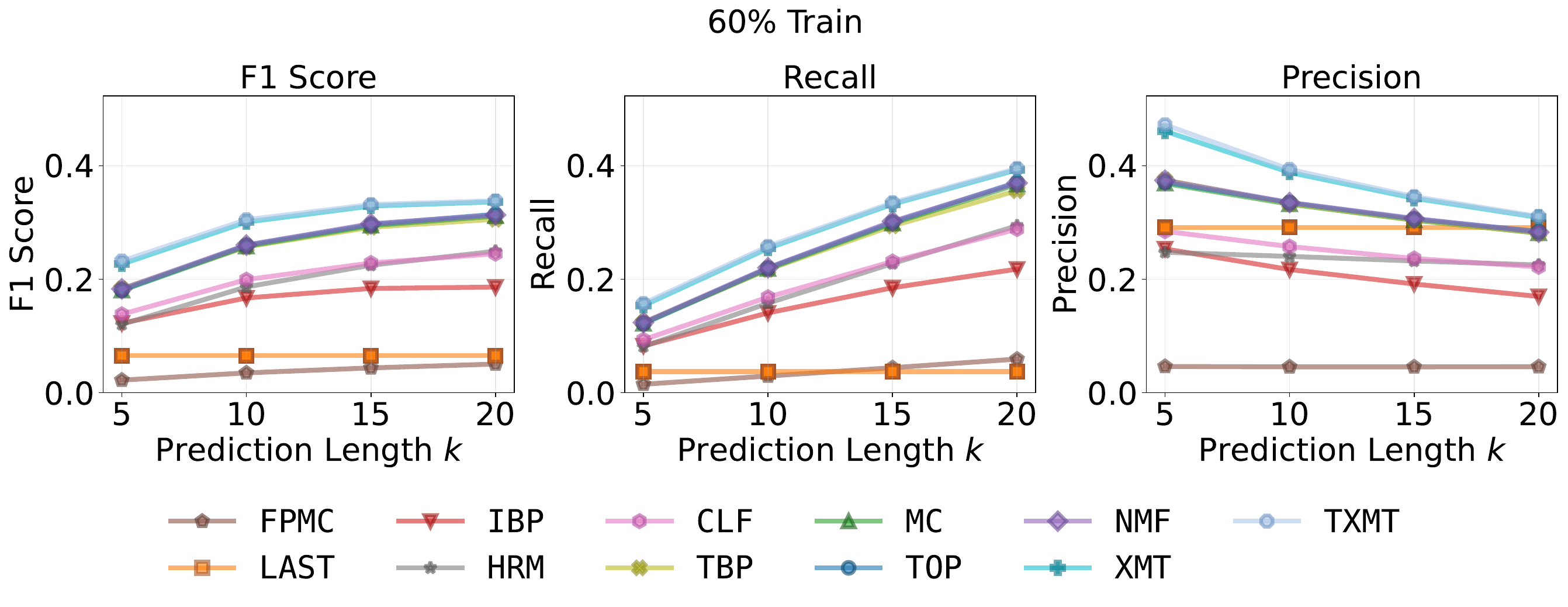}
    \caption{Comparison of \xmt and \txmt with Baseline methods using a \textit{60\% Train} and $h$=2}
    \label{fig:coop_60_day_2}
\end{figure}

%60% SPLIT
Figures~\ref{fig:coop_60_day_0}, ~\ref{fig:coop_60_day_1}and~\ref{fig:coop_60_day_2} presents the result with \textit{60\% Train} and $h = 0$, $h = 1$ and  $h = 2$ respectively. Tables~\ref{tab:comparison_precision_60},~\ref{tab:comparison_recall_60} and~\ref{tab:comparison_recall_60} report respectively the precision, recall and F1 Score with \textit{60\% Train}. Despite the increase of the training dataset, we notice only small improvements in terms of F1 Score, Recall, and Precision, which supports our thesis that our proposed methods offer good performance even when trained with a small amount of data.

\begin{table*}[ht]
            \caption{Precision comparison for different prediction lengths k and days $h$ with \textit{60\% Train}}
                \label{tab:comparison_precision_60}
            \vspace{0.5cm}
                \centering
                \small
                \setlength{\tabcolsep}{4pt}
                \resizebox{\linewidth}{!}{
\begin{tabular}{cc>{\columncolor{gray!20}}cc>{\columncolor{gray!20}}cc>{\columncolor{gray!20}}cc>{\columncolor{gray!20}}cc>{\columncolor{gray!20}}cc>{\columncolor{gray!20}}c}
                \hline
                \multicolumn{13}{c}{60\% Train} \\
                \hline
                \textbf{k} & \textbf{$h$} & \textbf{clf} & \textbf{fpmc} & \textbf{hrm} & \textbf{ibp} & \textbf{last} & \textbf{mc} & \textbf{nmf} & \textbf{tbp} & \textbf{top} & \textbf{txmt} & \textbf{xmt} \\
                \hline
                \multirow{3}{*}{5} & 0 & 0.29 $\pm$ 0.22 & 0.05 $\pm$ 0.10 & 0.33 $\pm$ 0.22 & 0.28 $\pm$ 0.21 & 0.27 $\pm$ 0.00 & 0.37 $\pm$ 0.23 & \underline{0.38 $\pm$ 0.23} & \underline{0.38 $\pm$ 0.24} & 0.37 $\pm$ 0.23 & \textbf{0.50 $\pm$ 0.26} & \textbf{0.50 $\pm$ 0.26} \\
                 & 1 & 0.28 $\pm$ 0.21 & 0.05 $\pm$ 0.10 & 0.26 $\pm$ 0.\multirow{3}{*}{20} & 0.26 $\pm$ 0.21 & 0.29 $\pm$ 0.00 & 0.37 $\pm$ 0.23 & 0.37 $\pm$ 0.23 & 0.37 $\pm$ 0.23 & 0.37 $\pm$ 0.23 & \textbf{0.47 $\pm$ 0.25} & \underline{0.46 $\pm$ 0.24} \\
                 & 2 & 0.28 $\pm$ 0.21 & 0.05 $\pm$ 0.09 & 0.25 $\pm$ 0.\multirow{3}{*}{20} & 0.25 $\pm$ 0.21 & 0.29 $\pm$ 0.00 & 0.37 $\pm$ 0.23 & 0.37 $\pm$ 0.23 & 0.37 $\pm$ 0.23 & 0.37 $\pm$ 0.23 & \textbf{0.47 $\pm$ 0.24} & \underline{0.46 $\pm$ 0.24} \\
                \midrule
                \multirow{3}{*}{10} & 0 & 0.26 $\pm$ 0.16 & 0.05 $\pm$ 0.07 & 0.30 $\pm$ 0.16 & 0.24 $\pm$ 0.16 & 0.27 $\pm$ 0.00 & \underline{0.34 $\pm$ 0.17} & \underline{0.34 $\pm$ 0.17} & \underline{0.34 $\pm$ 0.17} & \underline{0.34 $\pm$ 0.17} & \textbf{0.41 $\pm$ 0.19} & \textbf{0.41 $\pm$ 0.19} \\
                 & 1 & 0.25 $\pm$ 0.15 & 0.05 $\pm$ 0.07 & 0.25 $\pm$ 0.14 & 0.22 $\pm$ 0.15 & 0.29 $\pm$ 0.00 & 0.33 $\pm$ 0.17 & 0.33 $\pm$ 0.17 & 0.33 $\pm$ 0.17 & 0.33 $\pm$ 0.17 & \textbf{0.39 $\pm$ 0.18} & \underline{0.38 $\pm$ 0.18} \\
                 & 2 & 0.26 $\pm$ 0.15 & 0.05 $\pm$ 0.07 & 0.24 $\pm$ 0.14 & 0.22 $\pm$ 0.15 & 0.29 $\pm$ 0.00 & 0.33 $\pm$ 0.17 & \underline{0.34 $\pm$ 0.16} & 0.33 $\pm$ 0.17 & 0.33 $\pm$ 0.17 & \textbf{0.39 $\pm$ 0.18} & \textbf{0.39 $\pm$ 0.18} \\
                \midrule
                \multirow{3}{*}{15} & 0 & 0.24 $\pm$ 0.13 & 0.05 $\pm$ 0.06 & 0.27 $\pm$ 0.13 & 0.21 $\pm$ 0.13 & 0.27 $\pm$ 0.00 & \underline{0.31 $\pm$ 0.14} & \underline{0.31 $\pm$ 0.14} & \underline{0.31 $\pm$ 0.14} & \underline{0.31 $\pm$ 0.14} & \textbf{0.36 $\pm$ 0.16} & \textbf{0.36 $\pm$ 0.15} \\
                 & 1 & 0.23 $\pm$ 0.12 & 0.05 $\pm$ 0.05 & 0.24 $\pm$ 0.12 & 0.20 $\pm$ 0.12 & 0.29 $\pm$ 0.00 & \underline{0.30 $\pm$ 0.14} & \underline{0.30 $\pm$ 0.14} & \underline{0.30 $\pm$ 0.14} & \underline{0.30 $\pm$ 0.14} & \textbf{0.34 $\pm$ 0.15} & \textbf{0.34 $\pm$ 0.15} \\
                 & 2 & 0.24 $\pm$ 0.12 & 0.05 $\pm$ 0.05 & 0.23 $\pm$ 0.12 & 0.19 $\pm$ 0.12 & 0.29 $\pm$ 0.00 & 0.30 $\pm$ 0.14 & 0.31 $\pm$ 0.13 & 0.30 $\pm$ 0.14 & 0.31 $\pm$ 0.14 & \textbf{0.35 $\pm$ 0.15} & \underline{0.34 $\pm$ 0.15} \\
                \midrule
                \multirow{3}{*}{20} & 0 & 0.22 $\pm$ 0.11 & 0.05 $\pm$ 0.05 & 0.26 $\pm$ 0.12 & 0.19 $\pm$ 0.11 & 0.27 $\pm$ 0.00 & \underline{0.29 $\pm$ 0.12} & \underline{0.29 $\pm$ 0.12} & \underline{0.29 $\pm$ 0.13} & \underline{0.29 $\pm$ 0.12} & \textbf{0.32 $\pm$ 0.13} & \textbf{0.32 $\pm$ 0.13} \\
                 & 1 & 0.22 $\pm$ 0.10 & 0.05 $\pm$ 0.05 & 0.23 $\pm$ 0.10 & 0.17 $\pm$ 0.10 & 0.29 $\pm$ 0.00 & 0.28 $\pm$ 0.12 & 0.28 $\pm$ 0.12 & 0.28 $\pm$ 0.12 & 0.28 $\pm$ 0.12 & \textbf{0.31 $\pm$ 0.13} & \underline{0.30 $\pm$ 0.13} \\
                 & 2 & 0.22 $\pm$ 0.11 & 0.05 $\pm$ 0.05 & 0.23 $\pm$ 0.10 & 0.17 $\pm$ 0.10 & \underline{0.29 $\pm$ 0.00} & 0.28 $\pm$ 0.12 & 0.28 $\pm$ 0.12 & 0.28 $\pm$ 0.12 & 0.28 $\pm$ 0.12 & \textbf{0.31 $\pm$ 0.13} & \textbf{0.31 $\pm$ 0.13} \\
                \midrule
                \end{tabular}}
                
                \end{table*}
\begin{table*}[ht]
            \caption{Recall comparison for different prediction lengths (k) and days (d) with \textit{60\% Train}}
                \label{tab:comparison_recall_60}
            \vspace{0.5cm}
                \centering
                \small
                \setlength{\tabcolsep}{4pt}
                \resizebox{\linewidth}{!}{
                \begin{tabular}{cc>{\columncolor{gray!20}}cc>{\columncolor{gray!20}}cc>{\columncolor{gray!20}}cc>{\columncolor{gray!20}}cc>{\columncolor{gray!20}}cc>{\columncolor{gray!20}}c}
                \hline
                \multicolumn{13}{c}{60\% Train} \\
                \hline
                \textbf{k} & \textbf{$h$} & \textbf{clf} & \textbf{fpmc} & \textbf{hrm} & \textbf{ibp} & \textbf{last} & \textbf{mc} & \textbf{nmf} & \textbf{tbp} & \textbf{top} & \textbf{txmt} & \textbf{xmt} \\
                \hline
                \multirow{3}{*}{5} & 0 & 0.10 $\pm$ 0.07 & 0.02 $\pm$ 0.03 & 0.11 $\pm$ 0.07 & 0.09 $\pm$ 0.07 & 0.04 $\pm$ 0.00 & 0.12 $\pm$ 0.08 & \underline{0.13 $\pm$ 0.08} & \underline{0.13 $\pm$ 0.08} & 0.12 $\pm$ 0.08 & \textbf{0.17 $\pm$ 0.09} & \textbf{0.17 $\pm$ 0.09} \\
                 & 1 & 0.09 $\pm$ 0.07 & 0.02 $\pm$ 0.03 & 0.09 $\pm$ 0.07 & 0.09 $\pm$ 0.07 & 0.04 $\pm$ 0.00 & 0.12 $\pm$ 0.08 & 0.12 $\pm$ 0.08 & \underline{0.13 $\pm$ 0.08} & 0.12 $\pm$ 0.08 & \textbf{0.16 $\pm$ 0.09} & \textbf{0.16 $\pm$ 0.09} \\
                 & 2 & 0.09 $\pm$ 0.07 & 0.01 $\pm$ 0.03 & 0.08 $\pm$ 0.07 & 0.08 $\pm$ 0.07 & 0.04 $\pm$ 0.00 & 0.12 $\pm$ 0.08 & 0.12 $\pm$ 0.08 & 0.12 $\pm$ 0.08 & 0.12 $\pm$ 0.08 & \textbf{0.16 $\pm$ 0.09} & \underline{0.15 $\pm$ 0.08} \\
                \midrule
                \multirow{3}{*}{10} & 0 & 0.17 $\pm$ 0.10 & 0.04 $\pm$ 0.05 & 0.20 $\pm$ 0.10 & 0.16 $\pm$ 0.10 & 0.04 $\pm$ 0.00 & 0.23 $\pm$ 0.10 & 0.23 $\pm$ 0.11 & 0.22 $\pm$ 0.11 & 0.23 $\pm$ 0.11 & \textbf{0.28 $\pm$ 0.12} & \underline{0.27 $\pm$ 0.12} \\
                 & 1 & 0.17 $\pm$ 0.10 & 0.03 $\pm$ 0.04 & 0.17 $\pm$ 0.09 & 0.15 $\pm$ 0.09 & 0.04 $\pm$ 0.00 & \underline{0.22 $\pm$ 0.11} & \underline{0.22 $\pm$ 0.10} & \underline{0.22 $\pm$ 0.11} & \underline{0.22 $\pm$ 0.10} & \textbf{0.26 $\pm$ 0.12} & \textbf{0.26 $\pm$ 0.12} \\
                 & 2 & 0.17 $\pm$ 0.09 & 0.03 $\pm$ 0.04 & 0.16 $\pm$ 0.09 & 0.14 $\pm$ 0.09 & 0.04 $\pm$ 0.00 & 0.22 $\pm$ 0.10 & 0.22 $\pm$ 0.10 & 0.22 $\pm$ 0.11 & 0.22 $\pm$ 0.10 & \textbf{0.26 $\pm$ 0.11} & \underline{0.25 $\pm$ 0.11} \\
                \midrule
                \multirow{3}{*}{15} & 0 & 0.24 $\pm$ 0.12 & 0.05 $\pm$ 0.06 & 0.27 $\pm$ 0.12 & 0.21 $\pm$ 0.11 & 0.04 $\pm$ 0.00 & 0.31 $\pm$ 0.12 & 0.31 $\pm$ 0.12 & 0.31 $\pm$ 0.13 & 0.31 $\pm$ 0.12 & \textbf{0.36 $\pm$ 0.14} & \underline{0.35 $\pm$ 0.14} \\
                 & 1 & 0.23 $\pm$ 0.11 & 0.04 $\pm$ 0.05 & 0.24 $\pm$ 0.11 & 0.19 $\pm$ 0.11 & 0.04 $\pm$ 0.00 & \underline{0.30 $\pm$ 0.12} & \underline{0.30 $\pm$ 0.12} & \underline{0.30 $\pm$ 0.13} & \underline{0.30 $\pm$ 0.12} & \textbf{0.34 $\pm$ 0.13} & \textbf{0.34 $\pm$ 0.13} \\
                 & 2 & 0.23 $\pm$ 0.11 & 0.04 $\pm$ 0.05 & 0.23 $\pm$ 0.10 & 0.19 $\pm$ 0.11 & 0.04 $\pm$ 0.00 & \underline{0.30 $\pm$ 0.12} & \underline{0.30 $\pm$ 0.12} & 0.29 $\pm$ 0.12 & \underline{0.30 $\pm$ 0.12} & \textbf{0.33 $\pm$ 0.13} & \textbf{0.33 $\pm$ 0.13} \\
                \midrule
                \multirow{3}{*}{20} & 0 & 0.29 $\pm$ 0.12 & 0.07 $\pm$ 0.07 & 0.34 $\pm$ 0.13 & 0.25 $\pm$ 0.13 & 0.04 $\pm$ 0.00 & \underline{0.38 $\pm$ 0.13} & \underline{0.38 $\pm$ 0.13} & 0.37 $\pm$ 0.14 & \underline{0.38 $\pm$ 0.13} & \textbf{0.42 $\pm$ 0.15} & \textbf{0.42 $\pm$ 0.15} \\
                 & 1 & 0.29 $\pm$ 0.12 & 0.06 $\pm$ 0.06 & 0.31 $\pm$ 0.12 & 0.23 $\pm$ 0.12 & 0.04 $\pm$ 0.00 & 0.37 $\pm$ 0.13 & 0.37 $\pm$ 0.13 & 0.36 $\pm$ 0.13 & \underline{0.38 $\pm$ 0.13} & \textbf{0.40 $\pm$ 0.14} & \textbf{0.40 $\pm$ 0.14} \\
                 & 2 & 0.29 $\pm$ 0.12 & 0.06 $\pm$ 0.06 & 0.29 $\pm$ 0.12 & 0.22 $\pm$ 0.12 & 0.04 $\pm$ 0.00 & 0.37 $\pm$ 0.13 & 0.37 $\pm$ 0.13 & 0.36 $\pm$ 0.13 & 0.37 $\pm$ 0.13 & \textbf{0.40 $\pm$ 0.14} & \underline{0.39 $\pm$ 0.14} \\
\midrule
                \end{tabular}
                }
                
                \end{table*}
\begin{table*}[ht]
\caption{F1 Score comparison for different prediction lengths (k) and days (d)  with \textit{60\% Train}}
\label{tab:comparison_f1_score_60}
\vspace{0.5cm}
\centering
\small
\setlength{\tabcolsep}{4pt}
\resizebox{\linewidth}{!}{
\begin{tabular}{cc>{\columncolor{gray!20}}cc>{\columncolor{gray!20}}cc>{\columncolor{gray!20}}cc>{\columncolor{gray!20}}cc>{\columncolor{gray!20}}cc>{\columncolor{gray!20}}c}
\hline
\multicolumn{13}{c}{60\% Train} \\
\hline
\textbf{k} & \textbf{$h$} & \textbf{clf} & \textbf{fpmc} & \textbf{hrm} & \textbf{ibp} & \textbf{last} & \textbf{mc} & \textbf{nmf} & \textbf{tbp} & \textbf{top} & \textbf{txmt} & \textbf{xmt} \\
\hline
\multirow{3}{*}{5} & 0 & 0.14 $\pm$ 0.10 & 0.03 $\pm$ 0.05 & 0.16 $\pm$ 0.10 & 0.14 $\pm$ 0.10 & 0.06 $\pm$ 0.00 & 0.18 $\pm$ 0.11 & \underline{0.19 $\pm$ 0.11} & \underline{0.19 $\pm$ 0.12} & 0.18 $\pm$ 0.11 & \textbf{0.25 $\pm$ 0.13} & \textbf{0.25 $\pm$ 0.13} \\
 & 1 & 0.14 $\pm$ 0.10 & 0.02 $\pm$ 0.05 & 0.13 $\pm$ 0.10 & 0.13 $\pm$ 0.10 & 0.07 $\pm$ 0.00 & \underline{0.18 $\pm$ 0.11} & \underline{0.18 $\pm$ 0.11} & \underline{0.18 $\pm$ 0.11} & \underline{0.18 $\pm$ 0.11} & \textbf{0.23 $\pm$ 0.12} & \textbf{0.23 $\pm$ 0.12} \\
 & 2 & 0.14 $\pm$ 0.10 & 0.02 $\pm$ 0.05 & 0.12 $\pm$ 0.10 & 0.12 $\pm$ 0.10 & 0.07 $\pm$ 0.00 & \underline{0.18 $\pm$ 0.11} & \underline{0.18 $\pm$ 0.11} & \underline{0.18 $\pm$ 0.11} & \underline{0.18 $\pm$ 0.11} & \textbf{0.23 $\pm$ 0.12} & \textbf{0.23 $\pm$ 0.12} \\
\midrule
\multirow{3}{*}{10} & 0 & 0.20 $\pm$ 0.11 & 0.04 $\pm$ 0.05 & 0.23 $\pm$ 0.11 & 0.19 $\pm$ 0.11 & 0.06 $\pm$ 0.00 & \underline{0.27 $\pm$ 0.12} & \underline{0.27 $\pm$ 0.12} & 0.26 $\pm$ 0.12 & 0.26 $\pm$ 0.12 & \textbf{0.32 $\pm$ 0.13} & \textbf{0.32 $\pm$ 0.13} \\
 & 1 & 0.20 $\pm$ 0.11 & 0.04 $\pm$ 0.05 & 0.20 $\pm$ 0.10 & 0.17 $\pm$ 0.11 & 0.07 $\pm$ 0.00 & 0.26 $\pm$ 0.12 & 0.26 $\pm$ 0.12 & 0.26 $\pm$ 0.12 & 0.26 $\pm$ 0.12 & \textbf{0.31 $\pm$ 0.13} & \underline{0.30 $\pm$ 0.13} \\
 & 2 & 0.20 $\pm$ 0.11 & 0.03 $\pm$ 0.05 & 0.19 $\pm$ 0.10 & 0.17 $\pm$ 0.11 & 0.07 $\pm$ 0.00 & \underline{0.26 $\pm$ 0.12} & \underline{0.26 $\pm$ 0.12} & \underline{0.26 $\pm$ 0.12} & \underline{0.26 $\pm$ 0.12} & \textbf{0.30 $\pm$ 0.13} & \textbf{0.30 $\pm$ 0.13} \\
\midrule
\multirow{3}{*}{15} & 0 & 0.23 $\pm$ 0.11 & 0.05 $\pm$ 0.05 & 0.27 $\pm$ 0.11 & 0.21 $\pm$ 0.11 & 0.06 $\pm$ 0.00 & 0.31 $\pm$ 0.12 & 0.30 $\pm$ 0.12 & 0.30 $\pm$ 0.12 & 0.30 $\pm$ 0.12 & \textbf{0.35 $\pm$ 0.13} & \underline{0.34 $\pm$ 0.13} \\
 & 1 & 0.23 $\pm$ 0.10 & 0.04 $\pm$ 0.05 & 0.24 $\pm$ 0.10 & 0.19 $\pm$ 0.11 & 0.07 $\pm$ 0.00 & \underline{0.30 $\pm$ 0.11} & \underline{0.30 $\pm$ 0.11} & 0.29 $\pm$ 0.12 & \underline{0.30 $\pm$ 0.11} & \textbf{0.33 $\pm$ 0.12} & \textbf{0.33 $\pm$ 0.12} \\
 & 2 & 0.23 $\pm$ 0.11 & 0.04 $\pm$ 0.05 & 0.22 $\pm$ 0.10 & 0.18 $\pm$ 0.11 & 0.07 $\pm$ 0.00 & 0.29 $\pm$ 0.11 & \underline{0.30 $\pm$ 0.11} & 0.29 $\pm$ 0.12 & \underline{0.30 $\pm$ 0.11} & \textbf{0.33 $\pm$ 0.12} & \textbf{0.33 $\pm$ 0.12} \\
\midrule
\multirow{3}{*}{20} & 0 & 0.25 $\pm$ 0.10 & 0.06 $\pm$ 0.05 & 0.28 $\pm$ 0.11 & 0.21 $\pm$ 0.11 & 0.06 $\pm$ 0.00 & 0.32 $\pm$ 0.11 & 0.32 $\pm$ 0.11 & 0.32 $\pm$ 0.12 & 0.32 $\pm$ 0.11 & \textbf{0.36 $\pm$ 0.12} & \underline{0.35 $\pm$ 0.12} \\
 & 1 & 0.24 $\pm$ 0.10 & 0.05 $\pm$ 0.05 & 0.26 $\pm$ 0.10 & 0.19 $\pm$ 0.10 & 0.07 $\pm$ 0.00 & 0.31 $\pm$ 0.11 & 0.31 $\pm$ 0.11 & 0.31 $\pm$ 0.11 & \underline{0.32 $\pm$ 0.11} & \textbf{0.34 $\pm$ 0.12} & \textbf{0.34 $\pm$ 0.12} \\
 & 2 & 0.24 $\pm$ 0.10 & 0.05 $\pm$ 0.05 & 0.25 $\pm$ 0.10 & 0.19 $\pm$ 0.10 & 0.07 $\pm$ 0.00 & \underline{0.31 $\pm$ 0.11} & \underline{0.31 $\pm$ 0.11} & \underline{0.31 $\pm$ 0.11} & \underline{0.31 $\pm$ 0.11} & \textbf{0.34 $\pm$ 0.12} & \textbf{0.34 $\pm$ 0.12} \\
\midrule
\end{tabular}}

\end{table*}

\begin{figure}[ht]
    \centering
    \includegraphics[width=\linewidth]{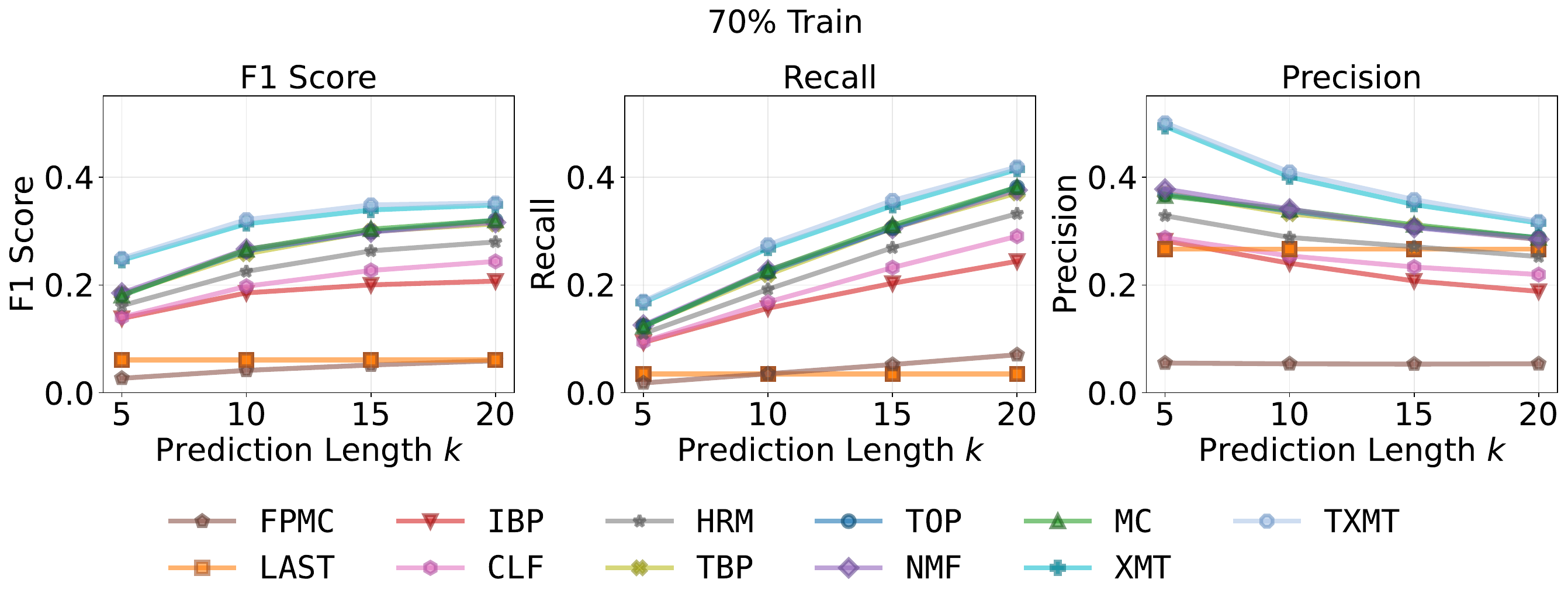}
    \caption{Comparison of \xmt and \txmt with Baseline methods using a \textit{70\% Train} and $h$=0}
    \label{fig:coop_70_day_0}
\end{figure}

\begin{figure}[ht]
    \centering
    \includegraphics[width=\linewidth]{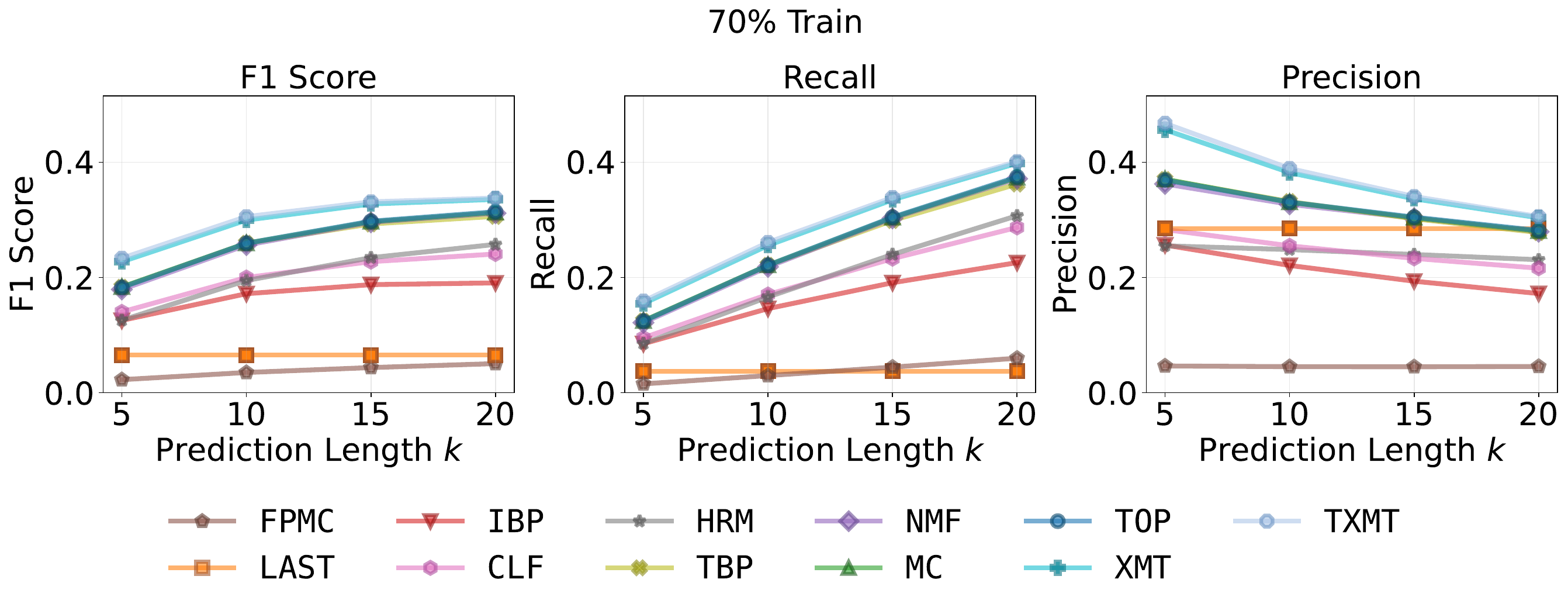}
    \caption{Comparison of \xmt and \txmt with Baseline methods using a \textit{70\% Train} and $h$=1}
    \label{fig:coop_70_day_1}
\end{figure}

%70% SPLIT
Figures~\ref{fig:coop_70_day_0}, and~\ref{fig:coop_70_day_1} presents the result with \textit{30\% Train} and $h = 0$ and $h = 1$ respectively. Tables~\ref{tab:comparison_precision_70},~\ref{tab:comparison_recall_70} and~\ref{tab:comparison_recall_70} report respectively the precision, recall and F1 Score with \textit{20\% Train}. This is one of the settings we considered in~\Cref{sec:exp_results}, here we offer a more in-depth view of the results, exploring how they change when we decrease the $h$.

\begin{table*}[ht]
                \caption{Precision comparison for different prediction lengths k and days $h$ with \textit{70\% Train}}
                    \label{tab:comparison_precision_70}
                \vspace{0.5cm}
                    \centering
                    \small
                    \setlength{\tabcolsep}{4pt}
                    \resizebox{\linewidth}{!}{
\begin{tabular}{cc>{\columncolor{gray!20}}cc>{\columncolor{gray!20}}cc>{\columncolor{gray!20}}cc>{\columncolor{gray!20}}cc>{\columncolor{gray!20}}cc>{\columncolor{gray!20}}c}
                    \hline
                    \multicolumn{13}{c}{70\% Train} \\
                    \hline
                    \textbf{k} & \textbf{$h$} & \textbf{clf} & \textbf{fpmc} & \textbf{hrm} & \textbf{ibp} & \textbf{last} & \textbf{mc} & \textbf{nmf} & \textbf{tbp} & \textbf{top} & \textbf{txmt} & \textbf{xmt} \\
                    \hline
                    \multirow{3}{*}{5} & 0 & 0.29 $\pm$ 0.21 & 0.06 $\pm$ 0.10 & 0.33 $\pm$ 0.22 & 0.28 $\pm$ 0.21 & 0.27 $\pm$ 0.00 & 0.37 $\pm$ 0.23 & 0.38 $\pm$ 0.23 & 0.37 $\pm$ 0.24 & 0.37 $\pm$ 0.23 & \textbf{0.50 $\pm$ 0.26} & \underline{0.49 $\pm$ 0.26} \\
                     & 1 & 0.28 $\pm$ 0.21 & 0.05 $\pm$ 0.09 & 0.25 $\pm$ 0.\multirow{3}{*}{20} & 0.26 $\pm$ 0.21 & 0.28 $\pm$ 0.00 & 0.37 $\pm$ 0.23 & 0.36 $\pm$ 0.23 & 0.37 $\pm$ 0.23 & 0.37 $\pm$ 0.23 & \textbf{0.47 $\pm$ 0.24} & \underline{0.46 $\pm$ 0.24} \\
                     & 2 & 0.29 $\pm$ 0.21 & 0.05 $\pm$ 0.10 & 0.24 $\pm$ 0.19 & 0.26 $\pm$ 0.21 & 0.29 $\pm$ 0.00 & 0.37 $\pm$ 0.23 & 0.37 $\pm$ 0.23 & 0.37 $\pm$ 0.23 & 0.37 $\pm$ 0.23 & \textbf{0.47 $\pm$ 0.25} & \underline{0.46 $\pm$ 0.24} \\
                    \midrule
                    \multirow{3}{*}{10} & 0 & 0.25 $\pm$ 0.16 & 0.05 $\pm$ 0.07 & 0.29 $\pm$ 0.16 & 0.24 $\pm$ 0.16 & 0.27 $\pm$ 0.00 & 0.34 $\pm$ 0.17 & 0.34 $\pm$ 0.17 & 0.33 $\pm$ 0.18 & 0.34 $\pm$ 0.17 & \textbf{0.41 $\pm$ 0.19} & \underline{0.40 $\pm$ 0.19} \\
                     & 1 & 0.25 $\pm$ 0.15 & 0.05 $\pm$ 0.07 & 0.25 $\pm$ 0.14 & 0.22 $\pm$ 0.15 & 0.28 $\pm$ 0.00 & 0.33 $\pm$ 0.17 & 0.33 $\pm$ 0.16 & 0.33 $\pm$ 0.17 & 0.33 $\pm$ 0.17 & \textbf{0.39 $\pm$ 0.18} & \underline{0.38 $\pm$ 0.18} \\
                     & 2 & 0.26 $\pm$ 0.15 & 0.05 $\pm$ 0.07 & 0.23 $\pm$ 0.14 & 0.22 $\pm$ 0.15 & 0.29 $\pm$ 0.00 & \underline{0.34 $\pm$ 0.17} & \underline{0.34 $\pm$ 0.16} & \underline{0.34 $\pm$ 0.17} & \underline{0.34 $\pm$ 0.17} & \textbf{0.39 $\pm$ 0.18} & \textbf{0.39 $\pm$ 0.18} \\
                    \midrule
                    \multirow{3}{*}{15} & 0 & 0.23 $\pm$ 0.13 & 0.05 $\pm$ 0.06 & 0.27 $\pm$ 0.13 & 0.21 $\pm$ 0.13 & 0.27 $\pm$ 0.00 & 0.31 $\pm$ 0.14 & 0.31 $\pm$ 0.14 & 0.31 $\pm$ 0.14 & 0.31 $\pm$ 0.14 & \textbf{0.36 $\pm$ 0.15} & \underline{0.35 $\pm$ 0.15} \\
                     & 1 & 0.23 $\pm$ 0.12 & 0.05 $\pm$ 0.05 & 0.24 $\pm$ 0.12 & 0.19 $\pm$ 0.12 & 0.28 $\pm$ 0.00 & \underline{0.30 $\pm$ 0.14} & \underline{0.30 $\pm$ 0.13} & \underline{0.30 $\pm$ 0.14} & \underline{0.30 $\pm$ 0.14} & \textbf{0.34 $\pm$ 0.15} & \textbf{0.34 $\pm$ 0.15} \\
                     & 2 & 0.24 $\pm$ 0.12 & 0.05 $\pm$ 0.06 & 0.23 $\pm$ 0.11 & 0.19 $\pm$ 0.12 & 0.29 $\pm$ 0.00 & 0.30 $\pm$ 0.14 & \underline{0.31 $\pm$ 0.14} & 0.30 $\pm$ 0.14 & \underline{0.31 $\pm$ 0.14} & \textbf{0.34 $\pm$ 0.15} & \textbf{0.34 $\pm$ 0.15} \\
                    \midrule
                    \multirow{3}{*}{20} & 0 & 0.22 $\pm$ 0.11 & 0.05 $\pm$ 0.05 & 0.25 $\pm$ 0.12 & 0.19 $\pm$ 0.11 & 0.27 $\pm$ 0.00 & \underline{0.29 $\pm$ 0.12} & 0.28 $\pm$ 0.12 & 0.28 $\pm$ 0.13 & \underline{0.29 $\pm$ 0.12} & \textbf{0.32 $\pm$ 0.13} & \textbf{0.32 $\pm$ 0.13} \\
                     & 1 & 0.22 $\pm$ 0.11 & 0.05 $\pm$ 0.05 & 0.23 $\pm$ 0.10 & 0.17 $\pm$ 0.10 & 0.28 $\pm$ 0.00 & 0.28 $\pm$ 0.12 & 0.28 $\pm$ 0.12 & 0.28 $\pm$ 0.12 & 0.28 $\pm$ 0.12 & \textbf{0.31 $\pm$ 0.13} & \underline{0.30 $\pm$ 0.13} \\
                     & 2 & 0.22 $\pm$ 0.11 & 0.05 $\pm$ 0.05 & 0.22 $\pm$ 0.10 & 0.17 $\pm$ 0.11 & \underline{0.29 $\pm$ 0.00} & 0.28 $\pm$ 0.12 & 0.28 $\pm$ 0.12 & 0.28 $\pm$ 0.12 & 0.28 $\pm$ 0.12 & \textbf{0.31 $\pm$ 0.13} & \textbf{0.31 $\pm$ 0.13} \\
                    \midrule
                    \end{tabular}}
                    
                    \end{table*}

\begin{table*}[ht]
            \caption{Recall comparison for different prediction lengths (k) and days (d) with \textit{70\% Train}}
            \label{tab:comparison_recall_70}
            \vspace{0.5cm}
            \centering
            \small
            \setlength{\tabcolsep}{4pt}
            \resizebox{\linewidth}{!}{
            \begin{tabular}{cc>{\columncolor{gray!20}}cc>{\columncolor{gray!20}}cc>{\columncolor{gray!20}}cc>{\columncolor{gray!20}}cc>{\columncolor{gray!20}}cc>{\columncolor{gray!20}}c}
            \hline
            \multicolumn{13}{c}{70\% Train} \\
            \hline
            \textbf{k} & \textbf{$h$} & \textbf{clf} & \textbf{fpmc} & \textbf{hrm} & \textbf{ibp} & \textbf{last} & \textbf{mc} & \textbf{nmf} & \textbf{tbp} & \textbf{top} & \textbf{txmt} & \textbf{xmt} \\
            \hline
            \multirow{3}{*}{5} & 0 & 0.09 $\pm$ 0.07 & 0.02 $\pm$ 0.04 & 0.11 $\pm$ 0.08 & 0.09 $\pm$ 0.07 & 0.03 $\pm$ 0.00 & 0.12 $\pm$ 0.08 & \underline{0.13 $\pm$ 0.08} & 0.12 $\pm$ 0.08 & 0.12 $\pm$ 0.08 & \textbf{0.17 $\pm$ 0.10} & \textbf{0.17 $\pm$ 0.10} \\
             & 1 & 0.10 $\pm$ 0.07 & 0.02 $\pm$ 0.03 & 0.09 $\pm$ 0.07 & 0.09 $\pm$ 0.07 & 0.04 $\pm$ 0.00 & 0.12 $\pm$ 0.08 & 0.12 $\pm$ 0.08 & 0.12 $\pm$ 0.08 & 0.12 $\pm$ 0.08 & \textbf{0.16 $\pm$ 0.09} & \underline{0.15 $\pm$ 0.09} \\
             & 2 & 0.09 $\pm$ 0.07 & 0.02 $\pm$ 0.03 & 0.08 $\pm$ 0.07 & 0.08 $\pm$ 0.07 & 0.04 $\pm$ 0.00 & 0.12 $\pm$ 0.08 & 0.12 $\pm$ 0.08 & 0.12 $\pm$ 0.08 & 0.12 $\pm$ 0.08 & \textbf{0.16 $\pm$ 0.09} & \underline{0.15 $\pm$ 0.09} \\
            \midrule
            \multirow{3}{*}{10} & 0 & 0.17 $\pm$ 0.10 & 0.04 $\pm$ 0.05 & 0.19 $\pm$ 0.10 & 0.16 $\pm$ 0.10 & 0.03 $\pm$ 0.00 & \underline{0.23 $\pm$ 0.11} & \underline{0.23 $\pm$ 0.11} & 0.22 $\pm$ 0.11 & 0.22 $\pm$ 0.11 & \textbf{0.27 $\pm$ 0.13} & \textbf{0.27 $\pm$ 0.12} \\
             & 1 & 0.17 $\pm$ 0.10 & 0.03 $\pm$ 0.04 & 0.17 $\pm$ 0.09 & 0.15 $\pm$ 0.10 & 0.04 $\pm$ 0.00 & \underline{0.22 $\pm$ 0.11} & \underline{0.22 $\pm$ 0.10} & \underline{0.22 $\pm$ 0.11} & \underline{0.22 $\pm$ 0.10} & \textbf{0.26 $\pm$ 0.12} & \textbf{0.26 $\pm$ 0.12} \\
             & 2 & 0.17 $\pm$ 0.09 & 0.03 $\pm$ 0.04 & 0.15 $\pm$ 0.09 & 0.14 $\pm$ 0.09 & 0.04 $\pm$ 0.00 & 0.22 $\pm$ 0.10 & 0.22 $\pm$ 0.10 & 0.22 $\pm$ 0.11 & 0.22 $\pm$ 0.10 & \textbf{0.26 $\pm$ 0.11} & \underline{0.25 $\pm$ 0.11} \\
            \midrule
            \multirow{3}{*}{15} & 0 & 0.23 $\pm$ 0.12 & 0.05 $\pm$ 0.06 & 0.27 $\pm$ 0.12 & 0.20 $\pm$ 0.12 & 0.03 $\pm$ 0.00 & 0.31 $\pm$ 0.13 & 0.30 $\pm$ 0.12 & 0.31 $\pm$ 0.13 & 0.31 $\pm$ 0.13 & \textbf{0.36 $\pm$ 0.14} & \underline{0.35 $\pm$ 0.14} \\
             & 1 & 0.23 $\pm$ 0.11 & 0.04 $\pm$ 0.05 & 0.24 $\pm$ 0.11 & 0.19 $\pm$ 0.11 & 0.04 $\pm$ 0.00 & 0.30 $\pm$ 0.12 & 0.30 $\pm$ 0.12 & 0.30 $\pm$ 0.12 & 0.30 $\pm$ 0.12 & \textbf{0.34 $\pm$ 0.13} & \underline{0.33 $\pm$ 0.13} \\
             & 2 & 0.23 $\pm$ 0.11 & 0.04 $\pm$ 0.05 & 0.22 $\pm$ 0.10 & 0.18 $\pm$ 0.11 & 0.04 $\pm$ 0.00 & \underline{0.30 $\pm$ 0.12} & \underline{0.30 $\pm$ 0.12} & 0.29 $\pm$ 0.12 & \underline{0.30 $\pm$ 0.12} & \textbf{0.33 $\pm$ 0.13} & \textbf{0.33 $\pm$ 0.13} \\
            \midrule
            \multirow{3}{*}{20} & 0 & 0.29 $\pm$ 0.13 & 0.07 $\pm$ 0.07 & 0.33 $\pm$ 0.13 & 0.24 $\pm$ 0.13 & 0.03 $\pm$ 0.00 & 0.38 $\pm$ 0.14 & 0.38 $\pm$ 0.13 & 0.37 $\pm$ 0.14 & 0.38 $\pm$ 0.13 & \textbf{0.42 $\pm$ 0.15} & \underline{0.41 $\pm$ 0.15} \\
             & 1 & 0.29 $\pm$ 0.12 & 0.06 $\pm$ 0.06 & 0.31 $\pm$ 0.12 & 0.23 $\pm$ 0.12 & 0.04 $\pm$ 0.00 & \underline{0.37 $\pm$ 0.13} & \underline{0.37 $\pm$ 0.13} & 0.36 $\pm$ 0.13 & \underline{0.37 $\pm$ 0.13} & \textbf{0.40 $\pm$ 0.14} & \textbf{0.40 $\pm$ 0.14} \\
             & 2 & 0.29 $\pm$ 0.12 & 0.06 $\pm$ 0.06 & 0.29 $\pm$ 0.12 & 0.22 $\pm$ 0.12 & 0.04 $\pm$ 0.00 & \underline{0.37 $\pm$ 0.13} & \underline{0.37 $\pm$ 0.13} & 0.35 $\pm$ 0.13 & \underline{0.37 $\pm$ 0.13} & \textbf{0.39 $\pm$ 0.14} & \textbf{0.39 $\pm$ 0.14} \\
            \midrule
            \end{tabular}
            }
            
            \end{table*}

\begin{table*}[ht]
\caption{F1 Score comparison for different prediction lengths (k) and days (d)  with \textit{70\% Train}}
\label{tab:comparison_f1_score_70}
\vspace{0.5cm}
\centering
\small
\setlength{\tabcolsep}{4pt}
\resizebox{\linewidth}{!}{
\begin{tabular}{cc>{\columncolor{gray!20}}cc>{\columncolor{gray!20}}cc>{\columncolor{gray!20}}cc>{\columncolor{gray!20}}cc>{\columncolor{gray!20}}cc>{\columncolor{gray!20}}c}
\hline
\multicolumn{13}{c}{70\% Train} \\
\hline
\textbf{k} & \textbf{$h$} & \textbf{clf} & \textbf{fpmc} & \textbf{hrm} & \textbf{ibp} & \textbf{last} & \textbf{mc} & \textbf{nmf} & \textbf{tbp} & \textbf{top} & \textbf{txmt} & \textbf{xmt} \\
\hline
\multirow{3}{*}{5} & 0 & 0.14 $\pm$ 0.10 & 0.03 $\pm$ 0.05 & 0.16 $\pm$ 0.11 & 0.14 $\pm$ 0.10 & 0.06 $\pm$ 0.00 & 0.18 $\pm$ 0.11 & \underline{0.19 $\pm$ 0.11} & 0.18 $\pm$ 0.12 & 0.18 $\pm$ 0.11 & \textbf{0.25 $\pm$ 0.13} & \textbf{0.25 $\pm$ 0.13} \\
 & 1 & 0.14 $\pm$ 0.10 & 0.02 $\pm$ 0.05 & 0.13 $\pm$ 0.10 & 0.13 $\pm$ 0.10 & 0.07 $\pm$ 0.00 & \underline{0.18 $\pm$ 0.11} & \underline{0.18 $\pm$ 0.11} & \underline{0.18 $\pm$ 0.11} & \underline{0.18 $\pm$ 0.11} & \textbf{0.23 $\pm$ 0.12} & \textbf{0.23 $\pm$ 0.12} \\
 & 2 & 0.14 $\pm$ 0.10 & 0.02 $\pm$ 0.05 & 0.12 $\pm$ 0.09 & 0.12 $\pm$ 0.10 & 0.07 $\pm$ 0.00 & \underline{0.18 $\pm$ 0.11} & \underline{0.18 $\pm$ 0.11} & \underline{0.18 $\pm$ 0.11} & \underline{0.18 $\pm$ 0.11} & \textbf{0.23 $\pm$ 0.12} & \textbf{0.23 $\pm$ 0.12} \\
\midrule
\multirow{3}{*}{10} & 0 & 0.20 $\pm$ 0.11 & 0.04 $\pm$ 0.06 & 0.23 $\pm$ 0.11 & 0.19 $\pm$ 0.11 & 0.06 $\pm$ 0.00 & 0.27 $\pm$ 0.12 & 0.27 $\pm$ 0.12 & 0.26 $\pm$ 0.12 & 0.26 $\pm$ 0.12 & \textbf{0.32 $\pm$ 0.14} & \underline{0.31 $\pm$ 0.13} \\
 & 1 & 0.20 $\pm$ 0.11 & 0.04 $\pm$ 0.05 & 0.19 $\pm$ 0.10 & 0.17 $\pm$ 0.11 & 0.07 $\pm$ 0.00 & 0.26 $\pm$ 0.12 & 0.26 $\pm$ 0.12 & 0.26 $\pm$ 0.12 & 0.26 $\pm$ 0.12 & \textbf{0.31 $\pm$ 0.13} & \underline{0.30 $\pm$ 0.13} \\
 & 2 & 0.20 $\pm$ 0.11 & 0.04 $\pm$ 0.05 & 0.18 $\pm$ 0.10 & 0.17 $\pm$ 0.11 & 0.07 $\pm$ 0.00 & \underline{0.26 $\pm$ 0.12} & \underline{0.26 $\pm$ 0.12} & \underline{0.26 $\pm$ 0.12} & \underline{0.26 $\pm$ 0.12} & \textbf{0.30 $\pm$ 0.13} & \textbf{0.30 $\pm$ 0.13} \\
\midrule
\multirow{3}{*}{15} & 0 & 0.23 $\pm$ 0.11 & 0.05 $\pm$ 0.06 & 0.26 $\pm$ 0.11 & 0.20 $\pm$ 0.11 & 0.06 $\pm$ 0.00 & 0.30 $\pm$ 0.12 & 0.30 $\pm$ 0.12 & 0.30 $\pm$ 0.12 & 0.30 $\pm$ 0.12 & \textbf{0.35 $\pm$ 0.13} & \underline{0.34 $\pm$ 0.13} \\
 & 1 & 0.23 $\pm$ 0.10 & 0.04 $\pm$ 0.05 & 0.23 $\pm$ 0.10 & 0.19 $\pm$ 0.11 & 0.07 $\pm$ 0.00 & \underline{0.30 $\pm$ 0.11} & \underline{0.30 $\pm$ 0.11} & 0.29 $\pm$ 0.12 & \underline{0.30 $\pm$ 0.11} & \textbf{0.33 $\pm$ 0.12} & \textbf{0.33 $\pm$ 0.12} \\
 & 2 & 0.23 $\pm$ 0.10 & 0.04 $\pm$ 0.05 & 0.22 $\pm$ 0.10 & 0.18 $\pm$ 0.11 & 0.07 $\pm$ 0.00 & 0.29 $\pm$ 0.11 & \underline{0.30 $\pm$ 0.11} & 0.29 $\pm$ 0.12 & \underline{0.30 $\pm$ 0.11} & \textbf{0.33 $\pm$ 0.12} & \textbf{0.33 $\pm$ 0.12} \\
\midrule
\multirow{3}{*}{20} & 0 & 0.24 $\pm$ 0.10 & 0.06 $\pm$ 0.06 & 0.28 $\pm$ 0.11 & 0.21 $\pm$ 0.11 & 0.06 $\pm$ 0.00 & \underline{0.32 $\pm$ 0.11} & \underline{0.32 $\pm$ 0.11} & 0.31 $\pm$ 0.12 & \underline{0.32 $\pm$ 0.11} & \textbf{0.35 $\pm$ 0.12} & \textbf{0.35 $\pm$ 0.12} \\
 & 1 & 0.24 $\pm$ 0.10 & 0.05 $\pm$ 0.05 & 0.26 $\pm$ 0.10 & 0.19 $\pm$ 0.10 & 0.07 $\pm$ 0.00 & \underline{0.31 $\pm$ 0.11} & \underline{0.31 $\pm$ 0.11} & \underline{0.31 $\pm$ 0.11} & \underline{0.31 $\pm$ 0.11} & \textbf{0.34 $\pm$ 0.12} & \textbf{0.34 $\pm$ 0.12} \\
 & 2 & 0.24 $\pm$ 0.10 & 0.05 $\pm$ 0.05 & 0.25 $\pm$ 0.10 & 0.19 $\pm$ 0.10 & 0.07 $\pm$ 0.00 & 0.31 $\pm$ 0.11 & 0.31 $\pm$ 0.11 & 0.30 $\pm$ 0.11 & 0.31 $\pm$ 0.11 & \textbf{0.34 $\pm$ 0.12} & \underline{0.33 $\pm$ 0.12} \\
\midrule
\end{tabular}}

\end{table*}

\begin{figure}[ht]
    \centering
    \includegraphics[width=\linewidth]{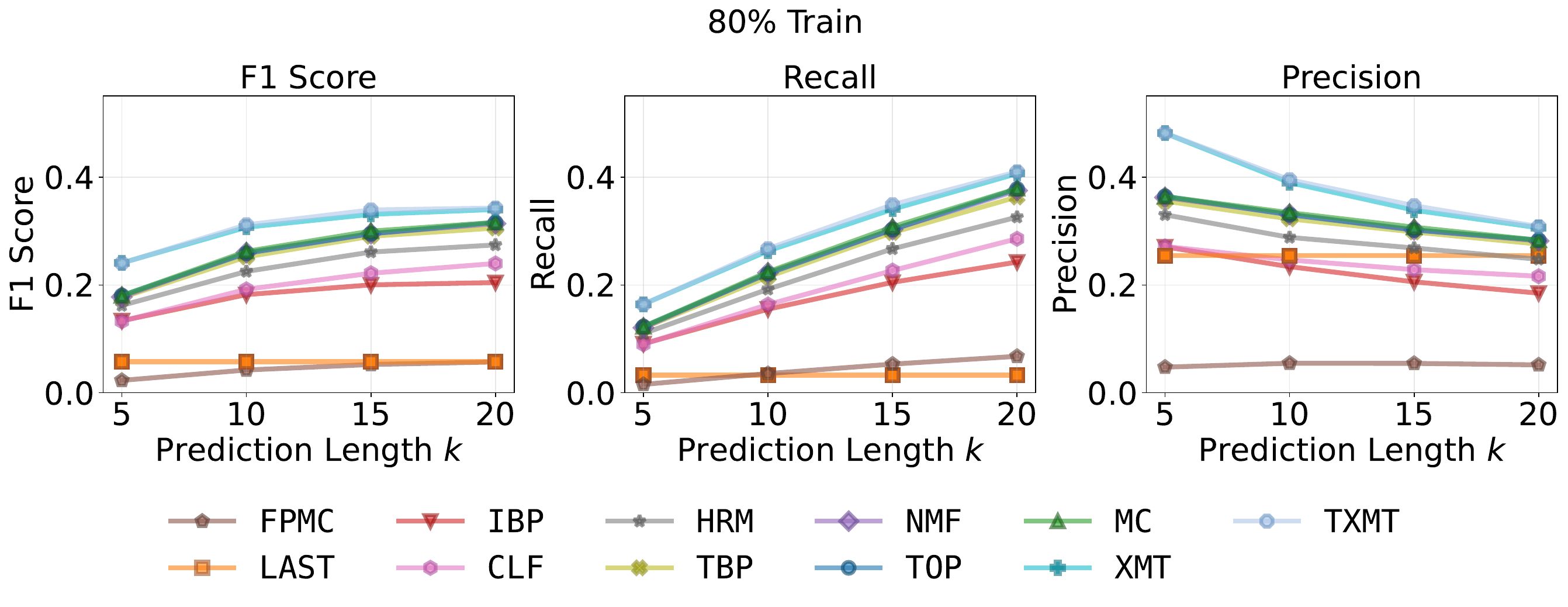}
    \caption{Comparison of \xmt and \txmt with Baseline methods using a \textit{80\% Train} and $h$=0}
    \label{fig:coop_80_day_0}
\end{figure}

\begin{figure}[ht]
    \centering
    \includegraphics[width=\linewidth]{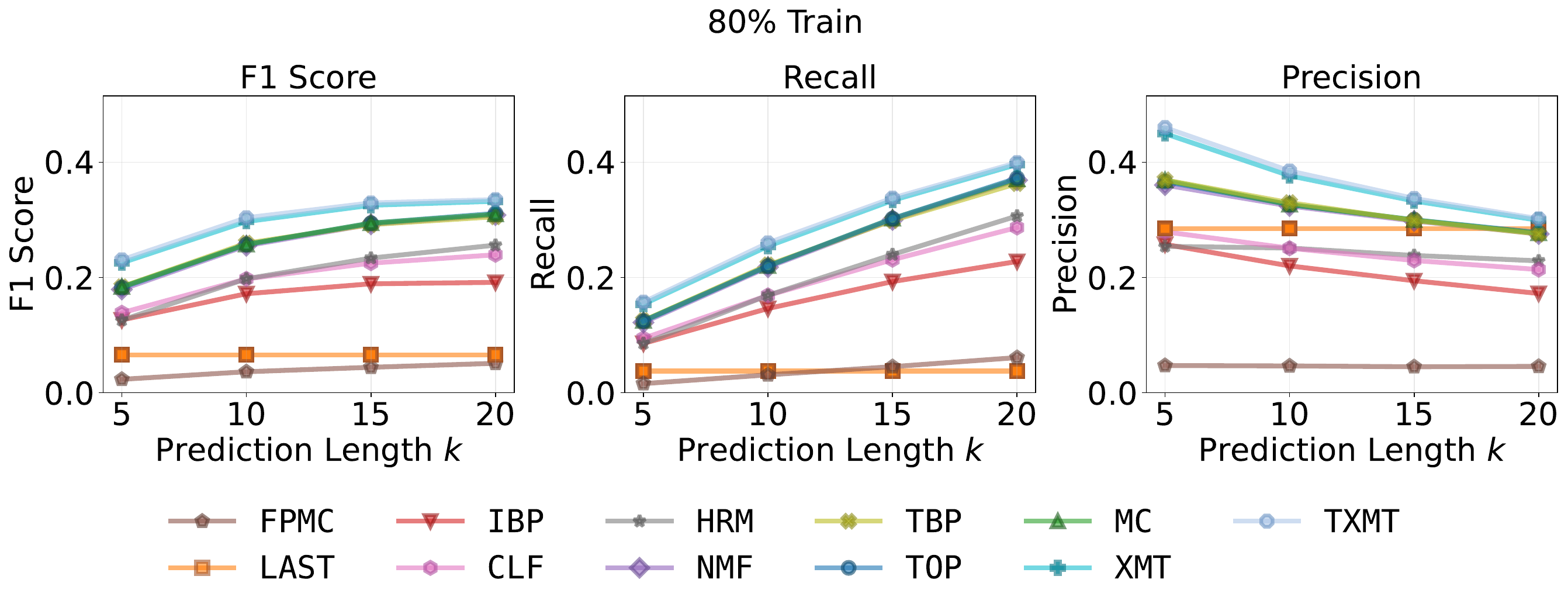}
    \caption{Comparison of \xmt and \txmt with Baseline methods using a \textit{80\% Train} and $h$=1}
    \label{fig:coop_80_day_1}
\end{figure}

\begin{figure}[ht]
    \centering
    \includegraphics[width=\linewidth]{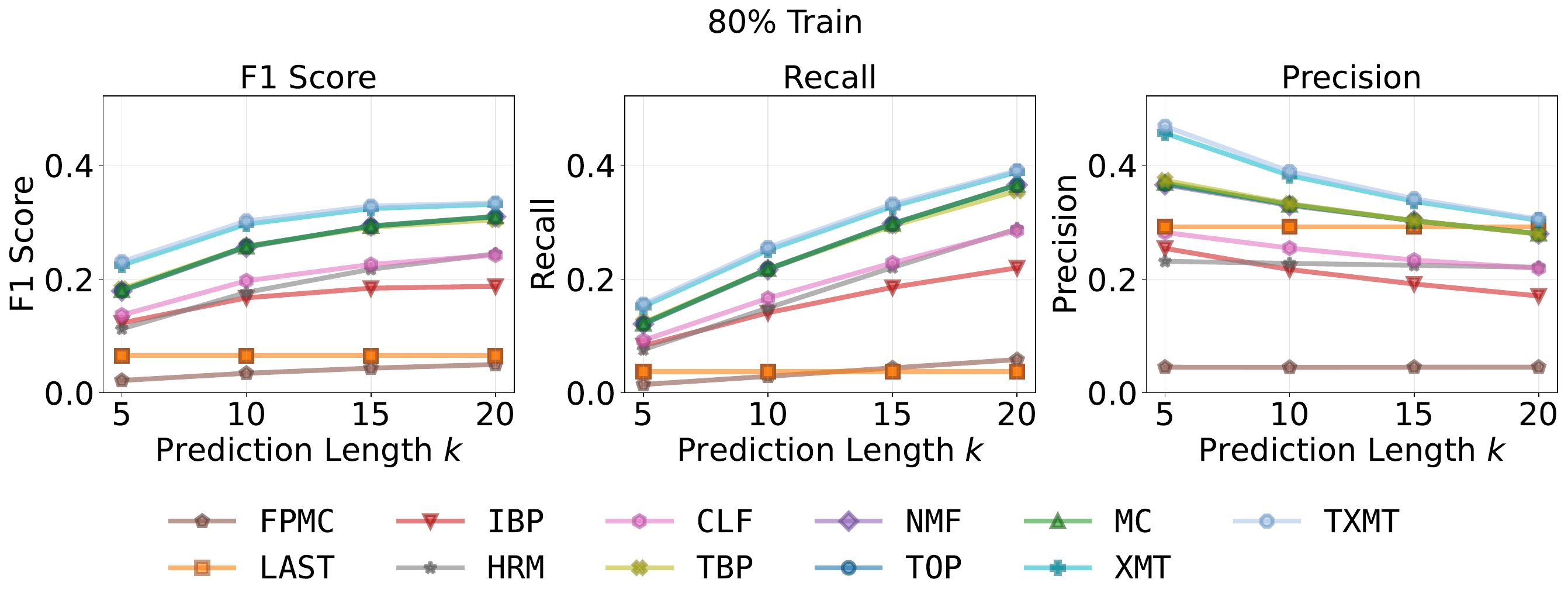}
    \caption{Comparison of \xmt and \txmt with Baseline methods using a \textit{80\% Train} and $h$=2}
    \label{fig:coop_80_day_2}
\end{figure}

%80% SPLIT
Lastly, figures~\ref{fig:coop_80_day_0}, ~\ref{fig:coop_80_day_1} and~\ref{fig:coop_80_day_2} presents the result with \textit{80\% Train} and $h = 0$, $h = 1$ and  $h = 2$ respectively. Tables~\ref{tab:comparison_precision_80},~\ref{tab:comparison_recall_80} and~\ref{tab:comparison_recall_80} report respectively the precision, recall and F1 Score with \textit{80\% Train}. This is the most extreme case that we considered in terms of training dataset size. However, the results are coherent and not dissimilar to the ones obtained when using smaller training datasets.


\begin{table*}[ht]
                    \caption{Precision comparison for different prediction lengths k and days $h$ with \textit{80\% Train}}
                        \label{tab:comparison_precision_80}
                    \vspace{0.5cm}
                        \centering
                        \small
                        \setlength{\tabcolsep}{4pt}
                        \resizebox{\linewidth}{!}{
\begin{tabular}{cc>{\columncolor{gray!20}}cc>{\columncolor{gray!20}}cc>{\columncolor{gray!20}}cc>{\columncolor{gray!20}}cc>{\columncolor{gray!20}}cc>{\columncolor{gray!20}}c}
                        \hline
                        \multicolumn{13}{c}{80\% Train} \\
                        \hline
                        \textbf{k} & \textbf{$h$} & \textbf{clf} & \textbf{fpmc} & \textbf{hrm} & \textbf{ibp} & \textbf{last} & \textbf{mc} & \textbf{nmf} & \textbf{tbp} & \textbf{top} & \textbf{txmt} & \textbf{xmt} \\
                        \hline
                        \multirow{3}{*}{5} & 0 & 0.27 $\pm$ 0.21 & 0.05 $\pm$ 0.09 & 0.33 $\pm$ 0.22 & 0.27 $\pm$ 0.21 & 0.25 $\pm$ 0.00 & \underline{0.36 $\pm$ 0.23} & \underline{0.36 $\pm$ 0.22} & \underline{0.36 $\pm$ 0.22} & \underline{0.36 $\pm$ 0.23} & \textbf{0.48 $\pm$ 0.26} & \textbf{0.48 $\pm$ 0.26} \\
                         & 1 & 0.28 $\pm$ 0.21 & 0.05 $\pm$ 0.09 & 0.25 $\pm$ 0.\multirow{3}{*}{20} & 0.26 $\pm$ 0.21 & 0.28 $\pm$ 0.00 & 0.37 $\pm$ 0.23 & 0.36 $\pm$ 0.23 & 0.37 $\pm$ 0.23 & 0.37 $\pm$ 0.23 & \textbf{0.46 $\pm$ 0.24} & \underline{0.45 $\pm$ 0.24} \\
                         & 2 & 0.28 $\pm$ 0.21 & 0.05 $\pm$ 0.09 & 0.23 $\pm$ 0.19 & 0.25 $\pm$ 0.21 & 0.29 $\pm$ 0.00 & 0.37 $\pm$ 0.23 & 0.37 $\pm$ 0.23 & 0.37 $\pm$ 0.23 & 0.37 $\pm$ 0.23 & \textbf{0.47 $\pm$ 0.24} & \underline{0.46 $\pm$ 0.24} \\
                        \midrule
                        \multirow{3}{*}{10} & 0 & 0.25 $\pm$ 0.15 & 0.05 $\pm$ 0.07 & 0.29 $\pm$ 0.16 & 0.23 $\pm$ 0.15 & 0.25 $\pm$ 0.00 & 0.33 $\pm$ 0.17 & 0.33 $\pm$ 0.16 & 0.32 $\pm$ 0.17 & 0.33 $\pm$ 0.17 & \textbf{0.40 $\pm$ 0.18} & \underline{0.39 $\pm$ 0.18} \\
                         & 1 & 0.25 $\pm$ 0.15 & 0.05 $\pm$ 0.07 & 0.25 $\pm$ 0.14 & 0.22 $\pm$ 0.15 & 0.28 $\pm$ 0.00 & \underline{0.33 $\pm$ 0.16} & 0.32 $\pm$ 0.16 & \underline{0.33 $\pm$ 0.17} & \underline{0.33 $\pm$ 0.17} & \textbf{0.38 $\pm$ 0.18} & \textbf{0.38 $\pm$ 0.18} \\
                         & 2 & 0.25 $\pm$ 0.15 & 0.04 $\pm$ 0.07 & 0.23 $\pm$ 0.14 & 0.22 $\pm$ 0.15 & 0.29 $\pm$ 0.00 & 0.33 $\pm$ 0.17 & 0.33 $\pm$ 0.17 & 0.33 $\pm$ 0.17 & 0.33 $\pm$ 0.17 & \textbf{0.39 $\pm$ 0.18} & \underline{0.38 $\pm$ 0.18} \\
                        \midrule
                        \multirow{3}{*}{15} & 0 & 0.23 $\pm$ 0.12 & 0.05 $\pm$ 0.06 & 0.27 $\pm$ 0.13 & 0.21 $\pm$ 0.13 & 0.25 $\pm$ 0.00 & 0.31 $\pm$ 0.14 & 0.30 $\pm$ 0.14 & 0.30 $\pm$ 0.14 & 0.30 $\pm$ 0.14 & \textbf{0.35 $\pm$ 0.15} & \underline{0.34 $\pm$ 0.14} \\
                         & 1 & 0.23 $\pm$ 0.12 & 0.05 $\pm$ 0.05 & 0.24 $\pm$ 0.12 & 0.19 $\pm$ 0.12 & 0.28 $\pm$ 0.00 & 0.30 $\pm$ 0.14 & 0.30 $\pm$ 0.13 & 0.30 $\pm$ 0.14 & 0.30 $\pm$ 0.14 & \textbf{0.34 $\pm$ 0.15} & \underline{0.33 $\pm$ 0.15} \\
                         & 2 & 0.23 $\pm$ 0.12 & 0.05 $\pm$ 0.05 & 0.22 $\pm$ 0.11 & 0.19 $\pm$ 0.12 & 0.29 $\pm$ 0.00 & \underline{0.30 $\pm$ 0.14} & \underline{0.30 $\pm$ 0.13} & \underline{0.30 $\pm$ 0.14} & \underline{0.30 $\pm$ 0.14} & \textbf{0.34 $\pm$ 0.15} & \textbf{0.34 $\pm$ 0.15} \\
                        \midrule
                        \multirow{3}{*}{20} & 0 & 0.22 $\pm$ 0.10 & 0.05 $\pm$ 0.05 & 0.25 $\pm$ 0.11 & 0.18 $\pm$ 0.11 & 0.25 $\pm$ 0.00 & \underline{0.28 $\pm$ 0.12} & \underline{0.28 $\pm$ 0.12} & \underline{0.28 $\pm$ 0.12} & \underline{0.28 $\pm$ 0.12} & \textbf{0.31 $\pm$ 0.12} & \textbf{0.31 $\pm$ 0.12} \\
                         & 1 & 0.21 $\pm$ 0.10 & 0.05 $\pm$ 0.05 & 0.23 $\pm$ 0.10 & 0.17 $\pm$ 0.11 & \underline{0.28 $\pm$ 0.00} & \underline{0.28 $\pm$ 0.12} & \underline{0.28 $\pm$ 0.12} & \underline{0.28 $\pm$ 0.12} & \underline{0.28 $\pm$ 0.12} & \textbf{0.30 $\pm$ 0.13} & \textbf{0.30 $\pm$ 0.13} \\
                         & 2 & 0.22 $\pm$ 0.11 & 0.05 $\pm$ 0.05 & 0.22 $\pm$ 0.10 & 0.17 $\pm$ 0.10 & 0.29 $\pm$ 0.00 & 0.28 $\pm$ 0.12 & 0.28 $\pm$ 0.12 & 0.28 $\pm$ 0.12 & 0.28 $\pm$ 0.12 & \textbf{0.31 $\pm$ 0.13} & \underline{0.30 $\pm$ 0.13} \\
                        \midrule
                        \end{tabular}}
                        
                        \end{table*}
\begin{table*}[ht]
            \caption{Recall comparison for different prediction lengths (k) and days (d)  with \textit{80\% Train}}
                \label{tab:comparison_recall_80}
            \vspace{0.5cm}
                \centering
                \small
                \setlength{\tabcolsep}{4pt}
                \resizebox{\linewidth}{!}{
                \begin{tabular}{cc>{\columncolor{gray!20}}cc>{\columncolor{gray!20}}cc>{\columncolor{gray!20}}cc>{\columncolor{gray!20}}cc>{\columncolor{gray!20}}cc>{\columncolor{gray!20}}c}
                \hline
                \multicolumn{13}{c}{80\% Train} \\
                \hline
                \textbf{k} & \textbf{$h$} & \textbf{clf} & \textbf{fpmc} & \textbf{hrm} & \textbf{ibp} & \textbf{last} & \textbf{mc} & \textbf{nmf} & \textbf{tbp} & \textbf{top} & \textbf{txmt} & \textbf{xmt} \\
                \hline
                \multirow{3}{*}{5} & 0 & 0.09 $\pm$ 0.07 & 0.02 $\pm$ 0.03 & 0.11 $\pm$ 0.07 & 0.09 $\pm$ 0.07 & 0.03 $\pm$ 0.00 & \underline{0.12 $\pm$ 0.08} & \underline{0.12 $\pm$ 0.08} & \underline{0.12 $\pm$ 0.08} & \underline{0.12 $\pm$ 0.08} & \textbf{0.16 $\pm$ 0.09} & \textbf{0.16 $\pm$ 0.09} \\
                 & 1 & 0.09 $\pm$ 0.07 & 0.02 $\pm$ 0.03 & 0.09 $\pm$ 0.07 & 0.09 $\pm$ 0.07 & 0.04 $\pm$ 0.00 & 0.12 $\pm$ 0.08 & 0.12 $\pm$ 0.08 & 0.12 $\pm$ 0.08 & 0.12 $\pm$ 0.08 & \textbf{0.16 $\pm$ 0.09} & \underline{0.15 $\pm$ 0.09} \\
                 & 2 & 0.09 $\pm$ 0.07 & 0.01 $\pm$ 0.03 & 0.08 $\pm$ 0.06 & 0.08 $\pm$ 0.07 & 0.04 $\pm$ 0.00 & 0.12 $\pm$ 0.08 & 0.12 $\pm$ 0.08 & 0.12 $\pm$ 0.08 & 0.12 $\pm$ 0.08 & \textbf{0.16 $\pm$ 0.08} & \underline{0.15 $\pm$ 0.08} \\
                \midrule
                \multirow{3}{*}{10} & 0 & 0.16 $\pm$ 0.10 & 0.04 $\pm$ 0.05 & 0.19 $\pm$ 0.10 & 0.16 $\pm$ 0.10 & 0.03 $\pm$ 0.00 & 0.22 $\pm$ 0.11 & 0.22 $\pm$ 0.11 & 0.22 $\pm$ 0.10 & 0.22 $\pm$ 0.11 & \textbf{0.27 $\pm$ 0.13} & \underline{0.26 $\pm$ 0.12} \\
                 & 1 & 0.17 $\pm$ 0.10 & 0.03 $\pm$ 0.05 & 0.17 $\pm$ 0.09 & 0.15 $\pm$ 0.10 & 0.04 $\pm$ 0.00 & 0.22 $\pm$ 0.10 & 0.22 $\pm$ 0.10 & 0.22 $\pm$ 0.11 & 0.22 $\pm$ 0.10 & \textbf{0.26 $\pm$ 0.12} & \underline{0.25 $\pm$ 0.12} \\
                 & 2 & 0.17 $\pm$ 0.09 & 0.03 $\pm$ 0.04 & 0.15 $\pm$ 0.09 & 0.14 $\pm$ 0.09 & 0.04 $\pm$ 0.00 & 0.22 $\pm$ 0.10 & 0.22 $\pm$ 0.10 & 0.22 $\pm$ 0.10 & 0.22 $\pm$ 0.10 & \textbf{0.26 $\pm$ 0.11} & \underline{0.25 $\pm$ 0.11} \\
                \midrule
                \multirow{3}{*}{15} & 0 & 0.23 $\pm$ 0.11 & 0.05 $\pm$ 0.06 & 0.27 $\pm$ 0.11 & 0.20 $\pm$ 0.12 & 0.03 $\pm$ 0.00 & 0.31 $\pm$ 0.13 & 0.30 $\pm$ 0.12 & 0.30 $\pm$ 0.12 & 0.30 $\pm$ 0.13 & \textbf{0.35 $\pm$ 0.14} & \underline{0.34 $\pm$ 0.14} \\
                 & 1 & 0.23 $\pm$ 0.11 & 0.05 $\pm$ 0.05 & 0.24 $\pm$ 0.11 & 0.19 $\pm$ 0.11 & 0.04 $\pm$ 0.00 & 0.30 $\pm$ 0.12 & 0.30 $\pm$ 0.12 & 0.30 $\pm$ 0.12 & 0.30 $\pm$ 0.12 & \textbf{0.34 $\pm$ 0.13} & \underline{0.33 $\pm$ 0.13} \\
                 & 2 & 0.23 $\pm$ 0.11 & 0.04 $\pm$ 0.05 & 0.22 $\pm$ 0.10 & 0.19 $\pm$ 0.11 & 0.04 $\pm$ 0.00 & \underline{0.30 $\pm$ 0.12} & \underline{0.30 $\pm$ 0.12} & 0.29 $\pm$ 0.12 & \underline{0.30 $\pm$ 0.12} & \textbf{0.33 $\pm$ 0.13} & \textbf{0.33 $\pm$ 0.13} \\
                \midrule
                \multirow{3}{*}{20} & 0 & 0.29 $\pm$ 0.12 & 0.07 $\pm$ 0.06 & 0.33 $\pm$ 0.13 & 0.24 $\pm$ 0.13 & 0.03 $\pm$ 0.00 & \underline{0.38 $\pm$ 0.14} & \underline{0.38 $\pm$ 0.13} & 0.36 $\pm$ 0.14 & \underline{0.38 $\pm$ 0.14} & \textbf{0.41 $\pm$ 0.15} & \textbf{0.41 $\pm$ 0.15} \\
                 & 1 & 0.29 $\pm$ 0.12 & 0.06 $\pm$ 0.06 & 0.31 $\pm$ 0.12 & 0.23 $\pm$ 0.12 & 0.04 $\pm$ 0.00 & \underline{0.37 $\pm$ 0.13} & \underline{0.37 $\pm$ 0.13} & 0.36 $\pm$ 0.13 & \underline{0.37 $\pm$ 0.13} & \textbf{0.40 $\pm$ 0.14} & \textbf{0.40 $\pm$ 0.14} \\
                 & 2 & 0.29 $\pm$ 0.12 & 0.06 $\pm$ 0.06 & 0.29 $\pm$ 0.11 & 0.22 $\pm$ 0.12 & 0.04 $\pm$ 0.00 & \underline{0.37 $\pm$ 0.13} & \underline{0.37 $\pm$ 0.13} & 0.36 $\pm$ 0.13 & \underline{0.37 $\pm$ 0.13} & \textbf{0.39 $\pm$ 0.14} & \textbf{0.39 $\pm$ 0.14} \\
\midrule
                \end{tabular}
                }
                
                \end{table*}
                
\begin{table*}[ht]
\caption{F1 Score comparison for different prediction lengths (k) and days (d)  with \textit{80\% Train}}
\label{tab:comparison_f1_score_80}
\vspace{0.5cm}
\centering
\small
\setlength{\tabcolsep}{4pt}
\resizebox{\linewidth}{!}{
\begin{tabular}{cc>{\columncolor{gray!20}}cc>{\columncolor{gray!20}}cc>{\columncolor{gray!20}}cc>{\columncolor{gray!20}}cc>{\columncolor{gray!20}}cc>{\columncolor{gray!20}}c}\hline
\multicolumn{13}{c}{80\% Train} \\
\hline
\textbf{k} & \textbf{$h$} & \textbf{clf} & \textbf{fpmc} & \textbf{hrm} & \textbf{ibp} & \textbf{last} & \textbf{mc} & \textbf{nmf} & \textbf{tbp} & \textbf{top} & \textbf{txmt} & \textbf{xmt} \\
\hline
\multirow{3}{*}{5} & 0 & 0.13 $\pm$ 0.10 & 0.02 $\pm$ 0.05 & 0.16 $\pm$ 0.11 & 0.13 $\pm$ 0.10 & 0.06 $\pm$ 0.00 & \underline{0.18 $\pm$ 0.11} & \underline{0.18 $\pm$ 0.11} & \underline{0.18 $\pm$ 0.11} & \underline{0.18 $\pm$ 0.11} & \textbf{0.24 $\pm$ 0.13} & \textbf{0.24 $\pm$ 0.13} \\
 & 1 & 0.14 $\pm$ 0.10 & 0.02 $\pm$ 0.05 & 0.13 $\pm$ 0.10 & 0.13 $\pm$ 0.10 & 0.07 $\pm$ 0.00 & \underline{0.18 $\pm$ 0.11} & \underline{0.18 $\pm$ 0.11} & \underline{0.18 $\pm$ 0.11} & \underline{0.18 $\pm$ 0.11} & \textbf{0.23 $\pm$ 0.12} & \textbf{0.23 $\pm$ 0.12} \\
 & 2 & 0.14 $\pm$ 0.10 & 0.02 $\pm$ 0.05 & 0.11 $\pm$ 0.09 & 0.12 $\pm$ 0.10 & 0.07 $\pm$ 0.00 & 0.18 $\pm$ 0.11 & 0.18 $\pm$ 0.11 & 0.18 $\pm$ 0.11 & 0.18 $\pm$ 0.11 & \textbf{0.23 $\pm$ 0.12} & \underline{0.22 $\pm$ 0.12} \\
\midrule
\multirow{3}{*}{10} & 0 & 0.19 $\pm$ 0.11 & 0.04 $\pm$ 0.05 & 0.23 $\pm$ 0.11 & 0.18 $\pm$ 0.11 & 0.06 $\pm$ 0.00 & \underline{0.26 $\pm$ 0.12} & \underline{0.26 $\pm$ 0.12} & 0.25 $\pm$ 0.12 & \underline{0.26 $\pm$ 0.12} & \textbf{0.31 $\pm$ 0.14} & \textbf{0.31 $\pm$ 0.13} \\
 & 1 & 0.20 $\pm$ 0.11 & 0.04 $\pm$ 0.05 & 0.20 $\pm$ 0.10 & 0.17 $\pm$ 0.11 & 0.07 $\pm$ 0.00 & \underline{0.26 $\pm$ 0.12} & \underline{0.26 $\pm$ 0.12} & \underline{0.26 $\pm$ 0.12} & \underline{0.26 $\pm$ 0.12} & \textbf{0.30 $\pm$ 0.13} & \textbf{0.30 $\pm$ 0.13} \\
 & 2 & 0.20 $\pm$ 0.11 & 0.03 $\pm$ 0.05 & 0.18 $\pm$ 0.10 & 0.17 $\pm$ 0.11 & 0.07 $\pm$ 0.00 & \underline{0.26 $\pm$ 0.12} & \underline{0.26 $\pm$ 0.12} & \underline{0.26 $\pm$ 0.12} & \underline{0.26 $\pm$ 0.12} & \textbf{0.30 $\pm$ 0.13} & \textbf{0.30 $\pm$ 0.13} \\
\midrule
\multirow{3}{*}{15} & 0 & 0.22 $\pm$ 0.11 & 0.05 $\pm$ 0.05 & 0.26 $\pm$ 0.11 & 0.20 $\pm$ 0.11 & 0.06 $\pm$ 0.00 & 0.30 $\pm$ 0.12 & 0.30 $\pm$ 0.12 & 0.29 $\pm$ 0.11 & 0.30 $\pm$ 0.12 & \textbf{0.34 $\pm$ 0.13} & \underline{0.33 $\pm$ 0.13} \\
 & 1 & 0.22 $\pm$ 0.11 & 0.04 $\pm$ 0.05 & 0.23 $\pm$ 0.10 & 0.19 $\pm$ 0.11 & 0.07 $\pm$ 0.00 & \underline{0.29 $\pm$ 0.11} & \underline{0.29 $\pm$ 0.11} & \underline{0.29 $\pm$ 0.12} & \underline{0.29 $\pm$ 0.11} & \textbf{0.33 $\pm$ 0.12} & \textbf{0.33 $\pm$ 0.12} \\
 & 2 & 0.23 $\pm$ 0.11 & 0.04 $\pm$ 0.05 & 0.22 $\pm$ 0.10 & 0.18 $\pm$ 0.11 & 0.07 $\pm$ 0.00 & 0.29 $\pm$ 0.11 & 0.29 $\pm$ 0.11 & 0.29 $\pm$ 0.12 & 0.29 $\pm$ 0.11 & \textbf{0.33 $\pm$ 0.12} & \underline{0.32 $\pm$ 0.12} \\
\midrule
\multirow{3}{*}{20} & 0 & 0.24 $\pm$ 0.10 & 0.06 $\pm$ 0.05 & 0.27 $\pm$ 0.11 & 0.20 $\pm$ 0.11 & 0.06 $\pm$ 0.00 & \underline{0.32 $\pm$ 0.11} & 0.31 $\pm$ 0.11 & 0.31 $\pm$ 0.11 & \underline{0.32 $\pm$ 0.11} & \textbf{0.34 $\pm$ 0.12} & \textbf{0.34 $\pm$ 0.12} \\
 & 1 & 0.24 $\pm$ 0.10 & 0.05 $\pm$ 0.05 & 0.26 $\pm$ 0.10 & 0.19 $\pm$ 0.10 & 0.07 $\pm$ 0.00 & 0.31 $\pm$ 0.11 & 0.31 $\pm$ 0.11 & 0.31 $\pm$ 0.11 & 0.31 $\pm$ 0.11 & \textbf{0.34 $\pm$ 0.12} & \underline{0.33 $\pm$ 0.12} \\
 & 2 & 0.24 $\pm$ 0.10 & 0.05 $\pm$ 0.05 & 0.24 $\pm$ 0.09 & 0.19 $\pm$ 0.10 & 0.07 $\pm$ 0.00 & \underline{0.31 $\pm$ 0.11} & \underline{0.31 $\pm$ 0.11} & 0.30 $\pm$ 0.11 & \underline{0.31 $\pm$ 0.11} & \textbf{0.33 $\pm$ 0.12} & \textbf{0.33 $\pm$ 0.12} \\
\midrule
\end{tabular}}

\end{table*}

\clearpage

\section{Example of explanations}
\label{sec:appendix:other_explanations}

\subsubsection{Example of \xmt Explanations}
\label{sub:xmt_explanations_example}

To illustrate how \xmt generates explanations, we examine here a real explanation for a prediction of the following sample taken from the \coopdataset dataset:

\begin{itemize}
\item \textbf{Customer ID:} 72764
\item \textbf{Shopping Date:} September 9th, 2013
\item \textbf{Current Basket (9 items):} rice, snacks, prepared food, bread, tomatoes, hazelnut fruit, wine, olive oil, cold cuts
\item \textbf{Prediction Size:} 5 items
\item \textbf{Predicted Forgotten Items:} drinks, dried fruit, rabbit, yogurt, vegetables
\end{itemize}

For each predicted item, \xmt generated the following evidence-based explanations:

\textbf{Item: vegetables}
\vspace{-1.0em}
\begin{enumerate}
\item Last purchased 3 days ago (typically bought every 2.0 days)
\item Often bought with current basket items: bread, cold cuts, wine
\item Often repurchased soon after large shopping trips (71.3\% of opportunities)
\end{enumerate}

\textbf{Item: drinks}
\vspace{-1.0em}
\begin{enumerate}
\item Last purchased 4 days ago (typically bought every 3.0 days)
\item Often bought with current basket items: bread, cold cuts, wine
\item Often repurchased soon after large shopping trips (50.2\% of opportunities)
\end{enumerate}

\textbf{Item: rabbit}
\vspace{-1.0em}
\begin{enumerate}
\item Last purchased 6 days ago (typically bought every 3.0 days)
\item Often bought with current basket items: bread, cold cuts, wine
\item Often repurchased soon after large shopping trips (40.3\% of opportunities)
\end{enumerate}

\textbf{Item: dried fruit}
\vspace{-1.0em}
\begin{enumerate}
\item Last purchased 4 days ago (typically bought every 3.0 days)
\item Often bought with current basket items: bread, cold cuts, wine
\item Often repurchased soon after large shopping trips (34.1\% of opportunities)
\end{enumerate}

\textbf{Item: yogurt}
\vspace{-1.0em}
\begin{enumerate}
\item Last purchased 2 days ago (typically bought every 3.0 days)
\item Often bought with current basket items: bread, cold cuts, wine
\item Often repurchased soon after large shopping trips (47.2\% of opportunities)
\end{enumerate}

These explanations demonstrate how XMT\xmt combines temporal patterns, co-purchase relationships, and repurchase behaviors to provide transparent, evidence-based justifications for its predictions.

\subsubsection{Example of \txmt Explanations}

To demonstrate how \txmt enhances the explanation framework through TARS integration, we consider the prediction done on the same sample used for \xmt in ~\Cref{sub:xmt_explanations_example}:

\begin{itemize}
\item \textbf{Customer ID:} 72764
\item \textbf{Shopping Date:} September 9th, 2013
\item \textbf{Current Basket (9 items):} rice, snacks, prepared food, bread, tomatoes, hazelnut fruit, wine, olive oil, cold cuts
\item \textbf{Prediction Size:} 5 items
\item \textbf{Predicted Forgotten Items:} yogurt, rabbit, vegetables, dried fruit, drinks
\end{itemize}

Notice that while \txmt predicts the same items as \xmt for this particular sample, its explanations incorporate additional TARS-based evidence:

\textbf{Item: vegetables}
\vspace{-1.0em}
\begin{enumerate}
\item Last purchased 3 days ago (typically bought every 2.0 days)
\item Often bought with current basket items: bread, cold cuts, wine
\item TARS pattern analysis suggests this item is likely to be needed (confidence: 50.5\%)
\item Often repurchased soon after large shopping trips (71.3\% of opportunities)
\end{enumerate}

\textbf{Item: drinks}
\vspace{-1.0em}
\begin{enumerate}
\item Last purchased 4 days ago (typically bought every 3.0 days)
\item Often bought with current basket items: bread, cold cuts, wine
\item TARS pattern analysis suggests this item is likely to be needed (confidence: 39.0\%)
\item Often repurchased soon after large shopping trips (50.2\% of opportunities)
\end{enumerate}

\textbf{Item: rabbit}
\vspace{-1.0em}
\begin{enumerate}
\item Last purchased 6 days ago (typically bought every 3.0 days)
\item Often bought with current basket items: bread, cold cuts, wine
\item TARS pattern analysis suggests this item is likely to be needed (confidence: 29.5\%)
\item Often repurchased soon after large shopping trips (40.3\% of opportunities)
\end{enumerate}

\textbf{Item: dried\_fruit}
\vspace{-1.0em}
\begin{enumerate}
\item Last purchased 4 days ago (typically bought every 3.0 days)
\item Often bought with current basket items: bread, cold cuts, wine
\item TARS pattern analysis suggests this item is likely to be needed (confidence: 28.6\%)
\item Often repurchased soon after large shopping trips (34.1\% of opportunities)
\end{enumerate}

\textbf{Item: yogurt}
\vspace{-1.0em}
\begin{enumerate}
\item Last purchased 2 days ago (typically bought every 3.0 days)
\item Often bought with current basket items: bread, cold cuts, wine
\item TARS pattern analysis suggests this item is likely to be needed (confidence: 37.2\%)
\item Often repurchased soon after large shopping trips (47.2\% of opportunities)
\end{enumerate}

The addition of TARS-based confidence scores provides users with a quantitative measure of how strongly each prediction is supported by recurring temporal patterns in their shopping history. This enhancement adds another layer to the explanations while maintaining the original, evidence-based structure of the original \xmt algorithm.

\section{Visualization of the explanations}
\label{sec:appendix:visualization}

\begin{figure*}
\centering
\includegraphics[width=\linewidth]{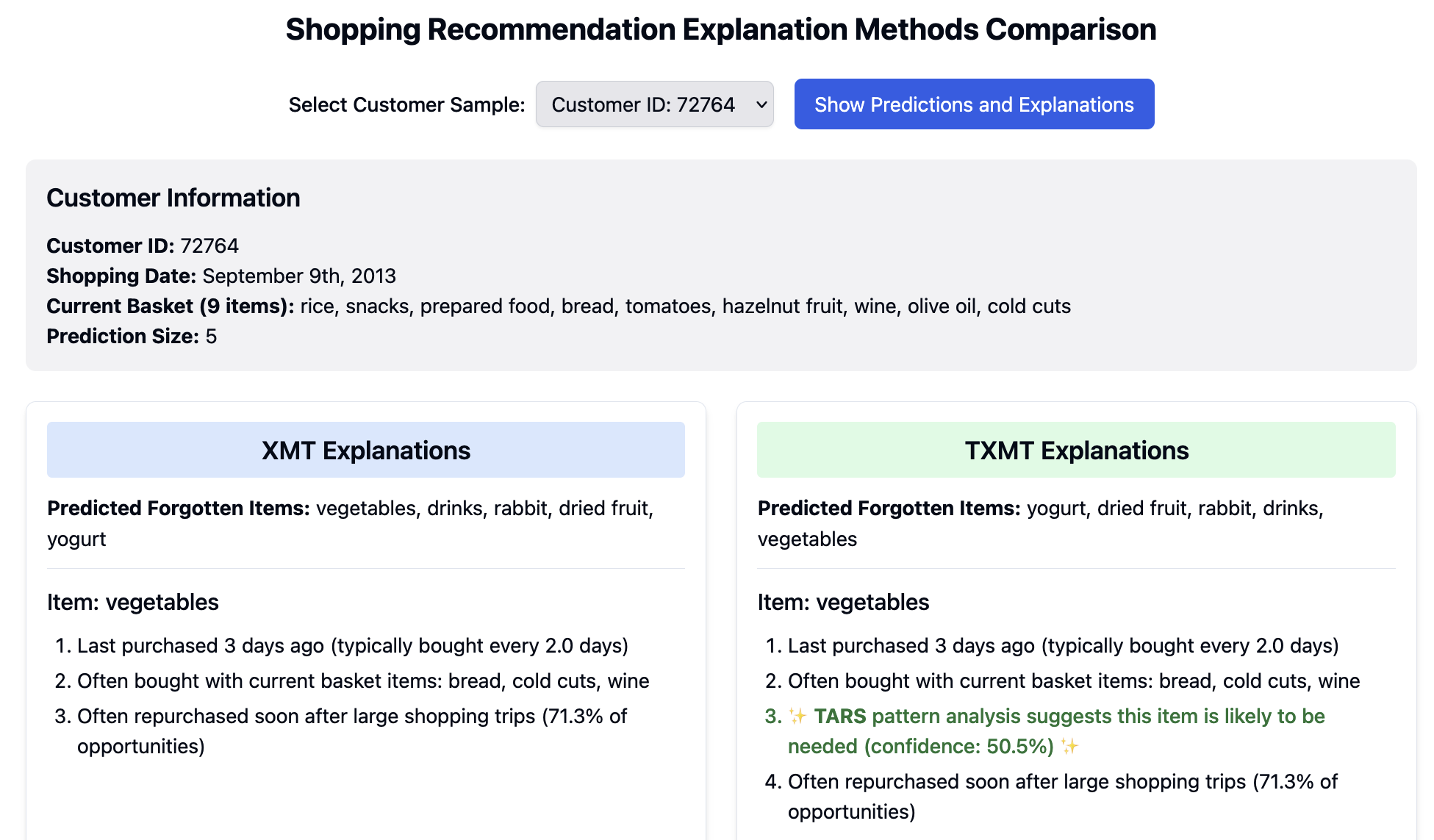}

\caption{Visualization of an explanation of a prediction obtained with \xmt and \txmt}
\label{fig:visualization}
\end{figure*}

Alongside the development of the two proposed algorithms, \xmt and \txmt, we developed a dashboard to visualize the predictions and the explanations. We report in~\Cref{fig:visualization} a screenshot taken from the dashboard that shows the differences between the explanation obtained using \xmt and \txmt. In particular, we highlight in green the TARS pattern extracted by the \txmt algorithm that increases the interpretability of the explanation. 

\end{document}